\documentclass{article}

\usepackage{microtype}
\usepackage{graphicx}
\usepackage{subcaption}
\usepackage{booktabs} 

\usepackage{hyperref}

\usepackage[preprint]{icml2026}

\usepackage{amsmath}
\usepackage{amssymb}
\usepackage{mathtools}
\usepackage{amsthm}
\usepackage{caption}
\usepackage{bm}
\usepackage{multirow}
\usepackage[most]{tcolorbox}
\usepackage{xcolor}
\usepackage{enumitem}

\definecolor{softbg}{RGB}{240, 244, 248} 
\definecolor{framegray}{RGB}{96, 110, 126} 
\definecolor{titlebg}{RGB}{220, 227, 233} 

\newtcolorbox[auto counter, number within=section]{WideBox}[2][]{
  enhanced,
  colback=softbg,
  colframe=framegray,
  coltitle=black,
  colbacktitle=titlebg,
  width=\textwidth,
  boxrule=0.5mm,
  arc=3mm,
  drop fuzzy shadow,
  title={\textbf{Example \thetcbcounter:} #2}, 
  fonttitle=\sffamily,
  attach boxed title to top left={xshift=4mm, yshift=-2mm},
  boxed title style={size=small, colframe=framegray, arc=2mm},
  #1
}

\usepackage[capitalize,noabbrev]{cleveref}

\theoremstyle{plain}

\theoremstyle{definition}

\theoremstyle{remark}

\usepackage[textsize=tiny]{todonotes}

\icmltitlerunning{How Do Language Models Understand Tables? A Mechanistic Analysis of Cell Location}

\begin{document}

\twocolumn[
  \icmltitle{How Do Language Models Understand Tables?\\A Mechanistic Analysis of Cell Location}

    \icmlsetsymbol{equal}{*}

  \begin{icmlauthorlist}
    \icmlauthor{Xuanliang Zhang}{hit}
    \icmlauthor{Dingzirui Wang}{hit}
    \icmlauthor{Keyan Xu}{hit}
    \icmlauthor{Qingfu Zhu}{hit}
    \icmlauthor{Wanxiang Che}{hit}
  \end{icmlauthorlist}

  \icmlaffiliation{hit}{Harbin Institute of Technology}

  \icmlcorrespondingauthor{Xuanliang Zhang}{xuanliangzhang@ir.hit.edu.cn}
  \icmlcorrespondingauthor{Wanxiang Che}{car@ir.hit.edu.cn}

  \icmlkeywords{Machine Learning, ICML}

  \vskip 0.3in
]



\printAffiliationsAndNotice{}  

\begin{abstract}
While Large Language Models (LLMs) are increasingly deployed for table-related tasks, the internal mechanisms enabling them to process linearized two-dimensional structured tables remain opaque. In this work, we investigate the process of table understanding by dissecting the atomic task of cell location. Through activation patching and complementary interpretability techniques, we delineate the table understanding mechanism into a sequential three-stage pipeline: Semantic Binding, Coordinate Localization, and Information Extraction. 
We demonstrate that models locate the target cell via an ordinal mechanism that counts discrete delimiters to resolve coordinates. Furthermore, column indices are encoded within a linear subspace that allows for precise steering of model focus through vector arithmetic.
Finally, we reveal that models generalize to multi-cell location tasks by multiplexing the identical attention heads identified during atomic location. Our findings provide a comprehensive explanation of table understanding within Transformer architectures.
\end{abstract}

\section{Introduction}
Tables serve as a ubiquitous format for organizing two-dimensional structured data across real-world applications \cite{badaro-etal-2023-transformers-table-survey}. Given the exceptional capabilities of Large Language Models (LLMs) in natural language comprehension and reasoning, researchers have increasingly applied these models to table-related tasks \cite{fang2024tabular-survey,zhang2025survey-table}. However, the Transformer models necessitates the linearization of two-dimensional structures into one-dimensional sequences, which differs from linearized natural language and could limit their understanding. 
While numerous methods have been proposed to enhance performance through prompt optimization, reasoning decomposition, or specialized training \cite{zhang-etal-2025-rot,yang2025triples,li2025table-modality}, the internal mechanisms governing how models understand tabular data remain largely opaque. This lack of transparency regarding the underlying dynamics significantly impedes the development of mechanistic interpretability and reasoning reliability.

\begin{figure}[t]
    \centering
    \includegraphics[width=.8\linewidth]{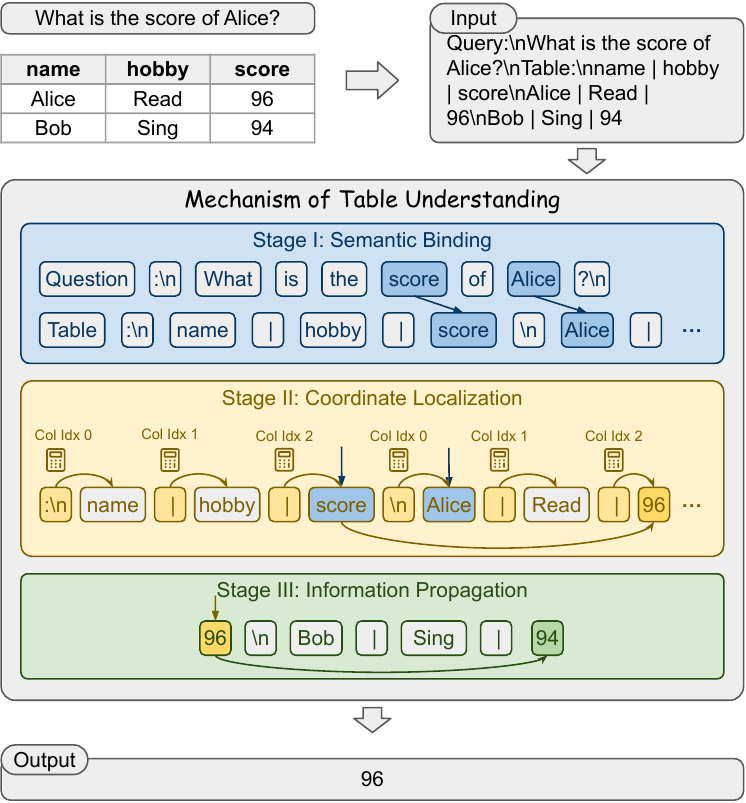}
    \caption{
      The mechanism interpretation of the sequential three-stage pipeline for table understanding. 
    }
    \label{fig:intro}
\end{figure}

Prior works in mechanistic interpretability primarily address tasks such as factual recall or indirect object identification \cite{wang2024grokking,ye2025how-transformers-implicit,huben2024sae,gan2026mechanism-survey}. 
However, \textbf{no existing study investigates the internal mechanics of table understanding}. 
This gap is critical because table understanding differs fundamentally from natural language understanding: it demands that the model adhere to rigid spatial coordinates within a linearized input. Deciphering the mechanisms is therefore a prerequisite for understanding how LLMs bridge the gap between sequential processing and two-dimensional structural logic.

We address this gap by discovering the atomic task of \textit{cell location}. We select this task as our focal point because it demands the synergistic execution of both semantic alignment (matching headers) and structural reasoning (navigating coordinates), thereby serving as a minimal yet comprehensive proxy for broader table understanding capabilities.
We analyze the internal states of state-of-the-art open-weights models including the Qwen and Llama series to discover the information flow from input headers to output cells.

To unravel the internal mechanics of this process, we employ activation patching analysis, cross-validated via linear probing and attention ablation. We delineate the table understanding mechanism into a sequential three-stage pipeline: \textbf{Semantic Binding}, \textbf{Coordinate Localization}, and \textbf{Information Extraction}, as shown in Figure~\ref{fig:intro}.
The process commences with Semantic Binding in early layers, where the model aligns query constraints with table headers. 
Then in Coordinate Localization, the model abstracts away from semantic content to navigate structural indices. 
The final stage involves Information Extraction in late layers, where the targeted value is propagated to the output distribution.

To further explore the nature of the Coordinate Localization phase, we examine the internal representations using linear probes and noise intervention experiments. We demonstrate that the model locate the target cell via a ordinal, distance-invariant mechanism that \textbf{counts discrete delimiters} to resolve coordinate states before propagating them to cell content. 
Furthermore, to investigate the geometric properties of this representation, we inject directional shift vectors into the residual stream. We find that scaling the unit vector successfully steers the model to focus on specific relative column positions, and that vectors representing different offsets exhibit additive compositionality, confirming that column indices are encoded within \textbf{a linear subspace navigated via vector arithmetic}.

To assess the scalability of these mechanisms to complex structural constraints, we extend our analysis to multi-cell location tasks. We observe that the models does not instantiate distinct mechanisms for complex tasks but rather \textbf{reuses the identical attention heads} identified for atomic cell location. By employing parallel semantic binding and multi-threaded coordinate localization, the models effectively multiplex these atomic operations to resolve multiple row and column constraints simultaneously.

We summarize our primary contributions as follows.
\begin{itemize}[nolistsep,leftmargin=*]
    \item We utilize activation patching to formalize the table understanding process as three stages: Semantic Binding, Coordinate Localization, and Information Extraction. 
    \item We explore the Coordinate Localization stage using linear probes and steering vectors. We discover that the models maintain an implicit coordinate system by counting delimiters and represent column indices in a linear subspace.
    \item We verify the generalization of these mechanisms and show that the models reuses the identical set of attention heads to process multi-cell location in parallel.
\end{itemize}

\section{Preliminaries}
    \label{sec:preliminaries}
    In this section, we formally define the task, introduce the models employed, and detail the mechanistic interpretability techniques used to dissect the internal location processing.

\subsection{Task Formulation: Cell Location}
We focus on the fundamental operation of table understanding: locating the specific cell value given its corresponding row and column headers, which is a common subproblem of complex table-related tasks \cite{pasupat-liang-2015-wikitq}. 
We select cell location as it requires the model to not only perform semantic alignment but also to execute structural reasoning by navigating the coordinates \cite{wang2025needleinatable}.
We first explore the atomic cell location task, and generalize our findings to the multi-cell location task.
Formally, let a table $T$ be defined by a set of column headers $\mathcal{C} = \{c_1, \dots, c_C\}$ and a set of row headers $\mathcal{R} = \{r_1, \dots, r_R\}$, where row headers correspond to the unique identifiers in the primary key column. The cell value located at the intersection of row header $r_i$ and column header $h_j$ is denoted as $v_{r_i,h_j}$. The table is serialized into a sequence of tokens $x_{table}$ using a Markdown format with pipe delimiters (`$\vert$') and newlines.

The model is presented with a prompt $x = [D, x_{table}, x_{query}]$, where $D$ represents a one-shot demonstration. 
The query $x_{query}$ is formulated as a natural language question that explicitly references a target row header constraint $r \in \mathcal{R}$ and a column header constraint $c \in \mathcal{C}$.
The objective is to generate the target token sequence $y = v_{r, c}$.
To rigorously isolate structural reasoning from parametric knowledge memorization, we generate synthetic tables where cell values are randomly sampled from an entity pool, ensuring the model must rely solely on the provided context to align the requested headers with the table structure \cite{cheng2025survey-data-contamination,wu2025memorization-contamination}.
Therefore, we construct a benchmark comprising $500$ independent samples to evaluate the performance on this atomic cell location task, with details in Appendix~\ref{app:data}.

\subsection{Models}
To ensure the generality of our findings, we conduct our analysis on a diverse set of decoder-only Transformer language models.
Specifically, we utilize the \textbf{Qwen} series (including Qwen3-0.6B, Qwen3-4B, Qwen2.5-32B)~\cite{yang2025qwen3technicalreport} and the \textbf{Llama} series (including Llama-3.2-3B, Llama-3.1-8B)~\cite{grattafiori2024llama3}. 
These models employ Rotary Positional Embeddings (RoPE)~\cite{su2024rope} and represent state-of-the-art open-weights architectures. 
We primarily report the results of Qwen3-4B as it achieves a high accuracy of $90\%$ on our benchmark.
Our core findings consistently generalize across the other evaluated models, with detailed results and analysis in Appendix~\ref{app:llms}.

\subsection{Analytic Techniques}
\label{subsec:techniques}
To unravel the mechanism of table understanding, we employ a suite of mechanistic interpretability techniques.

\paragraph{Activation Patching}
We employ Activation Patching to localize specific information flow within the model. Let $\mathbf{h}_t^{(l)} \in \mathbb{R}^d$ denote the activation vector (residual stream) at layer $l$ and token position $t$. To evaluate the causal contribution of a specific component, we perform an intervention by replacing its activation in a clean run (input $x_{\text{clean}}$, target $y$) with the corresponding activation from a corrupted run (input $x_{\text{corrupt}}$, foil $y'$).
We consider two levels of granularity for patching, following previous works \cite{wang2023interpretability-wild,ye2025how-transformers-implicit,li2025track-state}:
\begin{itemize}[nolistsep,leftmargin=*]
    \item \textbf{Layer-level Patching:} We replace the residual stream $\mathbf{h}_t^{(l)}$ across the specific token span $t \in \mathcal{T}$ to identify which layers and positions mediate the location process.
    \item \textbf{Head-level Patching:} We replace the output of a specific attention head $H^{(l, a)}$ (where $a \in \{1, \dots, A\}$ denotes the head index) at the final input token position. This details the causal contributions of precise heads.
\end{itemize}
To quantify the importance of the patched component, we use the \textit{Effect Score} based on the Logit Difference (LD) \cite{hong2025a-implies-b}. We define $\text{LD}(x) = \text{logit}(x)_{y} - \text{logit}(x)_{y'}$ as the confidence margin between the target and the foil. The Effect Score is the normalized change in LD:
\begin{equation}
    \text{Effect} = \frac{\text{LD}_{\text{patch}} - \text{LD}_{\text{corrupt}}}{\text{LD}_{\text{clean}} - \text{LD}_{\text{corrupt}}}
\end{equation}
where $\text{LD}_{\text{patch}}$ denotes the logit difference after intervention. Scores near or above $1.0$ indicate that the patched component is a pivotal mediator of the task.

\paragraph{Linear Probing}
To investigate whether the model explicitly encodes geometric structures within its internal representations, we employ linear probing on the residual stream. For a dataset of $N$ samples, we train a linear regressor $P$ on the activations $\mathbf{h}_t^{(l)}$ to predict the ground-truth geometric coordinates $z \in \{r_{idx}, c_{idx}\}$. These represent the numerical row and column indices of the token in the table grid:
\begin{equation}
    \hat{z} = P(\mathbf{h}_t^{(l)}) = \mathbf{W}_p \mathbf{h}_t^{(l)} + \mathbf{b}_p
\end{equation}
where $\mathbf{W}_p \in \mathbb{R}^{1 \times d}$ and $\mathbf{b}_p \in \mathbb{R}$ are learnable parameters. We evaluate the probe using the coefficient of determination ($R^2$ score):
\begin{equation}
    R^2 = 1 - \frac{\sum_{i=1}^{N} (z_i - \hat{z}_i)^2}{\sum_{i=1}^{N} (z_i - \bar{z})^2}
\end{equation}
A high $R^2$ score indicates that the model linearly encodes the structural coordinates of the table. We provide training and evaluation details in Appendix~\ref{subapp:probing_details}.


\paragraph{Ablation Studies}
To verify the causal necessity of specific attention heads, we employ ablation techniques. The total output of the multi-head attention (MHA) block at layer $l$ and position $t$ is the aggregation of independent head outputs: $\text{MHA}(\mathbf{h}_t^{(l)}) = \sum_{a=1}^{A} H_t^{(l,a)}$. We define a target set of heads $\mathcal{H}_{abl}$ to intervene upon.
\textbf{Zero Ablation} explicitly removes the information flow from the target heads by forcing their output activations to zero \cite{gould2024successor-heads}:
\begin{equation}
    \hat{H}_t^{(l,a)} = \mathbf{0}, \quad \forall (l,a) \in \mathcal{H}_{abl}
\end{equation}
By measuring the change of outputs, we observe whether the targeted heads are necessary components for cell location.


\section{The Three-Stage Information Flow of Table Understanding}
    \label{sec:understanding}
    \subsection{Overview}
\label{subsec:patch_overview}

\begin{figure}[t]
    \centering
    \includegraphics[width=.9\linewidth]{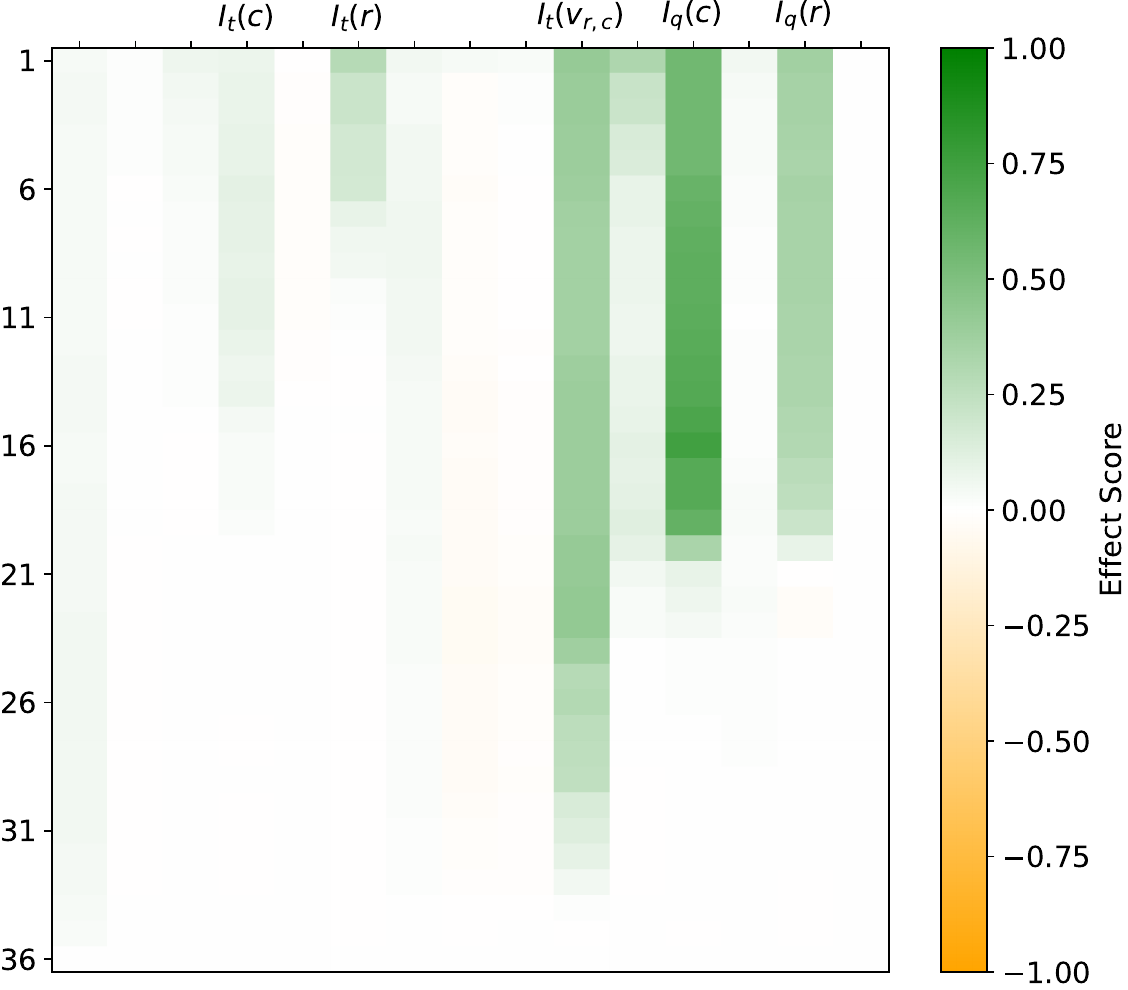}
    \caption{
      Average column patching effect score across layers and token positions. 
    }
    \label{fig:col_patch}
\end{figure}

\paragraph{Experimental Setup}
\label{sec:roi_def}
To map the information flow of table understanding, we employ activation patching on semantically segmented input sequences. 
Let $\mathcal{I}_{q}(r)$ and $\mathcal{I}_{q}(c)$ denote the query token spans for row and column constraints. Within the fixed table $x_{table}$, we define $\mathcal{I}_{t}(c)$ and $\mathcal{I}_{t}(r)$ as the token spans for the column header $c$ and row headers $r$, and $\mathcal{I}_{t}(v_{r,c})$ as the span for the cell value located by $r$ and $c$.
We construct corrupted inputs $x'$ by altering the query constraints to a counterfactual column or row header, thereby shifting the target value location. We define two settings:
\begin{itemize}[nolistsep,leftmargin=*]
    \item \textbf{Column Patching:} We replace the query column constraint $c$ with a counterfactual $c'$. This shifts the target value indices from the correct span $\mathcal{I}_{t}(v_{r, c})$ to the counterfactual span $\mathcal{I}_{t}(v_{r, c'})$.
    \item \textbf{Row Patching:} We replace the query row constraint $r$ with $r'$. This shifts the target indices to $\mathcal{I}_{t}(v_{r', c})$.
\end{itemize}
Figure~\ref{fig:col_patch} shows the Column Patching results, with detailed  configurations and explanations of coordinates in Appendix~\ref{subapp:patch_details}. Given the symmetric spatiotemporal patterns, results of Row Patching are provided in Appendix~\ref{subapp:patch_details}.

\paragraph{Analysis of Causal Dynamics}
Observing the distribution of Effect Scores across layers and positions reveals a sequential transition in causal importance, delineating the location process into three stages. 
Notably, this stratification reflects the concentration of the primary functional attention heads within these intervals but does not preclude the existence of a minority of relevant heads in other layers:

\begin{enumerate}[nolistsep,leftmargin=*]
    \item \textbf{Stage I: Semantic Binding (Early Layers 1-16).}
    Significant causal mass is concentrated on query constraints ($\mathcal{I}_{q}(r), \mathcal{I}_{q}(c)$), whereas the impact on corresponding table headers ($\mathcal{I}_{t}(r), \mathcal{I}_{t}(c)$) is comparatively marginal. 
    
    \item \textbf{Stage II: Coordinate Localization (Middle Layers 17-23).}
    The causal importance shift decisively from the headers to the target cell region ($\mathcal{I}_{t}(v_{r, c})$). This transition suggests a functional shift from semantic matching to structural navigation, where the model utilizes the constraint header information to locate the specific intersection coordinates in the serialized table sequence.
    
    \item \textbf{Stage III: Information Propagation (Late Layers 24+).}
    Interventions on all token positions yield negligible causal effects in this stage. This diminished sensitivity indicates that the answer $v_{r, c}$ has already been resolved and encoded within the internal states. Consequently, the primary mechanism shifts from location to the propagation to the final token position for decoding.
\end{enumerate}


\begin{figure}[t]
    \centering
    \begin{subfigure}[b]{0.48\linewidth} 
        \centering
        \includegraphics[width=\linewidth]{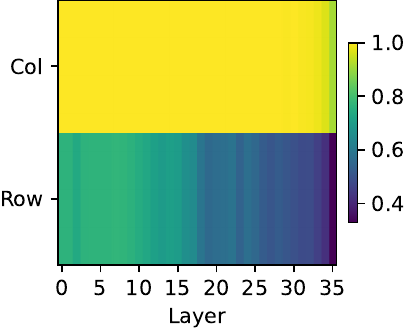}
        \caption{Layer-wise Alignment}
        \label{fig:row_col_similarity}
    \end{subfigure}
    \hfill
    \begin{subfigure}[b]{0.48\linewidth}
        \centering
        \includegraphics[width=\linewidth]{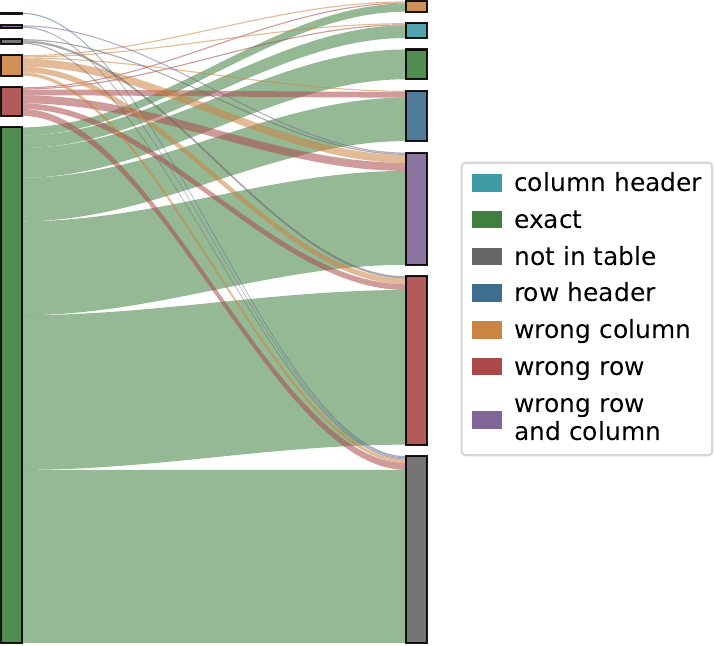}
        \caption{Ablation Error Flow}
        \label{fig:sankey_row}
    \end{subfigure}
    
    \caption{Analysis of semantic binding mechanisms. \textbf{(a)} Top-1 accuracy evaluated with cosine similarity between query constraints and all table headers across layers. \textbf{(b)} The impact of ablating row alignment heads by observing the changes of outputs.}
    \label{fig:both_figures}
\end{figure}

\subsection{Stage I: Semantic Binding}
\label{subsec:stage 1}

To investigate the semantic binding of query constraints to table headers, we first analyze the geometric alignment of their internal representations. We compute the cosine similarity between hidden states of query constraints and candidate table headers at each layer, and evaluate using the top-1 accuracy, defined as the frequency with which the gold header exhibits the highest similarity score among all candidates.
As illustrated in Figure~\ref{fig:row_col_similarity}, the accuracy for both dimensions converges close to $1.0$ in early layers, indicating that semantic mapping occurs well before information extraction. Notably, column headers exhibit tighter alignment than rows, suggesting easier binding, whereas row constraint resolution is strictly confined to the initial layers.

We further explore the causal mechanism via attention ablation. 
We identify specialized alignment heads, which exhibit preferential attention from query constraints to target table headers, with detailed setup in Appendix~\ref{subapp:ablation}. 
In Figure~\ref{fig:sankey_row}, only the \texttt{exact} flow represents correct predictions, while other flows ar error types.
Zero-ablating the top-20 of \textit{row alignment heads} ($95\%$ in layers 1-16) triggers a catastrophic performance collapse from $94.4\%$ to $5.2\%$.
Conversely, ablating the top-20 column alignment heads or a random selection of heads results in marginal performance degradations of $6.6\%$ and $13.8\%$, respectively. 
Notably, random head ablation results in a mere $11.0\%$ \texttt{wrong row} (the output in the correct column but error row) rate, which is significantly lower than $29.4\%$ with row alignment heads, confirming that the identified heads are responsible for binding row header semantic information.
Therefore, row grounding necessitates explicit semantic attention, column localization operates independently of strong attention weights, suggesting a process driven by signals beyond semantic alignment.

\begin{figure}[t]
    \centering
    \includegraphics[width=.8\linewidth]{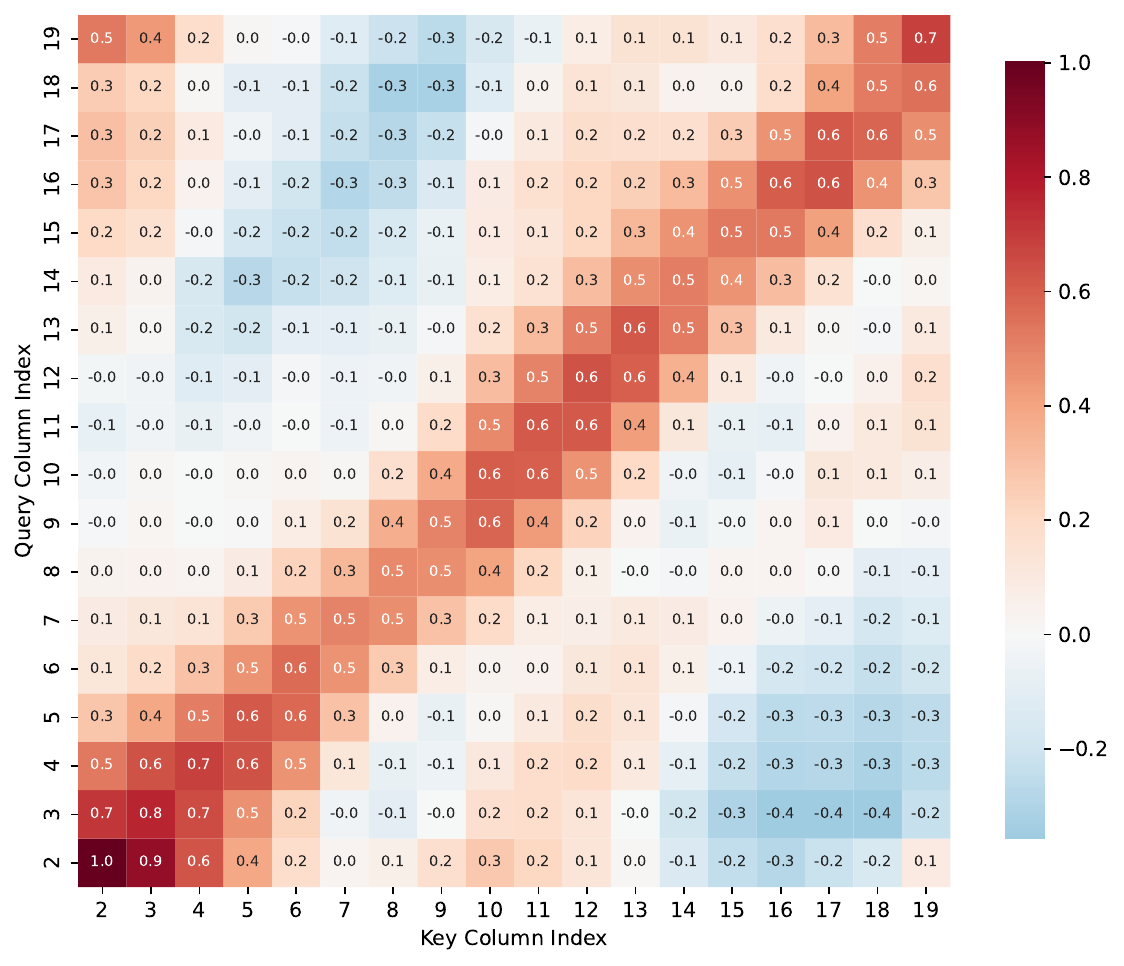}
    \caption{
        Visualization of the geometric interaction score $S_{j,l}$ in coordinate-encoding heads. 
        The heatmap illustrates the dot product between column-averaged query vectors ($\bar{q}_j$) and key vectors ($\bar{k}_l$) after injecting RoPE corresponding to their column indices. 
    }
    \label{fig:average_qk_heatmap}
\end{figure}

\subsection{Stage II: Coordinate Localization}
\label{subsec:stage 2}

Following semantic alignment, the model transitions to locating the geometric coordinates of the target cell. Specifically, once the row header is identified, the model must locate the target cell corresponding to the correct column header. 
To determine whether the model relies on structural geometry rather than semantic matching for this stage, we conduct the analysis of the Query-Key interactions.

We investigate whether specific attention heads explicitly encode column indices via RoPE. We extract query and key vectors from the residual stream and average them based on their respective column indices to eliminate specific semantic content. Let $\bar{q}_j$ denote the average query vector across column constraints where the target is the $j$-th column, and $\bar{k}_l$ denote the average key vector for table cells located in the $l$-th column. We artificially inject positional information using these column indices and compute the geometric interaction score $S_{j,l}$ as:
\begin{equation}
    S_{j,l} = \text{RoPE}(\bar{q}_j, j)^T \text{RoPE}(\bar{k}_l, l)
\end{equation}
where $\text{RoPE}(\cdot, p)$ represents the rotary embedding operation at position index $p$. We select the top 20 attention heads ($90\%$ in layers 17-23) that maximize the contrast between diagonal ($j=l$) and off-diagonal ($j \neq l$) interaction scores.
We term these \textit{coordinate-encoding heads}.

Figure~\ref{fig:average_qk_heatmap} exhibits a distinct diagonal pattern where $S_{j,l}$ is maximized significantly when the query target index matches the key column index. This geometric alignment provides strong evidence that the model utilizes a structural navigation mechanism in the middle layers. 
This suggests that the learned projections are optimized to leverage RoPE properties, facilitating an alignment mechanism based on the relative column indices of the query and the target cell.
This confirms that the model abandons semantic matching in this stage effectively treating the serialized table as a coordinate system to locate the correct value.

\begin{figure}[t]
    \centering
    \includegraphics[width=.9\linewidth]{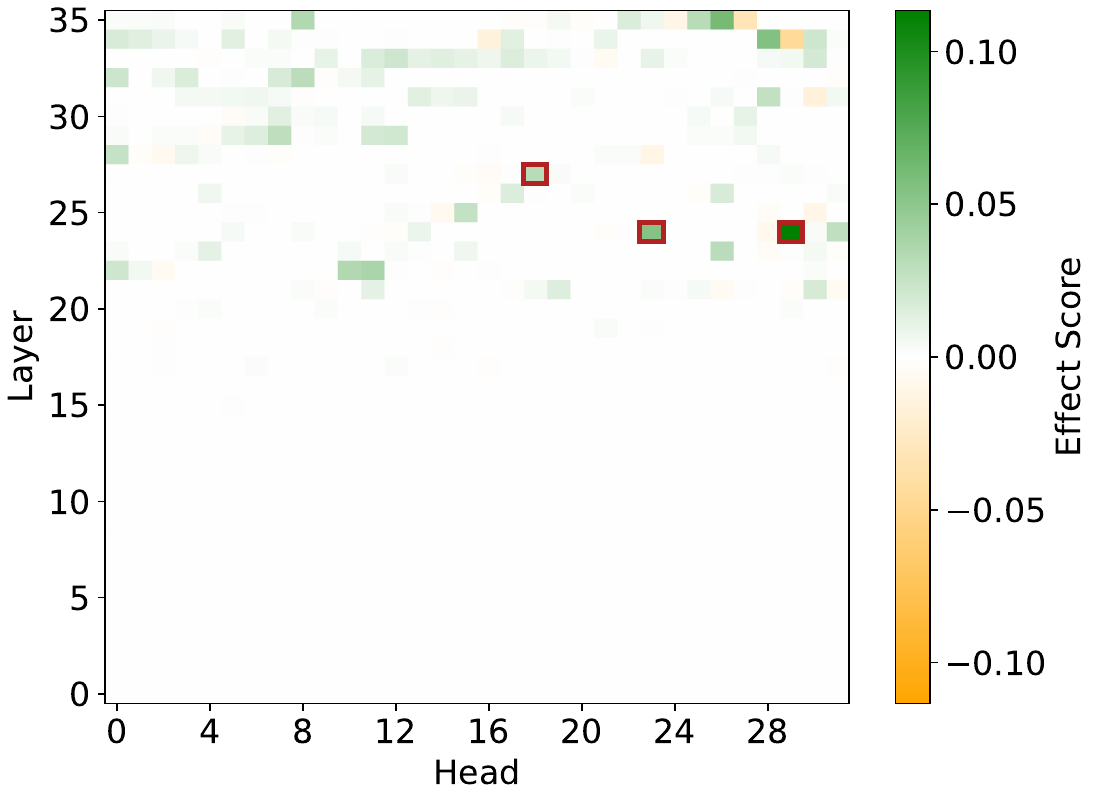}
    \caption{
      Effect Scores of patching attention heads at the last token position. 
      The red boxes highlight heads that exhibit positive Effect Scores and attention the target cell significantly.
    }
    \label{fig:rcba_lt_lg}
\end{figure}

\subsection{Stage III: Information Propagation}
\label{subsec:stage 3}

We investigate the final stage where the model moves the identified content to the last token position for generation. Therefore, we execute activation patching at the position of the last token. 
To isolate the mechanism responsible for location, we maintain the table context $x_{table}$ constant and modify the query $x_{query}$ to target a distinct row header $r'$ and column header $c'$. 
We compute the Effect Score for each attention head to quantify its contribution to predicting the correct target $y$ versus $y'$.

Figure~\ref{fig:rcba_lt_lg} presents the Effect Scores across all layers and heads at the final input token position. We observe that attention heads exhibiting an Effect Score greater than zero become densely distributed starting from Layer 24. To discern the functional mechanism of these high-scoring components we analyze their attention distribution over the input sequence. We discover that the three attention heads highlighted by red boxes allocate significant attention weights to the target cell token $v_{r,c}$ in over $70\%$ of the samples compared to other context tokens. This behavior characterizes them as \textit{information mover heads} that propagate the stored value from the resolved coordinates to the residual stream of the final token for decoding.

\section{Implicit Coordinate System via Delimiter Counting}
    \label{sec:coordinate}
    Having established that the model performs coordinate localization in the middle layers, we now investigate the precise mechanism by which these coordinates are constructed. 
We hypothesize that the model constructs an implicit coordinate system by linearly scanning the serialized sequence.

\begin{figure}[t]
    \centering
    \includegraphics[width=.95\linewidth]{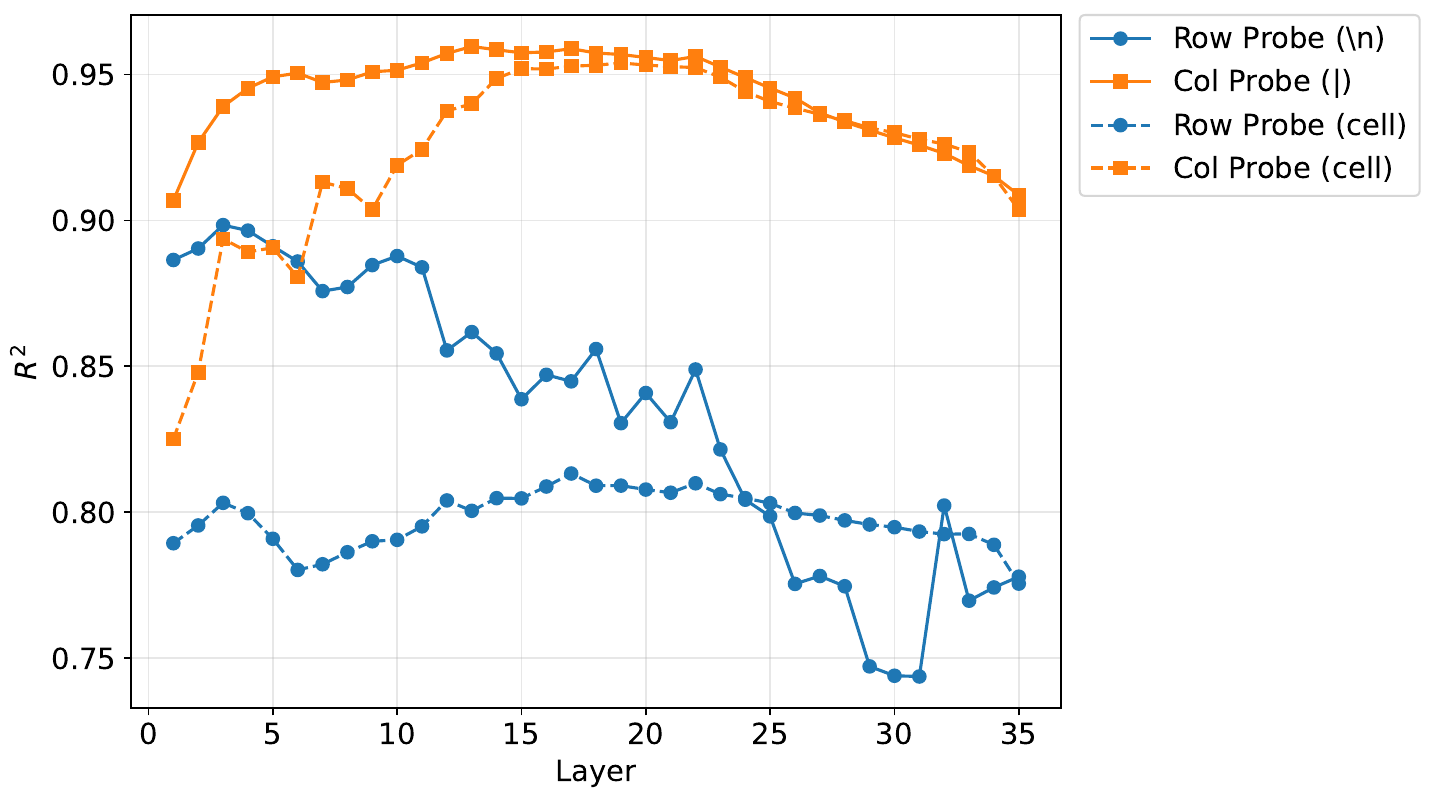}
    \caption{
      $R^2$ of column probes and row probes trained for cells and delimiters across layers. 
    }
    \label{fig:probe_r2}
\end{figure}

\begin{figure*}[t]
    \centering
    \begin{subfigure}[b]{0.45\textwidth} 
        \centering
        \includegraphics[width=\linewidth]{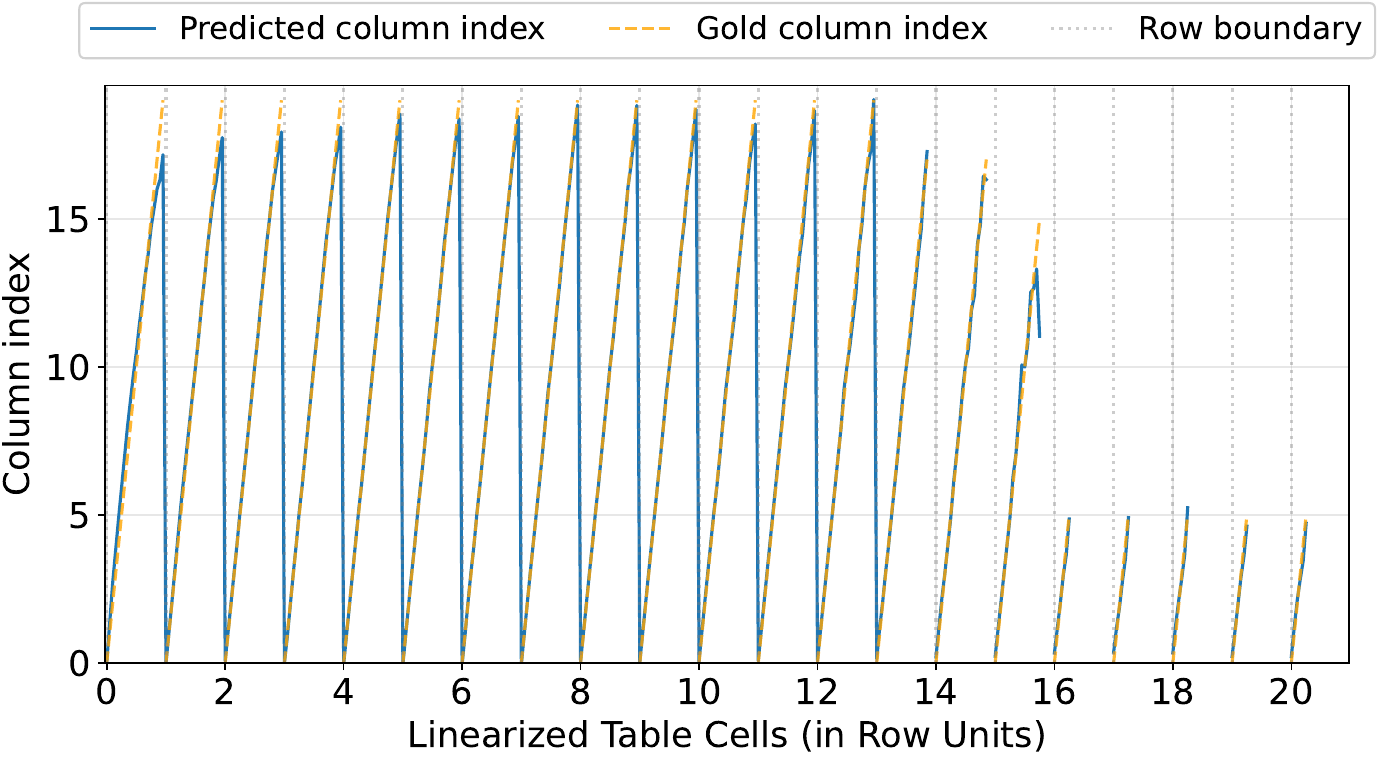}
        \caption{Column prob}
        \label{fig:col_probe_L17}
    \end{subfigure}
    \hfill
    \begin{subfigure}[b]{0.45\textwidth}
        \centering
        \includegraphics[width=\linewidth]{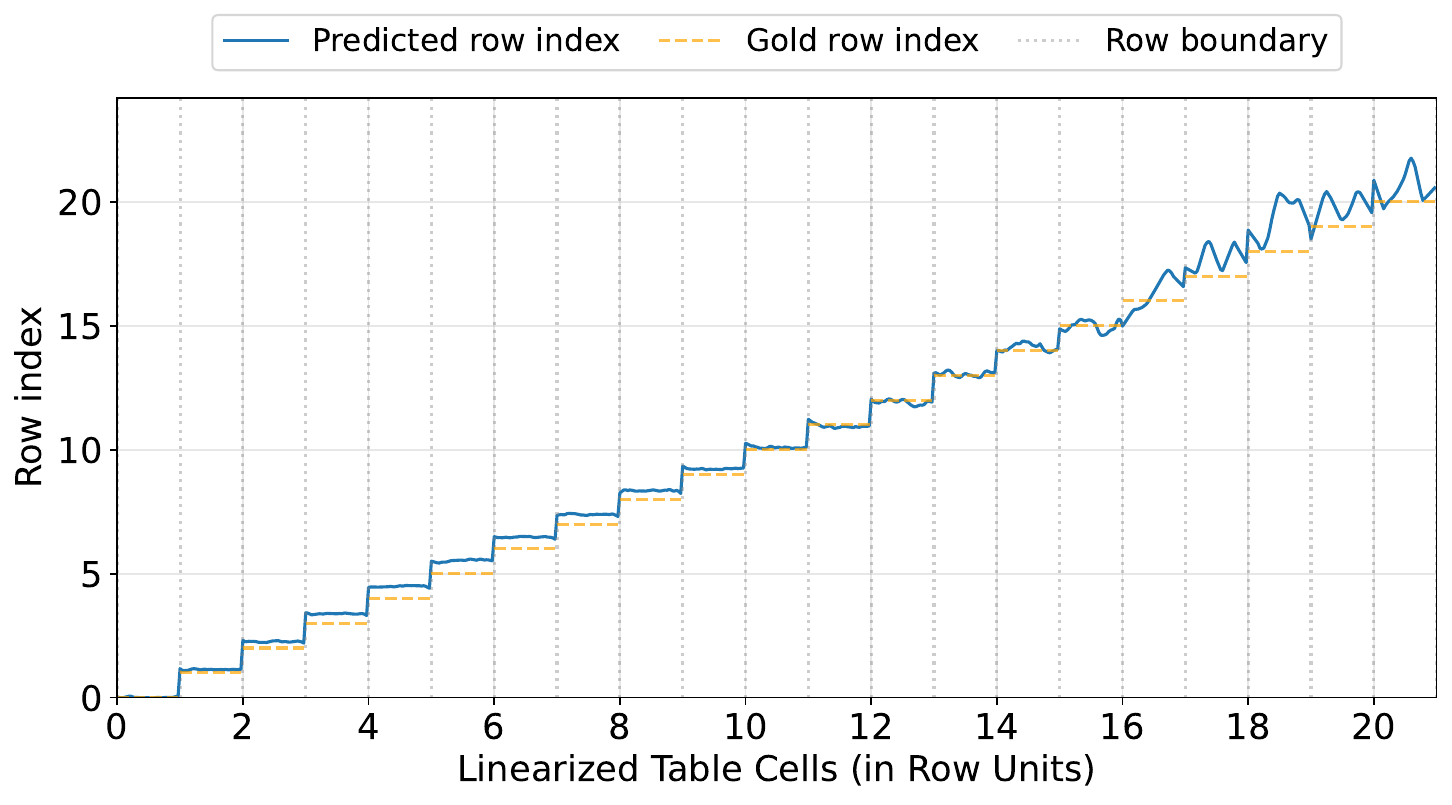}
        \caption{Row prob}
        \label{fig:row_probe_L17}
    \end{subfigure}

    \caption{The probe's predicted column and row indices (blue) versus ground truth (orange) across the linearized table cells at layer 17.}
    \label{fig:probe_predict}
\end{figure*}

\subsection{Representations of Column and Row Indices}
\label{subsec:Representations of Column and Row Indices}

To verify the existence of internal coordinate representations, we analyze the performance of the linear probes trained on the residual stream activations across different layers.
We train independent probes to predict the column index $c_{idx}$ and row index $r_{idx}$ for every cell token in the table sequence $x_{table}$ at each layer. 
We show the average $R^2$ across the tokens to observe the geometric structure representations across layers in Figure~\ref{fig:probe_r2}.
As indicated by the dashed lines, $R^2$ for both column and row probes stabilizes between layers 17 and 23, achieving peak values of $0.954$ and $0.809$ at layer 17.
Notably, the column probes exhibit superior performance compared to row probes, corroborating that the model utilizes column indices as its primary navigation mechanism, while treating row positions as secondary features derived cumulatively from row boundaries (implying a reliance on semantic content for row grounding).

Figure~\ref{fig:probe_predict} illustrates the behavior of the predicted column indices and row indices. We observe a distinct \textbf{sawtooth pattern} where the predicted column index increases linearly within each row. Crucially, as the model encounters the row boundary, the predicted column index sharply resets to zero before resuming its linear ascent for the subsequent row. This indicates that the model tracks column index via a counter that resets upon detecting a row delimiter.
In contrast, the row index representation manifests as a \textbf{stepwise function}. The predicted row value remains relatively constant throughout the tokens of a single row and undergoes a discrete increment only when the sequence transitions to the next line. 
Furthermore, the predicted indices increasingly diverge from the ground truth as the row and column counts grow, indicating that the model's capacity for table understanding degrades with increasing table dimensions.

\subsection{Role of Delimiters}

The sawtooth pattern observed above implies that the model relies on specific tokens to govern its counting mechanism. We posit that the pipe delimiter (`$\vert$') serves as the fundamental trigger for this incremental counting process.

\paragraph{Probe Comparison}
To disentangle whether coordinate information is intrinsic to cell content or anchored to structural delimiters, we analyze the layer-wise evolution of probe performance.
While both delimiter and cell-based probes eventually reach comparable peak $R^2$ values in deeper layers, their trajectory differs significantly (Figure~\ref{fig:probe_r2}).
Probes trained on delimiter tokens exhibit robust linear separability ($R^2 > 0.95$) immediately in the early layers. In contrast, cell content probes show a marked latency, stabilizing at high accuracy only after Layer 15.
This performance gap in the initial 23 layers suggests a hierarchical information flow: the model first resolves coordinate states at the delimiter positions and subsequently propagates this positional information to the constituent cell tokens. Thus, the delimiter acts not merely as a separator, but as the primary synchronization signal for the model's internal coordinate system.
We also present the results of using other tabular formats with different delimiters in Appendix~\ref{subapp:formats}.

\begin{table}[t]
    \centering
    \small
    \caption{
    Effect Scores of activation patching on top 20 coordinate-encoding heads, which are selected by the $R^2$ of probes trained on cells and delimiters, respectively.
    }
    \label{tab:delimiter_patch}
    \begin{small}
    \begin{tabular}{lcc}
    \toprule
    \multirow{2}{*}{\textbf{Head Function}} & \multicolumn{2}{c}{\textbf{Probing Position}} \\
    \cmidrule(lr){2-3}
     & \textbf{Cell} & \textbf{Delimiter} \\
    \midrule
    Row Index Heads & $0.006$ & $0.527$ \\
    Column Index Heads & $0.025$ & $0.958$ \\
    \bottomrule
\end{tabular}
    \end{small}
\end{table}

\paragraph{Activation Patching}
To corroborate the hypothesis that coordinate states originate at delimiters and subsequently propagate to cells, we perform a targeted causal intervention.
We identify the top-20 attention heads based on $R^2$ of probes for row and column indices at both cell and delimiter positions.
We then construct counterfactual inputs by swapping rows or columns to shift the geometric coordinates of the target cell while keeping the query constant.
By patching the activations of these specific heads from the original to the swapped sequence at the target cell position, we measure their causal contribution via the Effect Score.

Table~\ref{tab:delimiter_patch} summarizes the results, revealing a clear causal hierarchy that mirrors our probing observations.
Interventions on heads associated with cell content yield negligible Effect Scores ($0.006$ for rows, $0.025$ for columns).
This indicates that although cell tokens eventually encode positional information (as shown by late-layer probe accuracy in Figure~\ref{fig:probe_r2}), they are downstream recipients rather than drivers of the counting mechanism.
In contrast, heads operating at delimiter positions demonstrate decisive causal control.
The column index heads at delimiters achieve a near-perfect Effect Score of $0.958$, with row heads reaching $0.527$.
These findings confirm that the delimiter serves as the causal origin of the coordinate system, actively driving the indexing process before the state is synchronized to the cell content.

We also show the experiments of noise intervention in Appendix~\ref{subapp:Noise Intervention}.
Synthesizing findings from above, we conclude that the model locate cells by simulating a discrete ordinal coordinate system.
This system is strictly ordinal: the model resolves column indices through the sequential accumulation of delimiter synchronization signals. Similarly, row indices are updated via discrete state transitions at row boundaries. Then the model propagates the tabular indices to the constituent cell tokens. This mechanism ensures that the internal representation is driven exclusively by the logical structure of the input rather than semantic length.

\section{The Linear Geometry of Tabular Indices}
    \label{sec:geometry}
In this section, we investigate the geometric properties of column index representation.
We hypothesize that the model represents column indices within a structured linear subspace governed by translation invariance and additivity. 
We show the detailed experiment setup in Appendix~\ref{subapp:geometry_details}.

\subsection{Steering with Unit Shift Vectors}
\label{sec:unit_steering}


To characterize the geometric regularity and \textbf{translation invariance} of the column representation, we derive a global unit shift vector $\vec{v}_{unit}$ by averaging the mean activation differences between adjacent column header tokens in $x_{table}$ across layers 17-23. We then perform steering interventions by modifying the target activation $\mathbf{h}$ as $\tilde{\mathbf{h}} = \mathbf{h} + k \cdot \alpha \cdot \vec{v}_{unit}$ 
where $k \in \mathbb{Z}$ is an integer representing the desired column shift and $\alpha$ is a steering coefficient.

\begin{figure}[t]
    \centering
    \includegraphics[width=.9\linewidth]{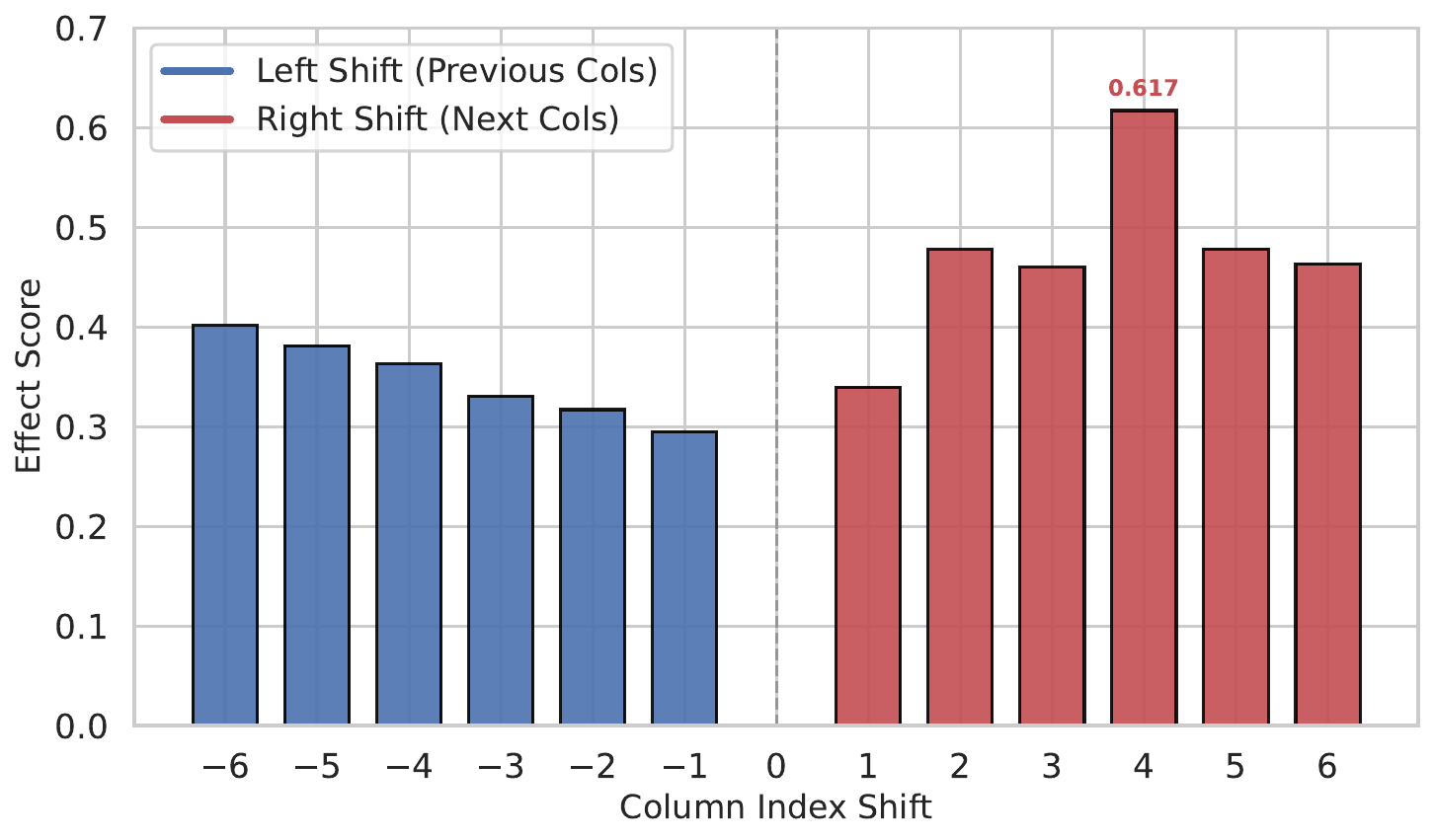}
    \caption{
      Effect Scores resulting from injecting the unit shift vector scaled by integer factors $k$. The positive integers denote rightward shifts and negative integers denote leftward shifts. 
    }
    \label{fig:single_effect}
\end{figure}

Figure~\ref{fig:single_effect} illustrates the Effect Score across different integer shifts $k$, which provides strong evidence for the linearity of the column representation. We observe that multiplying the unit shift vector by $k$ consistently steers the model to focus on the column located $k$ steps. 
This suggests that the model locates the cell by aligning its column position with that of the target column header, and encodes column positions as equidistant states along a specific direction in the activation space. 
The ability to traverse multiple columns by scaling a single vector confirms that the internal coordinate system possesses a robust linear structure.

\subsection{Compositionality of Column Vectors}
\label{sec:composition}

We further investigate the \textbf{additive compositionality} of the representation space. If the subspace is truly linear, the vector required to shift attention by $k$ columns should be decomposable: the sum of two vectors shifting by $a$ and $b$ columns should be functionally equivalent to a single shift of $a+b=k$ columns. 
We define the composite vector $\vec{v}_{comp} = \vec{v}_{a} + \vec{v}_{b}$ and compare its steering effect against the baseline vector $\vec{v}_{k}$, which is directly extracted from columns separated by distance $k$.
We conduct experiments for various target offsets $k$ using multiple combinations of constituent shifts $a$ and $b$ drawn from the range $[-6, 6]$.
We evaluate the interventions using the Effect Score.

\begin{figure}[t]
    \centering
    \includegraphics[width=.9\linewidth]{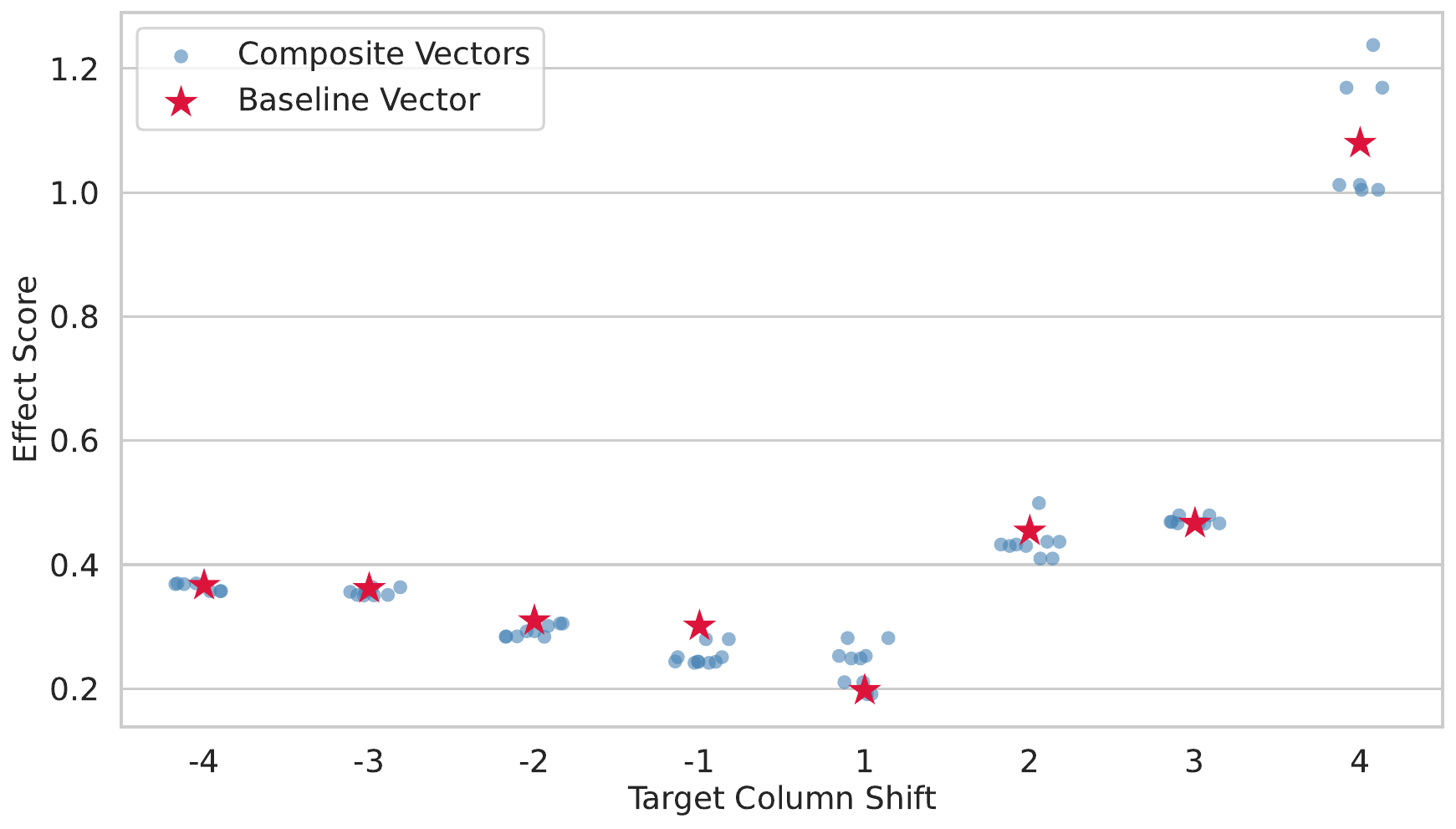}
    \caption{
      Comparison of Effect Scores between Baseline Vectors (red stars) and Composite Vectors (blue dots). The Baseline Vector is the shift vector directly extracted for target shift $k$. The Composite Vectors are formed by summing $\vec{v}_{a} + \vec{v}_{b}$ such that $a + b = k$.
    }
    \label{fig:comb_effect}
\end{figure}

Figure~\ref{fig:comb_effect} shows the results of baseline and composite vectors, with details in Appendix~\ref{subapp:geometry_details}. The alignment between the red baseline stars and the blue composite dots indicates a high degree of linearity. For a target shift of $k=4$ the baseline vector achieves an Effect Score of $1.079$, and the composite vectors formed by linear combinations such as $1+3$ yield comparable Effect Scores of $1.169$. This pattern holds for negative shifts as well where a target of $k=-4$ with a baseline score of $0.367$ is closely matched by the combination of $-2+(-2)$ yielding $0.357$. 
The results confirm that the column indices form a linear subspace where vector addition preserves the relative spatial geometry.

\section{Generalization to Multi-Cell Location}
    \label{sec:complex}
    \subsection{Experimental Setup}
To evaluate the generalization of the internal pipeline, we extend the atomic cell location task to multi-cell location scenarios.
We construct two distinct benchmarks each consisting of $500$ samples (details in Appendix~\ref{app:data}).
These benchmarks retain the same tables with the benchmark of the atomic cell location task, with the modifications applied solely to the natural language questions.
In the \textit{Multi-Row} setting the query $x_{query}$ explicitly references a subset of target row headers $\mathcal{R}_{sub} \subset \mathcal{R}$ and a single target column header $c \in \mathcal{C}$.
The objective is to locate the set of values $\{v_{r,c} \mid r \in \mathcal{R}_{sub}\}$ corresponding to the intersection of the specified rows and the target column.
In the \textit{Multi-Column} setting the query specifies a single row header $r \in \mathcal{R}$ and a subset of target column headers $\mathcal{C}_{sub} \subset \mathcal{C}$ requiring the location of values $\{v_{r,c} \mid c \in \mathcal{C}_{sub}\}$.
This design allows us to investigate whether the model resolves multiple constraints through sequential processing or by parallelizing the semantic binding and coordinate localization mechanisms.

\begin{figure}[t]
    \centering
    \begin{subfigure}[b]{0.55\linewidth} 
        \centering
        \includegraphics[width=\linewidth]{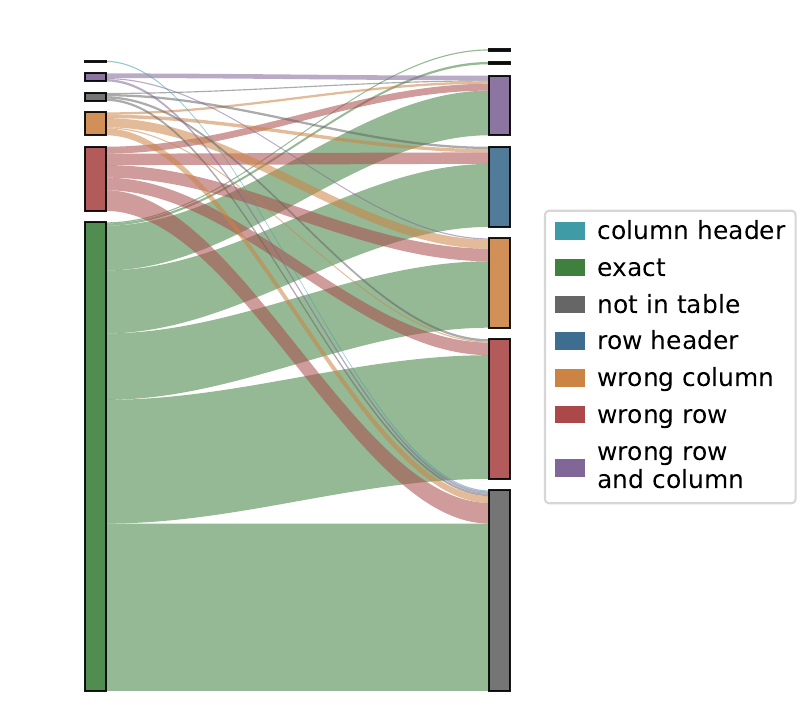}
        \caption{Error distribution post-ablation.}
        \label{fig:multi_sankey}
    \end{subfigure}
    \hfill
    \begin{subfigure}[b]{0.4\linewidth}
        \centering
        \includegraphics[width=\linewidth]{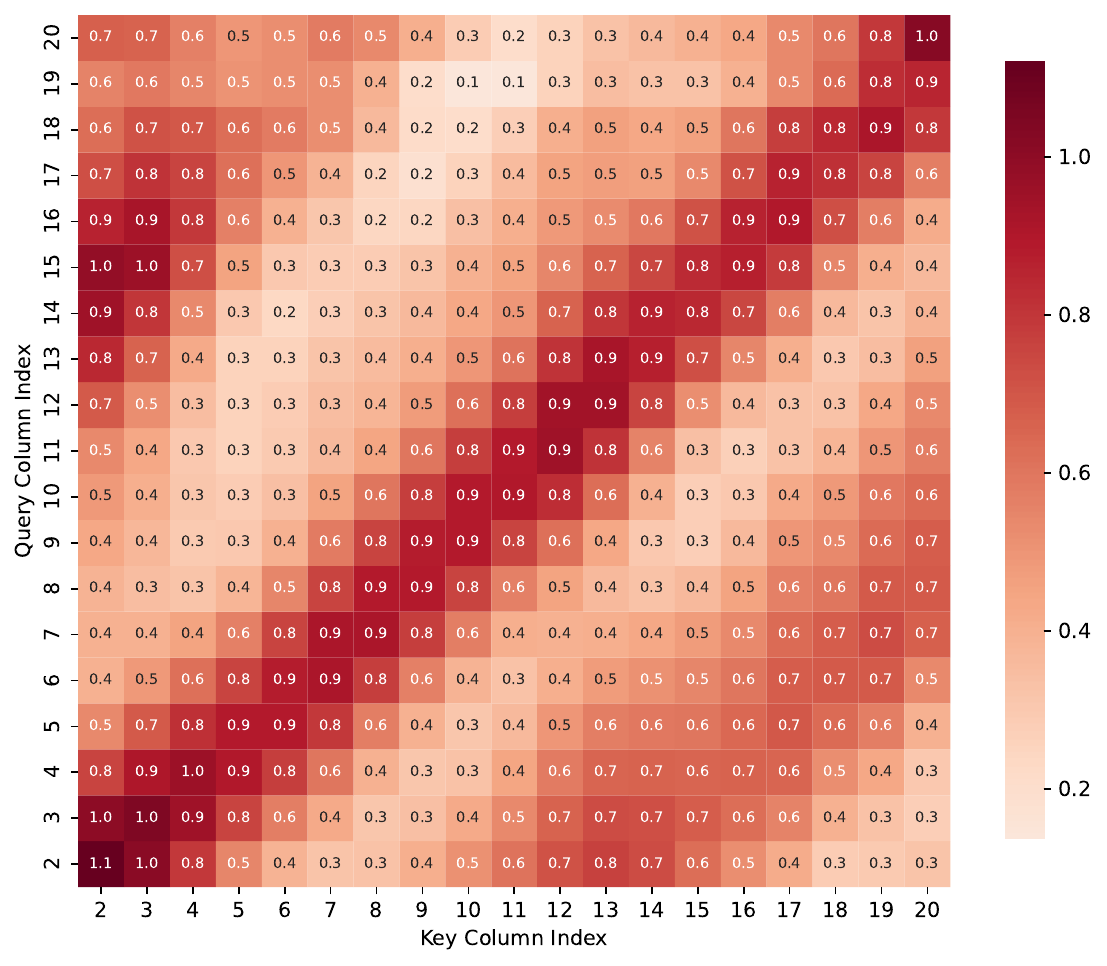}
    \caption{Multi-threaded coordinate localization.}
    \label{fig:multi_heatmap}
    \end{subfigure}
    
    \caption{
        Mechanistic analysis on the multi-cell location task. 
        \textbf{(a)} illustrates the impact of ablating the row alignment heads identified in Stage I on multi-row queries. 
        \textbf{(b)} depicts the Query-Key interaction scores for the top-20 coordinate-encoding heads from Stage II on multi-column queries. 
    }
    \label{fig:multi_mechanisms}
\end{figure}

\subsection{Parallel Semantic Binding}

We investigate whether the semantic binding mechanism identified in the atomic cell location task generalizes to multi-row scenarios. We hypothesize that the model reuses the specifically identified \textit{row alignment heads} from Stage I (\S\ref{subsec:stage 1}) to process multiple constraints in parallel rather than instantiating distinct mechanisms. 
Specifically, this parallelism is inherent to the self-attention mechanism, where these heads compute binding scores for all query constraints concurrently across the sequence dimension.
To verify this we perform an ablation study where we zero-ablate the top-20 row alignment heads, which are discovered in the atomic cell location task, during the processing of multi-row queries and observe the changes of model outputs. 

Figure~\ref{fig:multi_sankey} visualizes the distribution of error types following the ablation. The Sankey diagram reveals a catastrophic collapse in performance ($81.8\% \to 0.4\%$). Specifically, $38.4\%$ of samples shifting from \texttt{exact} matches to the \texttt{wrong row} or \texttt{row header} (directly outputting the row headers in the query) error categories. 
This specific failure mode mirrors the degradation observed in the atomic cell location setting, confirming that the model relies on the identical set of functional components to resolve multiple semantic targets. This demonstrates that the model does not resort to an iterative or multi-step reasoning strategy; instead, it possesses the capacity for parallel execution, where the row alignment operator is instantiated across representation sites within the latency of a single forward pass.

\subsection{Multi-Threaded Coordinate Localization}

We subsequently examine the coordinate localization mechanism for multi-column queries. Crucially, we utilize the top-20 \textit{coordinate-encoding heads} previously identified in the atomic cell location analysis (Stage II in \S\ref{subsec:stage 2}) to explore the model's behavior in multi-column settings. We apply the RoPE-based interaction metric to the heads to compute the geometric interaction scores between the query column representations and the key vectors of the table columns. 

Figure~\ref{fig:multi_heatmap} displays the average interaction scores for these specific top-20 coordinate-encoding heads. The heatmap reveals that the attention patterns of these heads exhibit strong positive interaction scores on the indices that correspond to all queried columns simultaneously. The preservation of the diagonal alignment pattern across multiple targets confirms that the coordinate-encoding heads are not limited to single-state tracking. This implies that the linear coordinate system operates as a superposition-ready vector space, where the same set of attention heads can calculate geometric offsets for distinct targets in parallel and locate the corresponding values in a single forward pass.


\section{Conclusion}
In this work, we provide a mechanistic dissection of table understanding in LLMs via the cell location task. Through activation patching, linear probing, and ablation study, we formalize this process into three stages: Semantic Binding, Coordinate Localization, and Information Extraction. We reveal that models navigate structures by maintaining an implicit coordinate system driven by delimiter counting, with column indices represented in a linear subspace navigable via vector arithmetic. Finally, we show that models generalize to complex constraints by multiplexing these atomic mechanisms in parallel. Our findings bridge the gap between sequential processing and structural logic, offering a robust framework for improving the reliability and interpretability of LLMs in structured data applications.

\clearpage
\section*{Impact Statement}
All datasets and models used in this paper are publicly available, and our usage follows their licenses and terms.
We employ AI tools for coding and writing polishing.

This paper presents the work whose goal is to advance the field of machine learning through a mechanistic interpretability analysis of table understanding in Large Language Models. By identifying the specific attention heads and internal representations responsible for cell location, we contribute to the broader effort of making AI systems more transparent and predictable. Such insights are foundational for building trustworthy AI that handles structured information across various domains. 






\bibliography{example_paper}
\bibliographystyle{icml2026}

\clearpage
\appendix
\section{Related Works}
    \label{sec:related}
    \subsection{Mechanistic Interpretability of Reasoning}
Recent advancements in mechanistic interpretability have extensively mapped the internal circuits responsible for logical and linguistic reasoning \cite{gan2026mechanism-survey}. 
Researchers reverse-engineer the algorithmic components of specific tasks such as indirect object identification where models employ specialized attention heads to move information between token positions \cite{wang2023interpretability-wild,zhang2024best-pratices}. 
Subsequent studies extend this analysis to propositional logic demonstrating that models utilize a modular pipeline of locating moving and processing heads to resolve implications \cite{hong2025a-implies-b}. 
Furthermore investigations into implicit multi-hop reasoning reveal that Transformers can learn to separate memorization from generalization often exhibiting distinct developmental stages \cite{wang2024grokking,ye2025how-transformers-implicit}. 
Also, research on Chain-of-Thought processes demonstrates that models utilize intermediate tokens as a computational scratchpad to decompose global logic into iterative state updates across layers \cite{wang2023understanding-cot,zhang2023cot-policy}.
Most recently, analysis reveals that Text-to-SQL generation operates via a strictly staged mechanism: the model first determines the abstract syntactic structure and subsequently fills in specific schema entities \cite{harrasse-etal-2025-tinysql}.

However, prior works primarily investigate tasks governed by explicit logical rules or generative syntax.
In contrast our work investigates the mechanism of table understanding where the model must resolve spatial constraints rather than purely semantic dependencies. 
We show that table understanding necessitates a distinct Coordinate Localization stage, where the model navigates the serialized grid structure via delimiter counting instead of semantic signals.

\subsection{Internal Representations of Structure}
A parallel line of research explores how LLMs encode geometric or structural knowledge within their internal representations. 
The existing work on synthetic tasks such as Othello proves that sequence models can spontaneously emerge internal world representations allowing them to track complex game states without explicit supervision \cite{li2023World-Representations}. 
Regarding sequential structure researchers identify successor heads that utilize linear subspaces to represent ordered sequences confirming that models possess linear arithmetic capabilities for progression \cite{gould2024successor-heads}. 
Additional studies on entity binding propose that models assign abstract binding identifiers to link entities with their attributes independent of their absolute positions \cite{feng2024bind-entities,yang2025emergent-symbolic}.
Furthermore, recent findings demonstrate that for many factual and linguistic relations, transformer language models implement a linear decoding strategy where an affine transformation on the intermediate subject representation directly maps to the corresponding object \cite{hernandez2024linearity}.

While these works establish that models can internalize static world models or sequential progressions, our research uncovers a dynamic mechanism for understanding tables. 
Unlike fixed game boards or sequential entities, tables require an internal coordinate system. 
We distinguish our contribution by proving that this structural navigation is governed by an implicit linear geometry constructed via delimiter counting, which allows the model to generalize coordinate localization to multi-target queries through vector arithmetic in the residual stream.

\subsection{LLMs for Table Understanding}
Since the integration of LLMs into table-related tasks, a burgeoning body of research has emerged to evaluate and enhance the tabular understanding capabilities of these models \cite{fang2024tabular-survey}. 
Regarding evaluation, datasets such as MMTU~\cite{xing2025mmtu} and ENTRANT~\cite{elias2024entrant} introduce multi-task and domain-specific datasets in the financial sector, revealing that current models still exhibit significant deficiencies in comprehensive table comprehension. 
Furthermore, existing studies indicate that LLMs lack robustness in this area, as modifications to the organizational order of contents or structural formats frequently change performance \cite{singha2023tabular-representation,sui2024table-meets-llm,liu-etal-2024-rethinking}. 
Efforts to improve tabular understanding are generally categorized into two paradigms. 
The first encompasses training-free methods which optimize prompts by decomposing complex questions and tables \cite{ye2023dater}, refining data serialization formats \cite{yang2025triples,zhang2024flextaf}, or integrating tools to facilitate reasoning \cite{wang2024chainoftable}. 
The second paradigm focuses on enhancing intrinsic capabilities through training, which involves designing specialized model architectures \cite{he-etal-2025-tablelora,Jin_25hegta} or conducting supervised fine-tuning and reinforcement learning on large-scale tabular corpora \cite{liu2025hippo,wu2025tabler1}.

Despite these advancements in evaluating and enhancing performance, the underlying mechanisms governing how LLMs understand tables remain a black box. Our work fills this gap by shifting the focus from external performance metrics to internal mechanism analysis. We focus on the atomic \textit{cell location} task and provide the first formal characterization of its internal stages: Semantic Binding, Coordinate Localization, and Information Extraction. By uncovering that models navigate linearized tables via delimiter-counting mechanisms and encode coordinates within a linear subspace, we offer a fundamental shift in perspective—from observing that models understand tables to explaining \textit{how} they bridge the gap between sequential input and two-dimensional logic.

\section{Data Details}
\label{app:data}

\begin{table*}[t]
\centering
\small
\caption{Statistics of the synthetic datasets for Table Understanding tasks. The table details the range of table dimensions and the scope of the query targets (number of rows or columns requested) for each experimental setting.}
\label{tab:data_stats}
\begin{tabular}{lcccc}
\toprule
\textbf{Task Type} & \textbf{Table Rows ($R$)} & \textbf{Table Cols ($C$)} & \textbf{Query Rows} & \textbf{Query Cols} \\
\midrule
Atomic Cell Location & $[4, 20]$ & $[4, 20]$ & $1$ & $1$ \\
Multi-Row Location   & $[4, 20]$ & $[4, 20]$ & $|\mathcal{R}_{sub}| \in [2, 11]$ & $1$ \\
Multi-Column Location & $[4, 20]$ & $[4, 20]$ & $1$ & $|\mathcal{C}_{sub}| \in [2, 11]$ \\
\bottomrule
\end{tabular}
\end{table*}

\begin{figure*}[t]
    \input{fig/example_single}
\end{figure*}

To rigorously investigate the internal mechanics of table understanding, we synthesize datasets specifically designed for the cell location task. We select cell location as the primary focus because it constitutes an atomic operation requiring the model to not only perform semantic alignment—matching query terms to row and column headers—but also to execute structural reasoning by navigating the two-dimensional coordinates of the table.
Also, it is a common and fundamental subproblem of more complex table-related reasoning problems in the wild, such as the Table Question-Answering task \cite{pasupat-liang-2015-wikitq,wu2025tablebench}.

We deliberately avoid utilizing existing datasets to preclude the risk of data contamination \cite{cheng2025survey-data-contamination,oren2024proving-contamination}. By constructing synthetic tables with randomized values, we ensure that the model cannot rely on parametric knowledge or memorization to answer queries \cite{wu2025memorization-contamination}. Instead, the model is forced to extract answers solely via in-context reasoning over the provided $x_{table}$. We describe the data generation process below, with dataset statistics summarized in Table~\ref{tab:data_stats}.

\subsection{Atomic Cell Location}

\paragraph{Table Generation}
We first construct a comprehensive entity pool comprising $83$ distinct semantic categories. To ensure value uniqueness, the sets of values across different categories are disjoint.
For the table structure, the primary key (the header of the first column $c_1$) is restricted to generic identifiers including \textit{name}, \textit{role}, and \textit{id}. The remaining column headers $\mathcal{C} \setminus \{c_1\}$ are randomly sampled from the entity pool categories.
The dimensions of the tables are randomized, with the number of rows $R$ and columns $C$ uniformly sampled from the interval $[4, 20]$. 
This upper bound $20$ is selected to ensure that the linearized table sequences remain within the model's context window while maintaining a challenging complexity profile, following the principals of the previous work \cite{hong2025a-implies-b}. Critically, this scale requires the model to perform robust two-dimensional coordinate mapping within a linear sequence without inducing performance collapse, thereby facilitating a granular investigation into the internal mechanisms of cell location.
We generate a total of $500$ independent tables.

\paragraph{Query Formulation}
For each synthesized table, we configure a corresponding query $x_{query}$. We randomly sample a target row header $r \in \mathcal{R}$ and a target column header $c \in \mathcal{C}$. The natural language question is generated using one of several templates, such as \textit{What is the c for r?} or \textit{According to the table, what is the value of c for r?}.
To clearly define the task and the expected output format, a one-shot demonstration $D$ is prepended to every input. Example~\ref{ex:single} illustrates a representative input sample.

\subsection{Multi-Cell Location}

To probe the model's ability to generalize to more complex structural constraints, we extend the benchmark while maintaining the same base table corpus ($500$ samples) used in the atomic cell location task. The queries are categorized into two distinct types:

\begin{itemize}
    \item \textbf{Multi-Row Location:} For a fixed column header $c$, we randomly select a subset of row headers $\mathcal{R}_{sub} \subset \mathcal{R}$ to form the query, requiring the model to locate multiple values in a vertical structure.
    \item \textbf{Multi-Column Location:} For a fixed row header $r$, we randomly select a subset of column headers $\mathcal{C}_{sub} \subset \mathcal{C}$, requiring the location of values across a horizontal structure.
\end{itemize}

Consistent with the atomic-cell setting, specific one-shot demonstrations are provided for each task type to guide the generation of list-formatted answers.

\section{Experiment Details and Supplement}
\label{app:experiment details}

\subsection{Details of Activation Patching}
\label{subapp:patch_details}

In this subsection, we provide the specific implementation details of the activation patching experiments and further analyze the results of Row Patching to corroborate the findings presented in \S\ref{subsec:patch_overview}.

\paragraph{Experimental Setup}
To achieve high-resolution localization of information flow, we employ a fine-grained tokenization strategy rather than averaging over broad token types. 
The input sequence is partitioned into $15$ distinct regions. The underlying segmentation principle is to isolate the semantic focus tokens, specifically the query constraints, their corresponding table headers, and the target cell values, while aggregating the surrounding non-focal text into context segments. This approach allows us to distinguish the causal role of structural coordinates from general context.

The coordinates on the x-axis (Indices 1-15) in Figure~\ref{fig:col_patch} correspond to the following token spans:

\begin{itemize}
    \item \textbf{Index 1:} The demonstration and initial table content preceding the relevant column headers.
    \item \textbf{Index 2:} The corrupt column header $\mathcal{I}_{t}(h^\prime)$ in the table schema definition.
    \item \textbf{Index 3:} The tokens between $\mathcal{I}_{t}(c^\prime)$ and $\mathcal{I}_{t}(c)$.
    \item \textbf{Index 4:} The relevant column header in the table schema, corresponding to the target column $\mathcal{I}_{t}(c)$.
    \item \textbf{Index 5:} The intermediate table content, including data rows positioned between the header definition and the target row.
    \item \textbf{Index 6:} The row header of the target row $\mathcal{I}_{t}(r)$, which serves as the primary vertical anchor.
    \item \textbf{Index 7:} The delimiter token separating the row header from cell values.
    \item \textbf{Index 8:} The corrupt cell value $\mathcal{I}_{t}(v_{r,c^\prime})$ within the target row.
    \item \textbf{Index 9:} The tokens between $\mathcal{I}_{t}(v_{r,c^\prime})$ and $\mathcal{I}_{t}(v_{r,c})$.
    \item \textbf{Index 10:} The target cell value $\mathcal{I}_{t}(v_{r,c})$ situated at the intersection of the specified row and column.
    \item \textbf{Index 11:} The remaining table content following the target row and the preamble query.
    \item \textbf{Index 12:} The column constraint token within the natural language query $\mathcal{I}_{q}(c)$.
    \item \textbf{Index 13:} The syntactic bridge tokens (e.g., prepositions) within the query between $\mathcal{I}_{q}(c)$ and $\mathcal{I}_{q}(r)$.
    \item \textbf{Index 14:} The row constraint token within the natural language query $\mathcal{I}_{q}(r)$.
    \item \textbf{Index 15:} The query suffix and answer prefix tokens.
\end{itemize}

We calculate the Effect Score at each layer and token span index. An Effect Score close or beyond $1.00$ indicates that the patch fully recovers the correct logit difference.

\paragraph{Row Patching Analysis}

\begin{figure}[t]
    \centering
    \includegraphics[width=.95\linewidth]{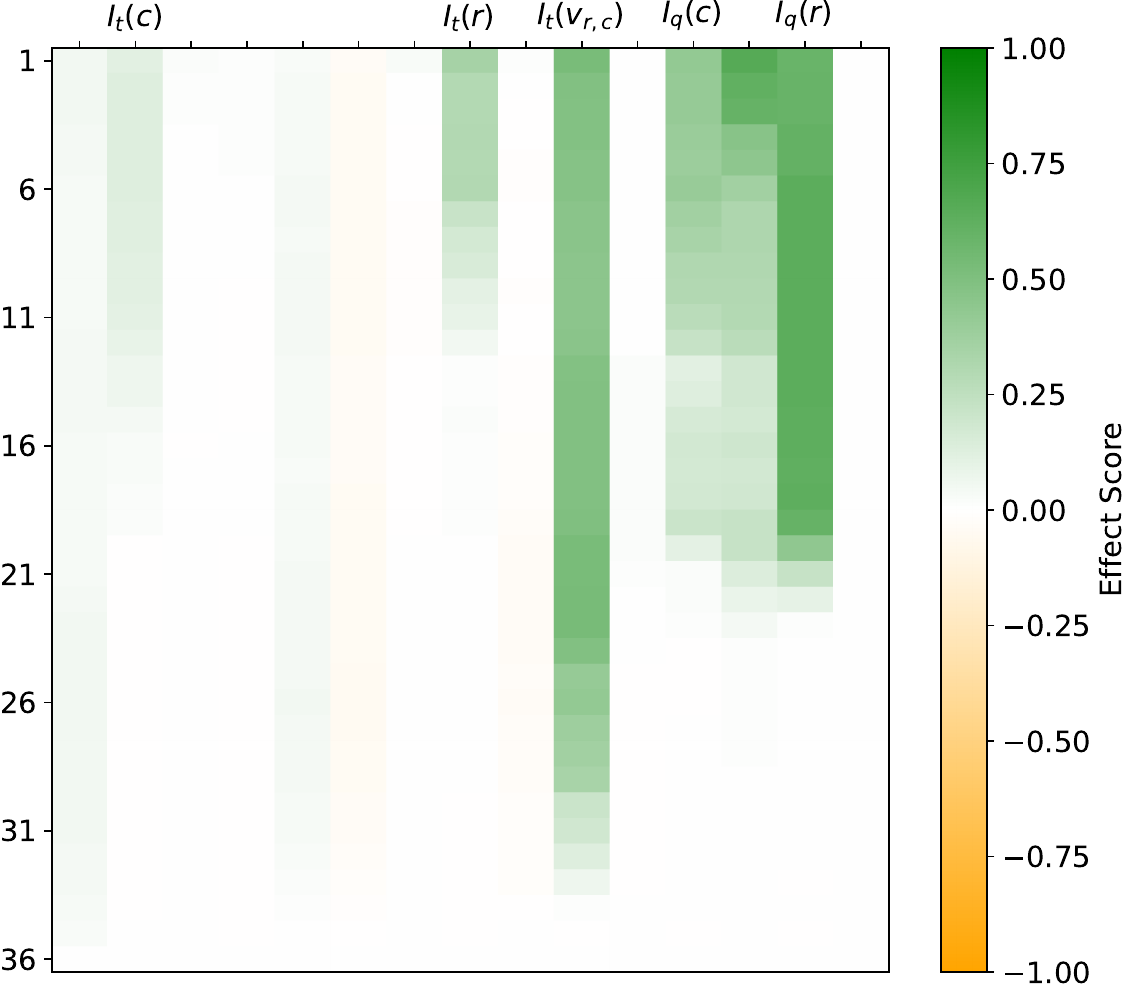}
    \caption{
      Average Row Patching Effect Score across layers and token positions. The heatmap visualizes the causal importance of specific token regions when the row constraint in the query is corrupted. 
    }
    \label{fig:row_patch}
\end{figure}

In the main text, we discuss the Three-Stage mechanism primarily through the lens of Column Patching. Here, we demonstrate that Row Patching (corrupting the query row constraint to shift the target to a different row) exhibits an identical mechanistic pattern, as shown in Figure~\ref{fig:row_patch}.

\begin{enumerate}
    \item \textbf{Stage 1: Semantic Binding (Layers 1-16).} 
    Consistent with column patching, the early layers exhibit high Effect Scores concentrated on the query constraints and their corresponding schema matches. Specifically, we observe strong activation on the Query Row $\mathcal{I}_{q}(r)$ (Index 14) and the corresponding Table Row Header $\mathcal{I}_{t}(r)$ (Index 6). This confirms that the model initiates the reasoning process by binding the semantic constraint from the question to the specific row identifier in the table structure.
    
    \item \textbf{Stage 2: Coordinate Localization (Layers 17-23).}
    In the middle layers, the causal mass shifts decisively rightward from the row header to the cell values. Notably, Index 10, which corresponds to the target value $\mathcal{I}_{t}(v_{r,c})$, displays a distinct band of high importance. This validates the Coordinate Localization stage, where the model, having anchored the correct row $r$ and column $c$, converges on the specific intersection coordinate.
    
    \item \textbf{Stage 3: Information Propagation (Layers 24+).}
    In the final layers, the heatmap fades to neutral values across all input tokens. This indicates that the extract ion is complete; the target information $v_{r,c}$ has been successfully encoded into the residual stream of the final token, and the generation process no longer relies on direct attention to the input sequence.
\end{enumerate}

These results reinforce our conclusion that table understanding in LLMs operates via a structured, sequential process of constraint matching followed by intersection navigation, independent of whether the location requires row-centric or column-centric reasoning.

\subsection{Ablation Study in Stage I}
\label{subapp:ablation}

In this subsection, we detail the experimental setup for identifying alignment heads and provide a granular error analysis of the ablation studies discussed in \S\ref{subsec:stage 1}.

\paragraph{Identification of Alignment Heads}
To pinpoint the attention heads responsible for Semantic Binding, we define an \textit{Alignment Score} that quantifies the preferential attention flow from query constraints to their corresponding table headers. 
Let $x$ be the input sequence. We denote the set of token indices corresponding to the row or column constraint in the query as $\mathcal{I}_{q} \in \{\mathcal{I}_{q}(r), \mathcal{I}_{q}(c)\}$. Similarly, let $\mathcal{I}_{tgt} \in \{\mathcal{I}_{t}(r), \mathcal{I}_{t}(c)\}$ denote the token indices of the ground-truth header in the table $x_{table}$, and $\mathcal{I}_{dist}$ denote the indices of all other same-type headers (distractors).
For a given attention head $h$ at layer $l$, we calculate the Alignment Score $S^{(l,h)}$ as the difference between the attention mass focused on the target and the average attention mass focused on distractors:

\begin{equation}
\begin{aligned}
S^{(l,h)} = \frac{1}{|\mathcal{I}_{q}|} \sum_{i \in \mathcal{I}_{q}} \Biggl( 
& \sum_{j \in \mathcal{I}_{tgt}} Att^{(l,h)}_{i,j} - \\ 
& \frac{1}{|\mathcal{I}_{dist}|} \sum_{k \in \mathcal{I}_{dist}} Att^{(l,h)}_{i,k} \Biggr)
\end{aligned}
\end{equation}

where $Att^{(l,h)}_{i,j}$ represents the attention weight from token $i$ to token $j$. We compute the global score $\bar{S}^{(l,h)}$ by averaging over $N=500$ samples. We identify the top-20 heads with the highest $\bar{S}^{(l,h)}$ for rows and columns respectively as \textit{row alignment heads} and \textit{column alignment heads}.

\paragraph{Error Categorization}
To rigorously assess the impact of ablation, we classify the model's output $y_{pred}$ into seven distinct categories relative to the ground truth value $v_{r,c}$:

\begin{itemize}
    \item \textbf{Exact:} The model correctly generates the target cell value ($y_{pred} = v_{r,c}$).
    \item \textbf{Wrong Row:} The output corresponds to a cell in the correct column but a different row ($y_{pred} = v_{r',c}$ where $r' \neq r$). This error indicates a specific failure in Row Semantic Binding.
    \item \textbf{Wrong Column:} The output corresponds to a cell in the correct row but a different column ($y_{pred} = v_{r,c'}$ where $c' \neq c$). This indicates a failure in Column Localization.
    \item \textbf{Wrong Row and Column:} The output is a value from the table but matches neither the target row nor column ($y_{pred} = v_{r',c'}$).
    \item \textbf{Row Header:} The model hallucinates the row header constraint and output it directly instead of a value ($y_{pred} = r$).
    \item \textbf{Column Header:} The model repeats the column header constraint ($y_{pred} = c$).
    \item \textbf{Not in Table:} The output is not found within the table context $x_{table}$.
\end{itemize}

\paragraph{Result Analysis}

\begin{figure}[h]
    \centering
    \begin{subfigure}[b]{0.48\linewidth}
        \centering
        \includegraphics[width=\linewidth]{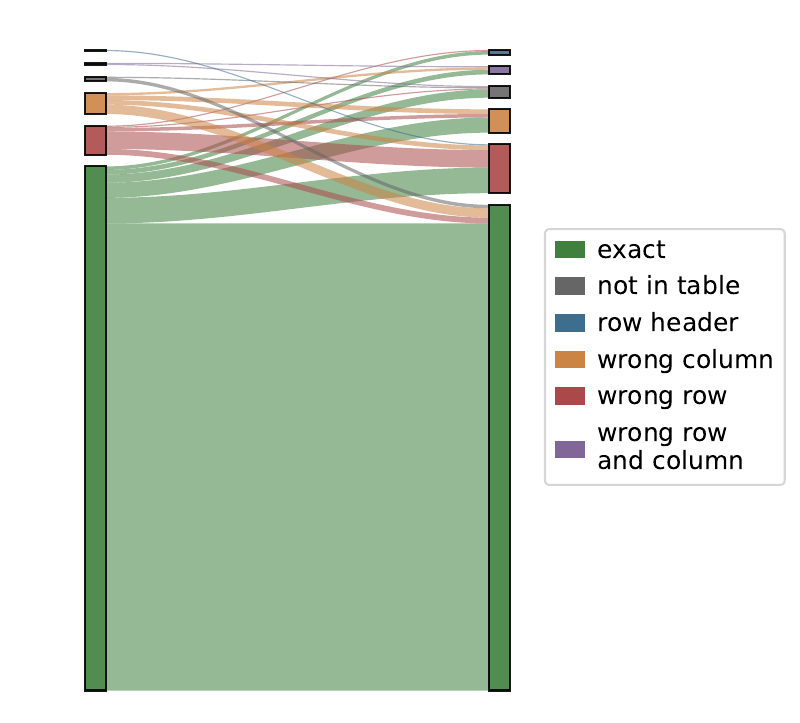}
        \caption{Column alignment heads.}
        \label{fig:sankey_col}
    \end{subfigure}
    \hfill
    \begin{subfigure}[b]{0.48\linewidth}
        \centering
        \includegraphics[width=\linewidth]{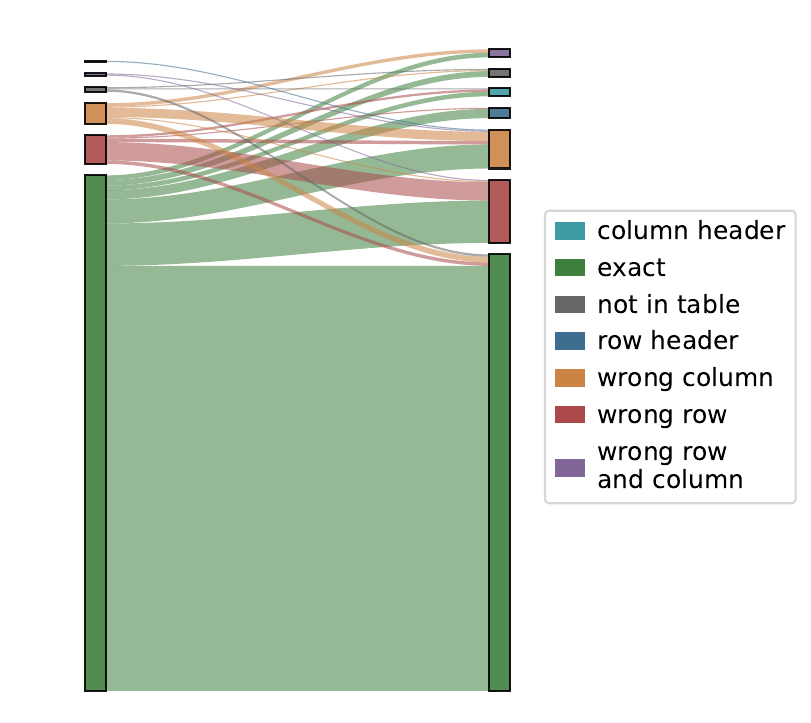}
        \caption{Random heads.}
        \label{fig:sankey_random}
    \end{subfigure}
    \caption{Sankey diagrams illustrating the flow of error types under different ablation conditions. Compared with Figure~\ref{fig:sankey_row} shows catastrophic failure for row ablation, \textbf{(a)} demonstrates that ablating column alignment heads maintains high performance, similar to the \textbf{(b)} random baseline.}
    \label{fig:ablation_comparison}
\end{figure}

We compare the ablation of row alignment with column alignment heads and random heads (Figure \ref{fig:ablation_comparison}). 
As established in \S\ref{subsec:stage 1}, zero-ablating row alignment heads caused a catastrophic drop in performance (Exact match falling to $5.8\%$), with the vast majority of errors shifting to the \texttt{wrong row} category. This confirms that specific attention heads are causally necessary for binding row constraints.
In contrast, Figure \ref{fig:sankey_col} reveals that ablating the top column alignment heads does not lead to a significant performance degradation. The dominant flow remains directed toward the \texttt{exact} category, comparable to the random baseline shown in Figure \ref{fig:sankey_random}. Specifically, the incidence of \texttt{wrong column} errors does not increase disproportionately.
This disparity highlights a fundamental mechanistic asymmetry in table understanding.
\textit{Row Binding is Explicit:} The model relies on sparse, dedicated attention heads in early layers to point to the correct row header based on semantic similarity. Removing these pointers breaks the location mechanism.
\textit{Column Binding is Implicit or Distributed:} The model does not rely on strong semantic attention weights to bind column headers. The resilience of column location to attention ablation suggests that column information is likely processed via the geometric mechanisms described in Stage II (Coordinate Localization), where position indices and RoPE interactions take precedence over direct semantic attention.

\subsection{Details of Linear Probing}
\label{subapp:probing_details}

In this subsection, we provide further details regarding the optimization objective and training setup for the linear probes described in \S\ref{sec:coordinate}.

\paragraph{Optimization Objective}
To mitigate the effects of multicollinearity inherent in high-dimensional neural representations and to improve generalization, we employ Ridge Regression (linear least squares with $L_2$ regularization). For a specific layer $l$, the parameters $\mathbf{W}_p$ and $\mathbf{b}_p$ are obtained by minimizing the following regularized loss function:

\begin{equation}
    \mathcal{L}(\mathbf{W}_p, \mathbf{b}_p) = \sum_{i=1}^{N} \left( z_i - (\mathbf{W}_p \mathbf{h}_{i}^{(l)} + \mathbf{b}_p) \right)^2 + \lambda \|\mathbf{W}_p\|_2^2
\end{equation}

where:
\begin{itemize}
    \item $N$ is the total number of tokens used for training the probe.
    \item $\mathbf{h}_{i}^{(l)} \in \mathbb{R}^d$ is the activation vector corresponding to the $i$-th token in the training set.
    \item $z_i \in \mathbb{R}$ is the ground-truth coordinate (normalized) for the $i$-th token.
    \item $\lambda$ is the regularization strength hyperparameter (set to $\lambda = 1.0$).
\end{itemize}

\paragraph{Coordinate Normalization}
Since tables in our dataset vary significantly in size, using absolute indices would introduce scale variance. We normalize the geometric coordinates to the unit interval $[0, 1]$. 
Let $R_{max}$ and $C_{max}$ denote the total number of rows and columns in the table containing the $i$-th token. The target variable $z$ is defined as:
\begin{equation}
    z_{row}^{(i)} = \frac{r_{idx}^{(i)}}{R_{max} - 1}, \quad z_{col}^{(i)} = \frac{c_{idx}^{(i)}}{C_{max} - 1}
\end{equation}
This normalization ensures that the probe learns relative geometric positions rather than memorizing specific integer indices.

\paragraph{Training Setup}
We construct the training dataset by extracting internal representations of specific tokens that correspond to structural markers across the corpus. 
For each layer $l$, this yields a dataset of $N$ token-coordinate pairs $\mathcal{D}_l = \{(\mathbf{h}_{i}^{(l)}, z_i)\}_{i=1}^N$. 
The dataset is randomly partitioned into a training set ($80\%$) and a held-out evaluation set ($20\%$). The $R^2$ scores are reported on the evaluation set.

\begin{figure*}[t]
    \centering
    \begin{subfigure}[b]{0.48\textwidth} 
        \centering
        \includegraphics[width=\linewidth]{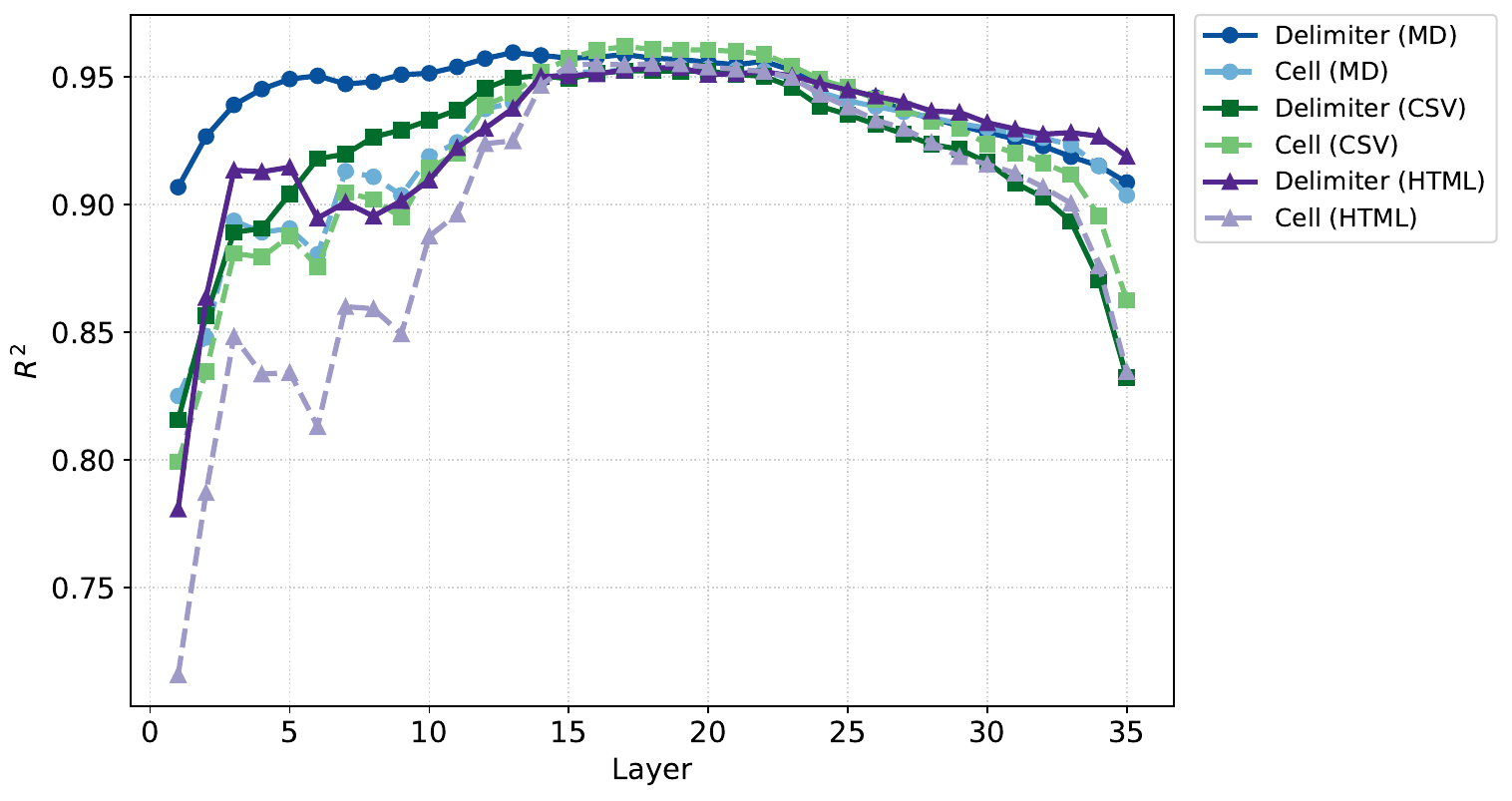}
        \caption{Column Probe Performance ($R^2$)}
        \label{fig:probe_col_formats}
    \end{subfigure}
    \hfill
    \begin{subfigure}[b]{0.48\textwidth}
        \centering
        \includegraphics[width=\linewidth]{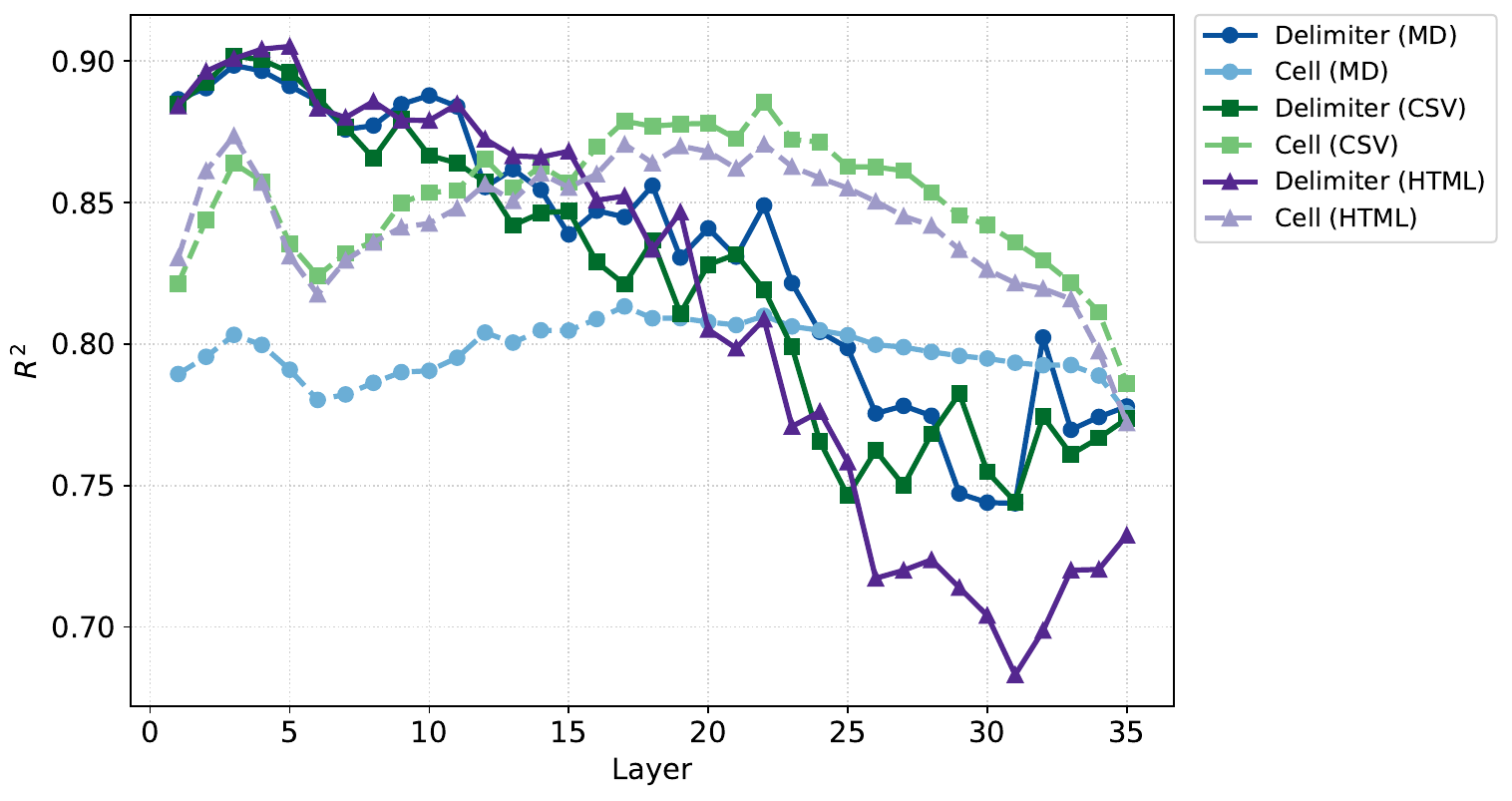}
        \caption{Row Probe Performance ($R^2$)}
        \label{fig:probe_row_formats}
    \end{subfigure}
    \caption{
        Coefficient of determination ($R^2$) for linear probes trained on varying tabular formats (MD, CSV, HTML). 
        Solid lines represent probes trained on delimiter tokens corresponding to formats, while dashed lines represent probes trained on cell content.
    }
    \label{fig:probe_formats}
\end{figure*}

\subsection{Generalization across Tabular Formats}
\label{subapp:formats}

To ensure that the implicit coordinate system observed in \S\ref{sec:coordinate} is not an artifact specific to the Markdown format, we extend our probing analysis to two additional common tabular serialization formats: Comma-Separated Values (CSV) and HyperText Markup Language (HTML).

\paragraph{Experimental Setup}
We adhere to the same data generation protocol described in \S\ref{sec:coordinate}, varying only the serialization function $\phi(T)$.
For a table $T$ with columns $\mathcal{C}$ and rows $\mathcal{R}$:
\begin{itemize}
    \item \textbf{Markdown (MD):} Uses the pipe character (`$\vert$') as the cell delimiter and newlines for row boundaries.
    \item \textbf{CSV:} Uses the comma (`,') as the cell delimiter and newlines for row boundaries. 
    \item \textbf{HTML:} Uses hierarchical tags, where \texttt{\textless td\textgreater} and \texttt{\textless /td\textgreater} delineate cells, and \texttt{\textless tr\textgreater} and \texttt{\textless /tr\textgreater} delineates rows.
\end{itemize}
We train linear probes on the intermediate activations of these sequences to predict column indices ($c_{idx}$) and row indices ($r_{idx}$).

\paragraph{Results and Analysis}
Figure~\ref{fig:probe_formats} presents the layer-wise $R^2$ scores for column and row probes across the three formats.
As shown in Figure~\ref{fig:probe_col_formats}, the model successfully constructs linear representations of column indices across all formats, achieving peak $R^2 > 0.90$ in the middle layers (Layers 17-23).
Crucially, the hierarchical lag between delimiter probes (solid lines) and cell probes (dashed lines) is preserved regardless of the format. 
For instance, in HTML, the probe on the \texttt{\textless /td\textgreater} tag consistently achieves high accuracy layers before the probe on the enclosed cell content. 
This confirms that the delimiter counting mechanism proposed in \S\ref{sec:coordinate} is format-invariant: the model identifies the ordinal significant separator—whether it is a pipe, a comma, or a closing tag—and utilizes it as a synchronization signal to update its internal column counter.


\begin{table}[t]
\centering
\small
\caption{
Accuracy comparison of Qwen3-4B under different noise settings. 
``Baseline'' denotes performance on clean inputs without noise injection. 
``Noise +'' and ``+ Noise'' indicate noise insertion preceding and succeeding the target column within the target row, respectively. 
The highest scores are highlighted in \textbf{bold}.
}
\label{tab:nosie}


\begin{tabular}{lccc}
    \toprule
    \textbf{Noise Type} & \textbf{Baseline} & \textbf{Noise +} & \textbf{+ Noise} \\
    \midrule
    Structural Injection & \multirow{2}{*}{\bm{$72.6$}} & $26.2$ & $68.8$ \\
    Length Injection     &      & $66.5$ & $68.0$ \\
    \bottomrule
\end{tabular}
\end{table}

\subsection{Noise Intervention}
\label{subapp:Noise Intervention}
To rigorously test the structural dependencies identified in our probing and patching experiments, we introduce targeted noise intervention to the table context $x_{table}$.
We design two intervention conditions to decouple token distance from structural count:
(\emph{i})~\textbf{Structural Injection:} Inserting additional pipe delimiters (`$\vert$') to simulate false coordinate boundaries.
(\emph{ii})~\textbf{Length Injection:} Inserting random alphanumeric characters of equivalent token length to test resilience against context expansion.

Table~\ref{tab:nosie} presents the performance impact under these noise settings.
Consistent with our hypothesis that coordinates are synchronized solely by delimiters, the model exhibits remarkable robustness to Length Injection.
The negligible performance drop indicates that the internal counting mechanism is invariant to absolute token distance, suggesting a reliance on discrete delimiter counting rather than token length proximity.
Conversely, Structural Injection preceding the target column induces a catastrophic failure, causing accuracy to plummet from $72.6\%$ to $26.2\%$.
Crucially, delimiter insertions succeeding the target column exert minimal influence, confirming that the counting mechanism is strictly sequential.

\subsection{Details on the Linear Geometry of Tabular Indices}
\label{subapp:geometry_details}

In this subsection, we provide further implementation details and comprehensive results for the vector composition experiments discussed in \S\ref{sec:geometry}.

\paragraph{Vector Extraction and Intervention Setup}
Based on preliminary probing results indicating that column information is most localized in the middle layers, we restrict our extraction and intervention to layers $L \in \{17, \dots, 23\}$.
For a given column header token at position $t$ in $x_{table}$, we extract the residual stream activations at positions $[t-1, t]$ (i.e., the header token and the immediate previous token) to capture the representations of column positions.
The unit shift vectors and composite vectors are computed by averaging differences across $N=100$ samples from the held-out benchmark.
During the steering experiments, we inject the computed vectors into the residual stream with a steering coefficient of $\alpha = 8.0$. The intervention is applied simultaneously to the same token positions $[t-1, t]$ used during extraction.

\paragraph{Vector Composition Analysis}
We validate the linearity of the representation space by comparing the causal effect of a ``true" shift vector (Baseline) against a ``composite" vector formed by summing two distinct shift vectors (e.g., aiming for a shift of $+4$ by summing vectors for $+2$ and $+2$).
Table~\ref{tab:composition_results} details the Effect scores for various target shifts $k \in \{-4, \dots, 4\}$.
The results strongly support the hypothesis of a linear subspace. 
For instance, for a target rightward shift of $k=4$, the baseline vector yields an Effect score of $1.079$. The composite vector formed by $\vec{v}_{+2} + \vec{v}_{+2}$ yields an Effect score of $1.238$, actually surpassing the baseline, suggesting a strong constructive interference of the positional signal. Similarly, for $k=2$, the additive combination $\vec{v}_{+1} + \vec{v}_{+1}$ achieves an Effect score of $0.499$, compared to the baseline of $0.454$.
Even for subtractive compositions (e.g., reaching a target of $-1$ via $\vec{v}_{+1} + \vec{v}_{-2}$), the model maintains robust performance (Effect Score $0.280$ vs. Baseline $0.301$), confirming that the geometric representation supports standard vector arithmetic operations.

\begin{table}[t]
    \centering
    \caption{
        Detailed results of the Vector Composition Experiment. \textbf{Target} denotes the desired column shift $k$. \textbf{Baseline} indicates the vector was extracted directly from column pairs with distance $k$; \textbf{Composition} describes how the vector was constructed.
        \textbf{Effect} reports the Effect Score (higher is better).
    }
    \label{tab:composition_results}
    \tiny
        \begin{tabular}{lccc}
        \toprule
        \textbf{Type} & \textbf{Target ($k$)} & \textbf{Composition ($a+b$)} & \textbf{Effect} \\
        \midrule
\textbf{Baseline} & $\mathbf{-1}$ & - & $0.301$ \\
Composite & -1 & $-6 + 5$ & $0.244$ \\
Composite & -1 & $-5 + 4$ & $0.243$ \\
Composite & -1 & $-4 + 3$ & $0.242$ \\
Composite & -1 & $-3 + 2$ & $0.251$ \\
Composite & -1 & $-2 + 1$ & $0.280$ \\
Composite & -1 & $1 + (-2)$ & $0.280$ \\
Composite & -1 & $2 + (-3)$ & $0.251$ \\
Composite & -1 & $3 + (-4)$ & $0.242$ \\
Composite & -1 & $4 + (-5)$ & $0.243$ \\
Composite & -1 & $5 + (-6)$ & $0.244$ \\
\midrule

\textbf{Baseline} & $\mathbf{1}$ & - & $0.197$ \\
Composite & 1 & $-5 + 6$ & $0.191$ \\
Composite & 1 & $-4 + 5$ & $0.253$ \\
Composite & 1 & $-3 + 4$ & $0.210$ \\
Composite & 1 & $-2 + 3$ & $0.249$ \\
Composite & 1 & $-1 + 2$ & $0.282$ \\
Composite & 1 & $2 + (-1)$ & $0.282$ \\
Composite & 1 & $3 + (-2)$ & $0.249$ \\
Composite & 1 & $4 + (-3)$ & $0.210$ \\
Composite & 1 & $5 + (-4)$ & $0.253$ \\
Composite & 1 & $6 + (-5)$ & $0.191$ \\
\midrule

\textbf{Baseline} & $\mathbf{2}$ & - & $0.454$ \\
Composite & 2 & $-4 + 6$ & $0.432$ \\
Composite & 2 & $-3 + 5$ & $0.437$ \\
Composite & 2 & $-2 + 4$ & $0.430$ \\
Composite & 2 & $-1 + 3$ & $0.410$ \\
Composite & 2 & $1 + 1$ & $0.499$ \\
Composite & 2 & $3 + (-1)$ & $0.410$ \\
Composite & 2 & $4 + (-2)$ & $0.430$ \\
Composite & 2 & $5 + (-3)$ & $0.437$ \\
Composite & 2 & $6 + (-4)$ & $0.432$ \\
\midrule

\textbf{Baseline} & $\mathbf{-2}$ & - & $0.310$ \\
Composite & -2 & $-6 + 4$ & $0.284$ \\
Composite & -2 & $-5 + 3$ & $0.284$ \\
Composite & -2 & $-4 + 2$ & $0.293$ \\
Composite & -2 & $-3 + 1$ & $0.305$ \\
Composite & -2 & $-1 + (-1)$ & $0.301$ \\
Composite & -2 & $1 + (-3)$ & $0.305$ \\
Composite & -2 & $2 + (-4)$ & $0.293$ \\
Composite & -2 & $3 + (-5)$ & $0.284$ \\
Composite & -2 & $4 + (-6)$ & $0.284$ \\
\midrule

\textbf{Baseline} & $\mathbf{3}$ & - & $0.467$ \\
Composite & 3 & $-3 + 6$ & $0.469$ \\
Composite & 3 & $-2 + 5$ & $0.466$ \\
Composite & 3 & $-1 + 4$ & $0.467$ \\
Composite & 3 & $1 + 2$ & $0.479$ \\
Composite & 3 & $2 + 1$ & $0.479$ \\
Composite & 3 & $4 + (-1)$ & $0.467$ \\
Composite & 3 & $5 + (-2)$ & $0.466$ \\
Composite & 3 & $6 + (-3)$ & $0.469$ \\
\midrule

\textbf{Baseline} & $\mathbf{-3}$ & - & $0.361$ \\
Composite & -3 & $-6 + 3$ & $0.351$ \\
Composite & -3 & $-5 + 2$ & $0.356$ \\
Composite & -3 & $-4 + 1$ & $0.364$ \\
Composite & -3 & $-2 + (-1)$ & $0.350$ \\
Composite & -3 & $-1 + (-2)$ & $0.350$ \\
Composite & -3 & $1 + (-4)$ & $0.364$ \\
Composite & -3 & $2 + (-5)$ & $0.356$ \\
Composite & -3 & $3 + (-6)$ & $0.351$ \\
\midrule

\textbf{Baseline} & $\mathbf{4}$ & - & $1.079$ \\
Composite & 4 & $-2 + 6$ & $1.004$ \\
Composite & 4 & $-1 + 5$ & $1.012$ \\
Composite & 4 & $1 + 3$ & $1.169$ \\
Composite & 4 & $2 + 2$ & $1.238$ \\
Composite & 4 & $3 + 1$ & $1.169$ \\
Composite & 4 & $5 + (-1)$ & $1.012$ \\
Composite & 4 & $6 + (-2)$ & $1.004$ \\
\midrule

\textbf{Baseline} & $\mathbf{-4}$ & - & $0.367$ \\
Composite & -4 & $-6 + 2$ & $0.369$ \\
Composite & -4 & $-5 + 1$ & $0.370$ \\
Composite & -4 & $-3 + (-1)$ & $0.357$ \\
Composite & -4 & $-2 + (-2)$ & $0.357$ \\
Composite & -4 & $-1 + (-3)$ & $0.357$ \\
Composite & -4 & $1 + (-5)$ & $0.370$ \\
Composite & -4 & $2 + (-6)$ & $0.369$ \\
        \bottomrule
    \end{tabular}
\end{table}

\section{Experiments of LLMs}
\label{app:llms}

In the section, we show the experiments and analysis of other evaluated models (including Qwen3-0.6B, Qwen2.5-32B, Llama-3.2-3B, Llama-3.1-8B). We conclude that the findings in the main text can generalize to the models with different scales successfully.

\begin{table}[t]
\centering
\small
\caption{EM performance of evaluated models on the cell location benchmark.}
\label{tab:models_performance}
\begin{tabular}{lc}
\toprule
\textbf{Model} & \textbf{EM Score} \\
\midrule
Qwen3-0.6B    & $38.8$ \\
Qwen3-4B      & $90.0$ \\
Qwen2.5-32B   & \bm{$98.0$} \\
Llama3.2-3B   & $52.8$ \\
Llama3.1-8B   & $70.0$ \\
\bottomrule
\end{tabular}
\end{table}

\subsection{Evaluation of LLMs}
\label{subapp:evaluation of llms}


In this subsection, we present the Exact Match (EM) scores of various models evaluated on our benchmark of atomic cell location. The EM metric requires the model's response to be an identical match with the gold answer. All results are obtained using greedy decoding with the temperature set to $0$. As shown in Table~\ref{tab:models_performance}, Qwen3-4B and Qwen2.5-32B achieve superior performance with scores of $90.0$ and above. In the subsequent sections, we conduct a mechanistic analysis of the other models on the cell location task to determine if their table understanding processes align with those of Qwen3-4B and to investigate the underlying causes of their performance gaps.

\subsection{The Three-Stage Information Flow of Table Understanding}
\label{subapp:three-stage}

\begin{figure*}[t]
    \centering
    \begin{subfigure}[b]{0.24\textwidth} 
        \centering
        \includegraphics[width=\linewidth]{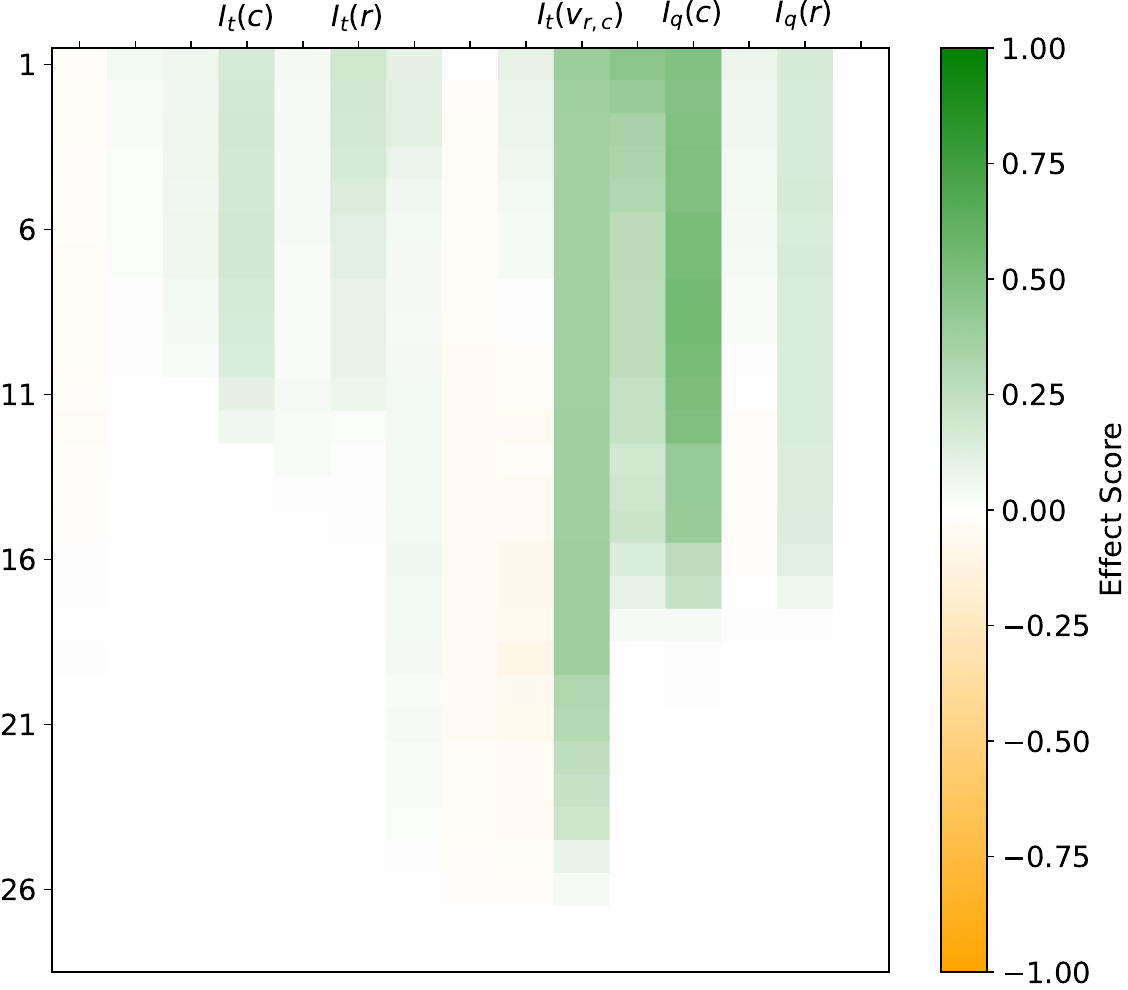}
        \caption{Qwen3-0.6B}
        \label{fig:Qwen3-0.6B_patch_columnba}
    \end{subfigure}
    \hfill
    \begin{subfigure}[b]{0.24\textwidth}
        \centering
        \includegraphics[width=\linewidth]{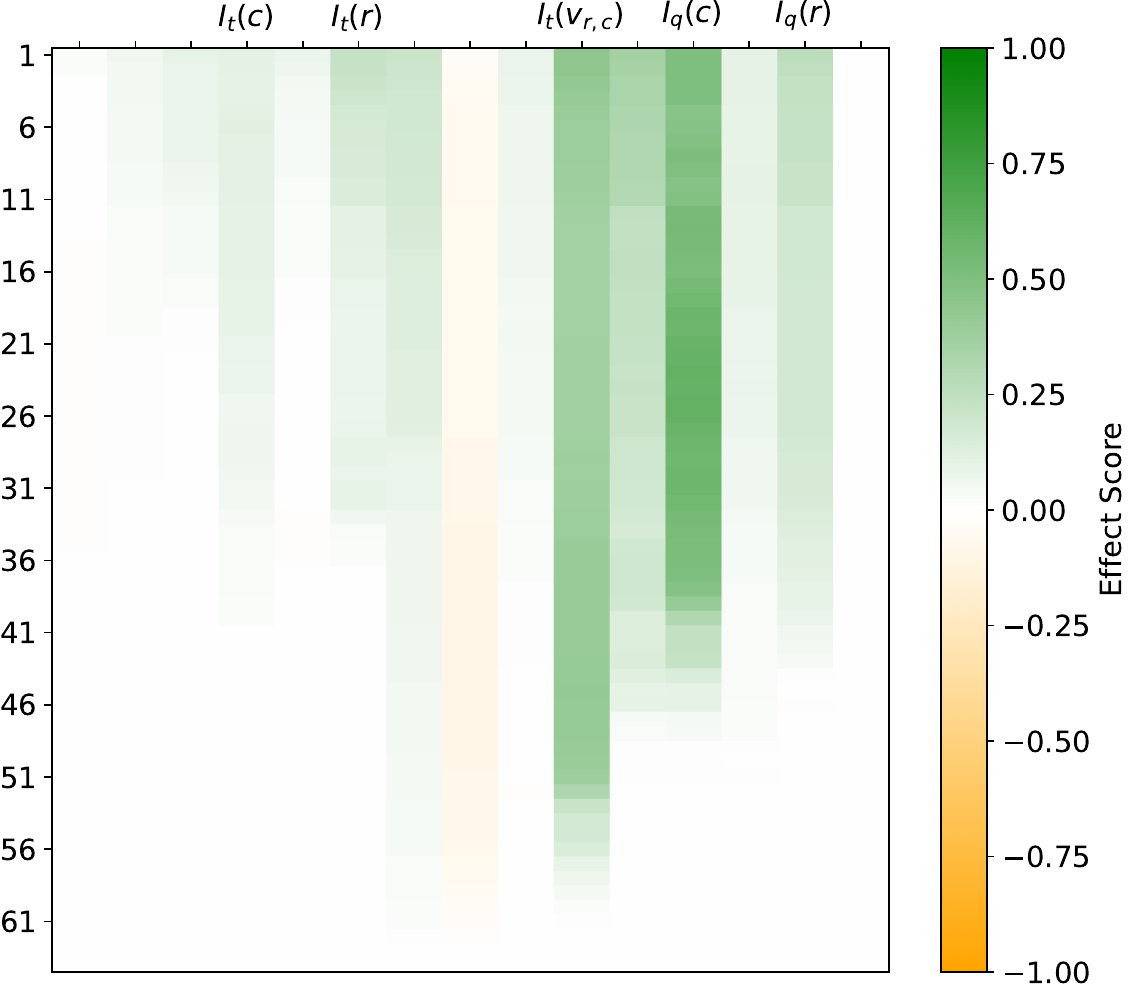}
        \caption{Qwen2.5-32B}
        \label{fig:Qwen2.5-32B_patch_columnba}
    \end{subfigure}
    \hfill
    \begin{subfigure}[b]{0.24\textwidth}
        \centering
        \includegraphics[width=\linewidth]{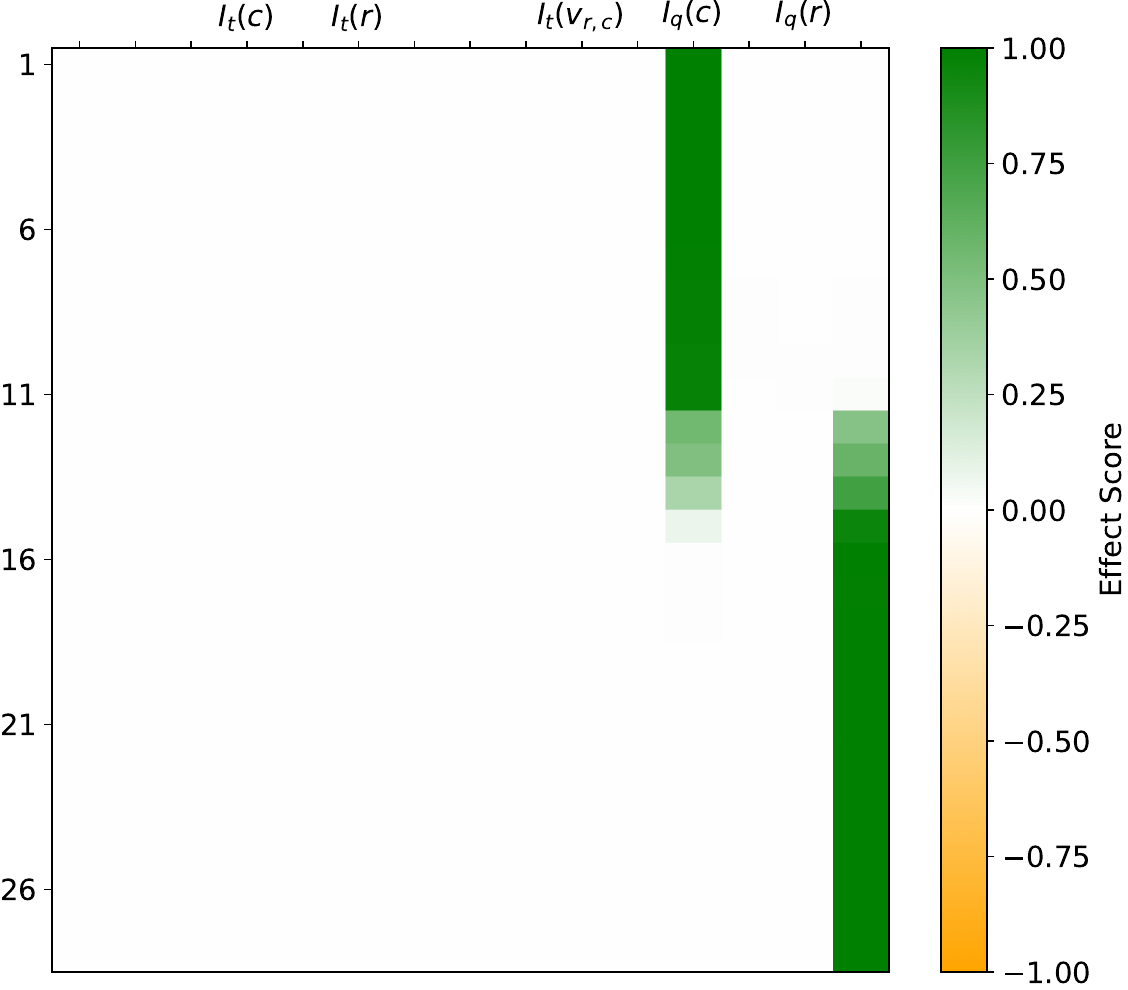}
        \caption{Llama-3.2-3B}
        \label{fig:Llama-3.2-3B_patch_columnba}
    \end{subfigure}
    \hfill
    \begin{subfigure}[b]{0.24\textwidth}
        \centering
        \includegraphics[width=\linewidth]{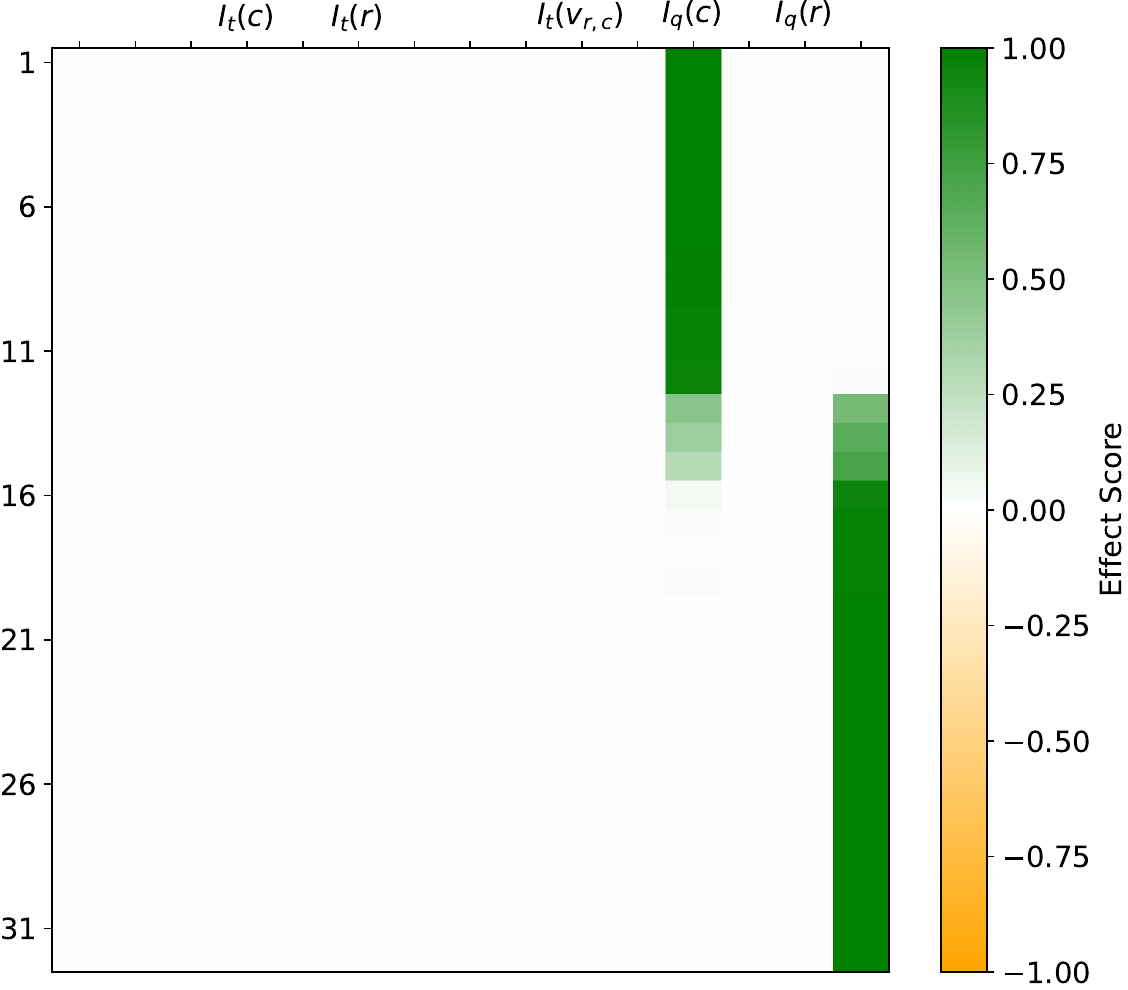}
        \caption{Llama-3.1-8B}
        \label{fig:Llama-3.1-8B_patch_columnba}
    \end{subfigure}
    
    \caption{Average column patching effect score across layers and token positions on various LLMs. }
    \label{fig:patch_llms}
\end{figure*}

\begin{table}[t]
\centering
\small
\caption{Approximate layer-wise distribution of the three-stage information flow across different LLMs.}
\label{tab:model_stages}

\begin{tabular}{l|ccc}
\toprule
\textbf{Model} & \textbf{Stage I} & \textbf{Stage II} & \textbf{Stage III} \\ 
\midrule
Qwen3-0.6B    & Layers 1-12               & Layers 13-18                    & Layers 19-28                   \\
Qwen2.5-32B   & Layers 1-26               & Layers 27-46                    & Layers 47-64                   \\
Llama-3.2-3B  & Layers 1-11               & Layers 12-14                    & Layers 15-28                   \\
Llama-3.1-8B  & Layers 1-12               & Layers 13-15                    & Layers 16-32                   \\ 
\bottomrule
\end{tabular}
\end{table}

\subsubsection{Overview}

To evaluate the universality of the three-stage information flow, we extend our analysis to Qwen3-0.6B, Qwen2.5-32B, Llama-3.2-3B, and Llama-3.1-8B. While all models transition through semantic grounding and localization, the heatmaps in Figure~\ref{fig:patch_llms} reveal two distinct mechanistic paradigms in how the internal representations utilize the input sequence.
The Qwen models (Figures~\ref{fig:Qwen3-0.6B_patch_columnba} and \ref{fig:Qwen2.5-32B_patch_columnba}) align closely with the mechanism described in the main text. 
We summarize the layer-wise distribution of the three-stage information flow in Table~\ref{tab:model_stages} and analyze them in detail.
\begin{itemize}
    \item \textbf{Qwen3-0.6B (28 Layers):} Stage I (Layers 1-12) focuses on query constraints. Stage II (Layers 13-18) shows a distinct causal bridge where interventions on the table tokens ($\mathcal{I}_t$) significantly affect the output. Stage III (Layers 19-28) shifts the mass to the final generation position.
    \item \textbf{Qwen2.5-32B (64 Layers):} The Stage II localization (Layers 27-46) involves nearly the entire vertical span of the table sequence, confirming that the model actively reads the serialized coordinates from the table tokens to resolve the target value.
\end{itemize}

In contrast, the Llama series (Figures~\ref{fig:Llama-3.2-3B_patch_columnba} and \ref{fig:Llama-3.1-8B_patch_columnba}) displays a sparse causal pattern. Notably, the table token positions remain almost entirely white (effect score $\approx 0$), suggesting a more direct mapping:
\begin{enumerate}
    \item \textbf{Stage I:} For Llama-3.2-3B (Layers 1-10) and Llama-3.1-8B (Layers 1-11), significant patching effects are concentrated strictly on the query constraints ($\mathcal{I}_q$). This indicates that the initial semantic binding is purely focused on identifying the constraints within the question.
    \item \textbf{Stage II:} Between the early and late layers, there is a narrow transitional window (e.g., Layers 12-14 in Llama-3.2) where the causal mass begins to shift. However, unlike Qwen, this transition bypasses the table tokens in the residual stream.
    \item \textbf{Stage III:} High effect scores (approaching $1.0$) appear abruptly at the position of the final token (e.g., Layer 15+ for 3B, Layer 16+ for 8B). 
\end{enumerate}
The absence of causal effects in the table region suggests that Llama models do not necessarily move information through the table tokens' hidden states. Instead, the late-layer attention heads at the final token position likely perform long-range retrieval directly from the table's KV-cache based on the semantic signals processed in the early-layer query tokens.
The comparison reveals that while what the models compute is the same (the three stages), where they store that computation differs. Qwen utilizes the table tokens as intermediate state-holders for localization, whereas Llama performs a more direct computation from the query constraints to the final output generation.

\begin{figure*}[t]
    \centering
    \begin{subfigure}[b]{0.24\textwidth} 
        \centering
        \includegraphics[width=\linewidth]{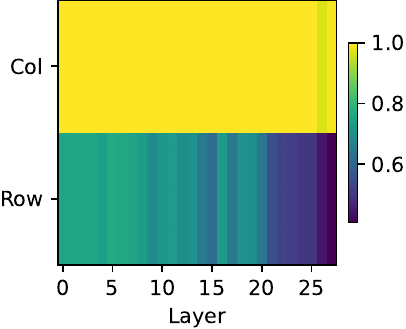}
        \caption{Qwen3-0.6B}
        \label{fig:Qwen3-0.6B-rc_similarity}
    \end{subfigure}
    \hfill
    \begin{subfigure}[b]{0.24\textwidth}
        \centering
        \includegraphics[width=\linewidth]{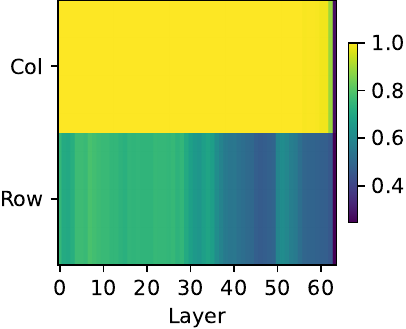}
        \caption{Qwen2.5-32B}
        \label{fig:Qwen2.5-32B-rc_similarity}
    \end{subfigure}
    \hfill
    \begin{subfigure}[b]{0.24\textwidth}
        \centering
        \includegraphics[width=\linewidth]{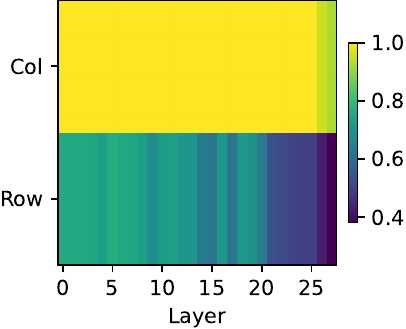}
        \caption{Llama-3.2-3B}
        \label{fig:Llama3.2-3B-rc_similarity}
    \end{subfigure}
    \hfill
    \begin{subfigure}[b]{0.24\textwidth}
        \centering
        \includegraphics[width=\linewidth]{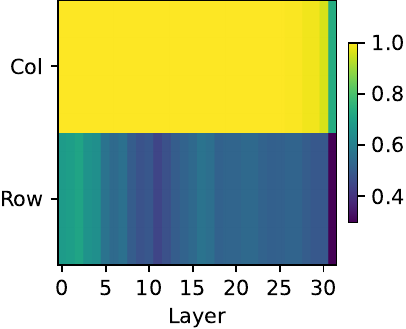}
        \caption{Llama-3.1-8B}
        \label{fig:Llama3.1-8B-rc_similarity}
    \end{subfigure}
    
    \caption{Analysis of semantic binding mechanisms: the top-1 accuracy evaluated with cosine similarity between query constraints and all table headers across layers of LLMs.}
    \label{fig:rc_similarity_llms}
\end{figure*}

\begin{figure*}[t]
    \centering
    \begin{subfigure}[b]{0.24\textwidth} 
        \centering
        \includegraphics[width=\linewidth]{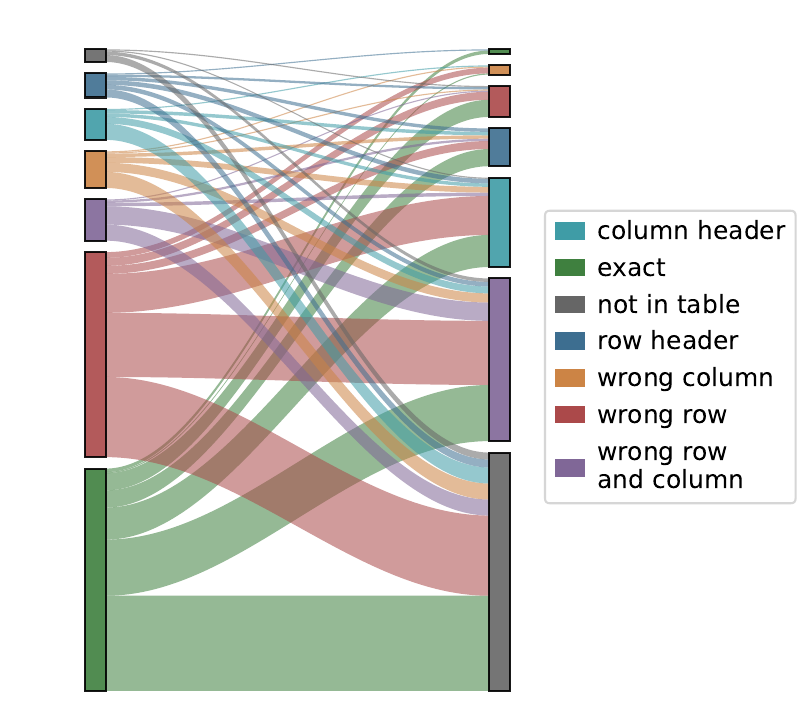}
        \caption{Qwen3-0.6B}
        \label{fig:Qwen3-0.6B_sankey}
    \end{subfigure}
    \hfill
    \begin{subfigure}[b]{0.24\textwidth}
        \centering
        \includegraphics[width=\linewidth]{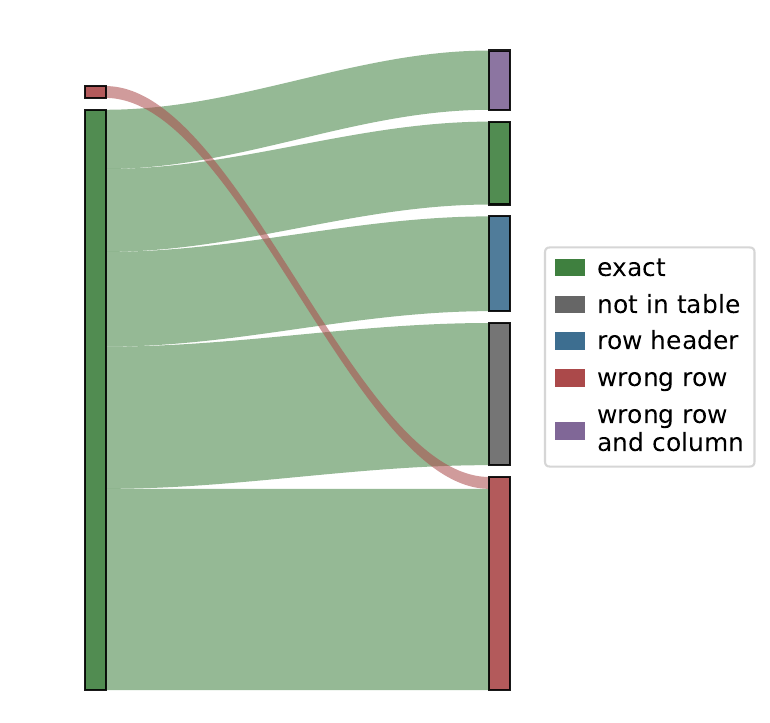}
        \caption{Qwen2.5-32B}
        \label{fig:Qwen2.5-32B_sankey}
    \end{subfigure}
    \hfill
    \begin{subfigure}[b]{0.24\textwidth}
        \centering
        \includegraphics[width=\linewidth]{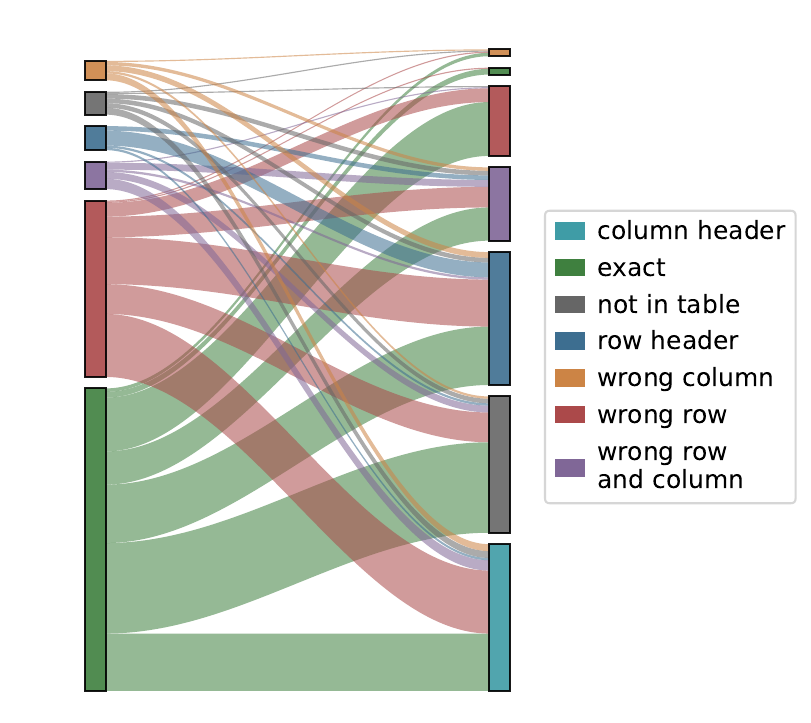}
        \caption{Llama-3.2-3B}
        \label{fig:Llama3.2-3B_sankey}
    \end{subfigure}
    \hfill
    \begin{subfigure}[b]{0.24\textwidth}
        \centering
        \includegraphics[width=\linewidth]{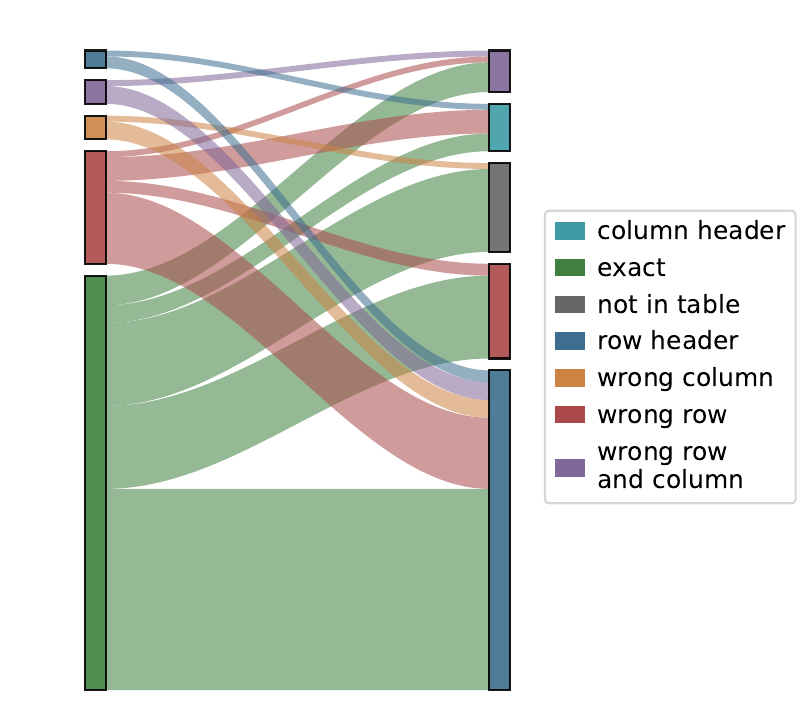}
        \caption{Llama-3.1-8B}
        \label{fig:Llama3.1-8B_sankey}
    \end{subfigure}
    
    \caption{Analysis of semantic binding mechanisms: the impact of ablating row alignment heads on various LLMs.}
    \label{fig:sankey_llms}
\end{figure*}

\subsubsection{Stage I: Semantic Binding}

To validate the universality of the semantic binding stage, we extend our analysis to the broader model suite. As illustrated in Figures~\ref{fig:rc_similarity_llms} and \ref{fig:sankey_llms}, while the specific layer boundaries vary according to total depth, the fundamental mechanism of binding query constraints ($\mathcal{I}_q$) to table headers ($\mathcal{I}_t$) remains consistent across different models and scales.
The layer-wise cosine similarity heatmaps (Figure~\ref{fig:rc_similarity_llms}) demonstrate that all models achieve near-perfect retrieval accuracy ($1.0$) for both row and column headers in their respective early layers. 
\begin{itemize}
    \item \textbf{Qwen Series:} For Qwen3-0.6B and Qwen2.5-32B, the alignment converges rapidly. In Qwen3-0.6B, binding is stabilized by Layer 10, while in the larger Qwen2.5-32B, this process is completed within the first 15 layers. Consistent with the main text, column alignment consistently reaches a high-similarity plateau earlier than row alignment, indicating a lower inherent difficulty in binding column-wise semantic information.
    \item \textbf{Llama Series:} Llama-3.2-3B and Llama-3.1-8B exhibit a similar convergent pattern within Layers 1-12. However, a notable observation in Figures~\ref{fig:Llama3.2-3B-rc_similarity} and \ref{fig:Llama3.1-8B-rc_similarity} is the slight noise or delayed convergence in row retrieval compared to the Qwen series. This suggests that Llama models may require more transformation steps in the residual stream to align row semantics before moving to the localization stage.
\end{itemize}

The Sankey diagrams (Figure~\ref{fig:sankey_llms}) visualize the performance collapse following the zero-ablation of identified \textit{row alignment heads}. 
\begin{itemize}
    \item \textbf{Qwen3-0.6B \& Qwen2.5-32B:} These models show a highly concentrated causal dependency. In Qwen2.5-32B, abating these heads leads to a near-total elimination of \texttt{exact} matches, with the majority of errors flowing into \texttt{not in table} and \texttt{wrong row}. This confirms that the identified heads are the primary functional units responsible for grounding the query row constraint to the table's structure.
    \item \textbf{Llama-3.2-3B \& Llama-3.1-8B:} The Llama series exhibits a more complex error distribution. While \texttt{exact} performance drops significantly, a non-negligible portion of the flow transitions to \texttt{row header} and \texttt{column header} errors. These specific error types represent cases where the model directly outputs the query constraints rather than the cell value. Combined with the paradigm observed in Figure~\ref{fig:patch_llms}, this suggests that when row binding heads are ablated, Llama models default to a copy-mechanism from the query tokens in the KV-cache, failing to initiate the retrieval process from the table.
\end{itemize}

In summary, across all tested LLMs, Stage I is characterized by a rapid geometric alignment in the early residual stream and is causally driven by a sparse set of specialized attention heads. The experimental results for these models align with the layer-wise distributions proposed in Table~\ref{tab:model_stages}, confirming that semantic binding is a prerequisite for subsequent coordinate localization.

\begin{figure*}[t]
    \centering
    \begin{subfigure}[b]{0.24\textwidth} 
        \centering
        \includegraphics[width=\linewidth]{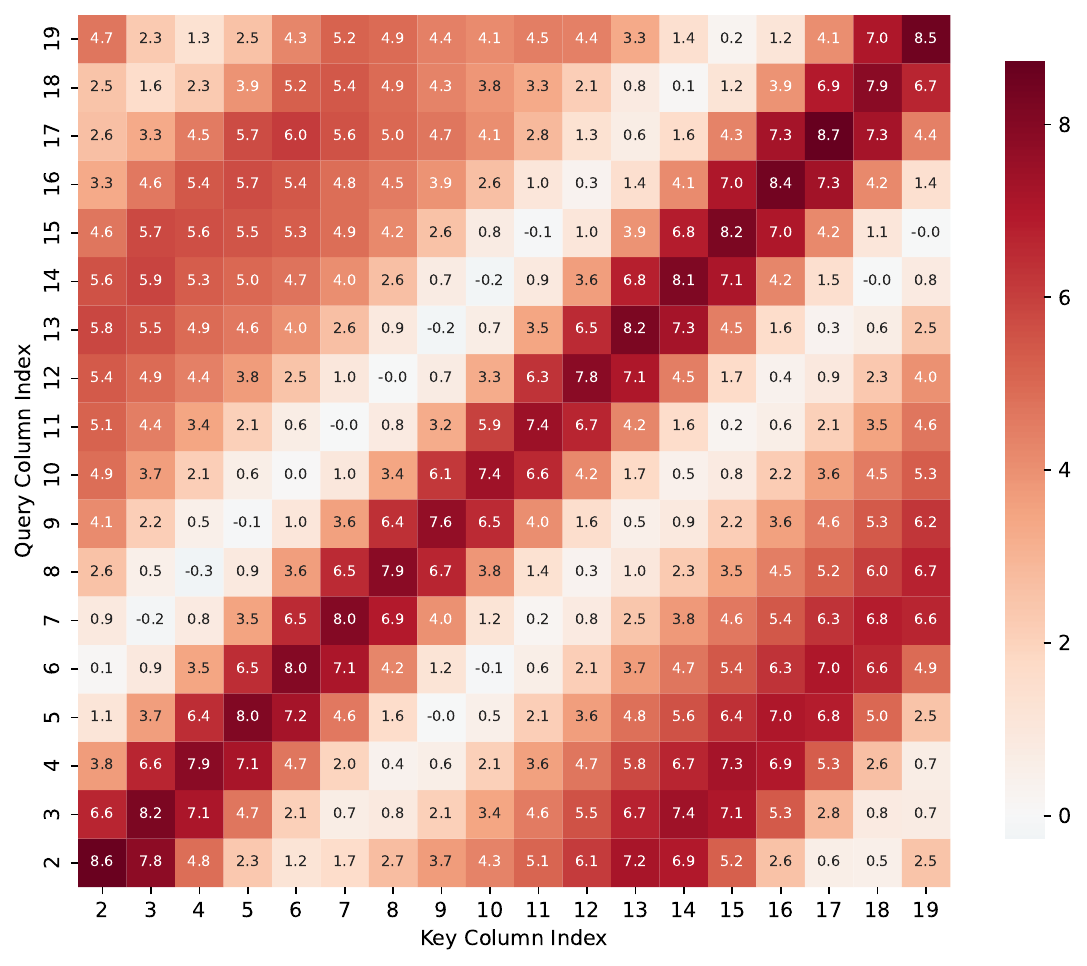}
        \caption{Qwen3-0.6B}
        \label{fig:Qwen3-0.6B_qk_heatmap}
    \end{subfigure}
    \hfill
    \begin{subfigure}[b]{0.24\textwidth}
        \centering
        \includegraphics[width=\linewidth]{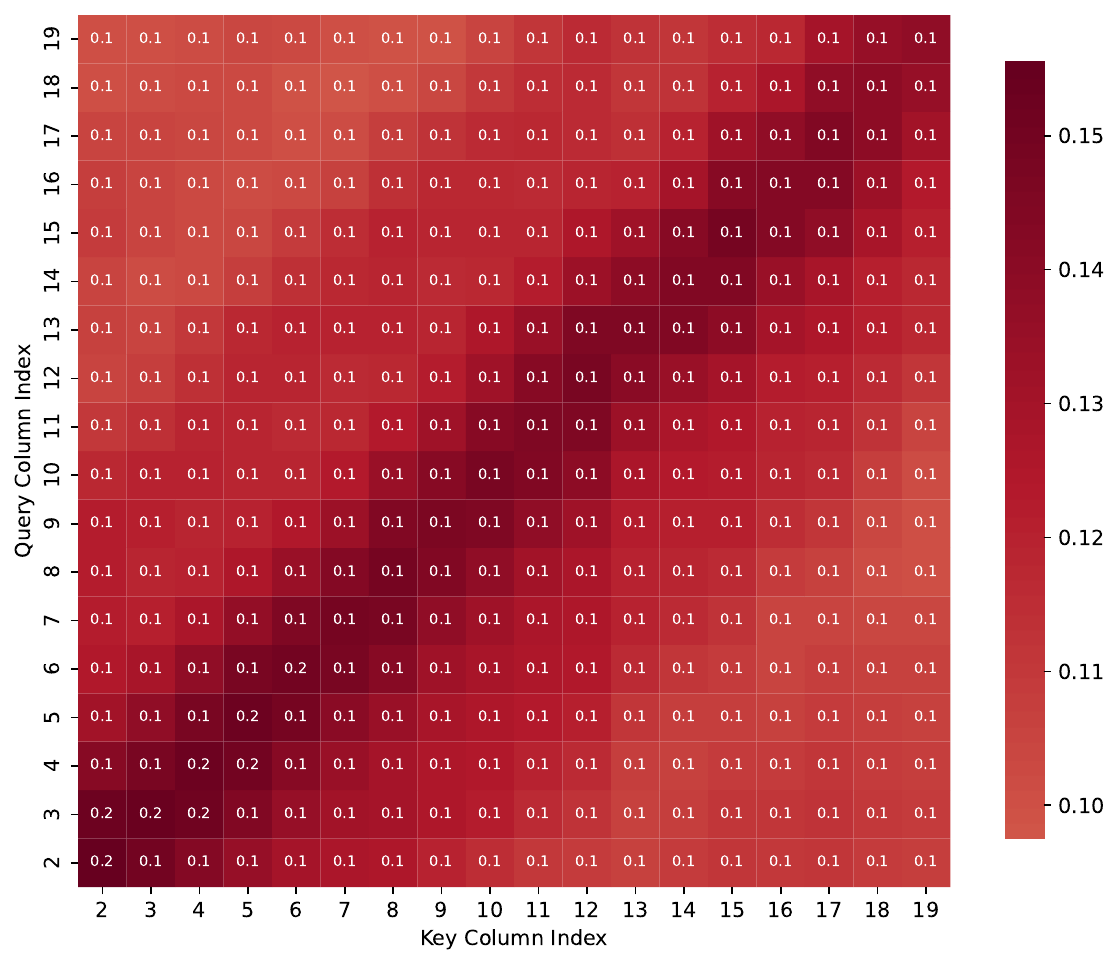}
        \caption{Qwen2.5-32B}
        \label{fig:Qwen2.5-32B_qk_heatmap}
    \end{subfigure}
    \hfill
    \begin{subfigure}[b]{0.24\textwidth}
        \centering
        \includegraphics[width=\linewidth]{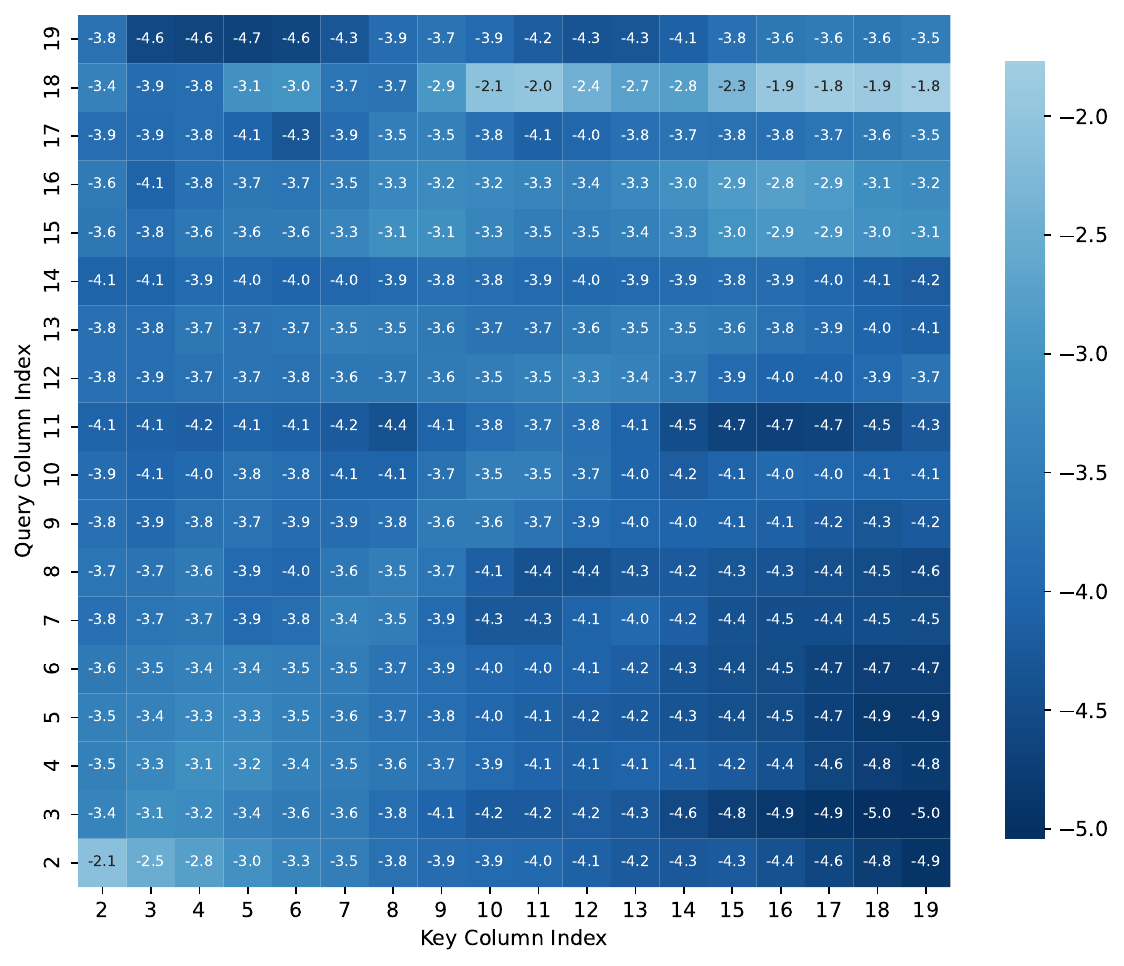}
        \caption{Llama-3.2-3B}
        \label{fig:Llama3.2-3B_qk_heatmap}
    \end{subfigure}
    \hfill
    \begin{subfigure}[b]{0.24\textwidth}
        \centering
        \includegraphics[width=\linewidth]{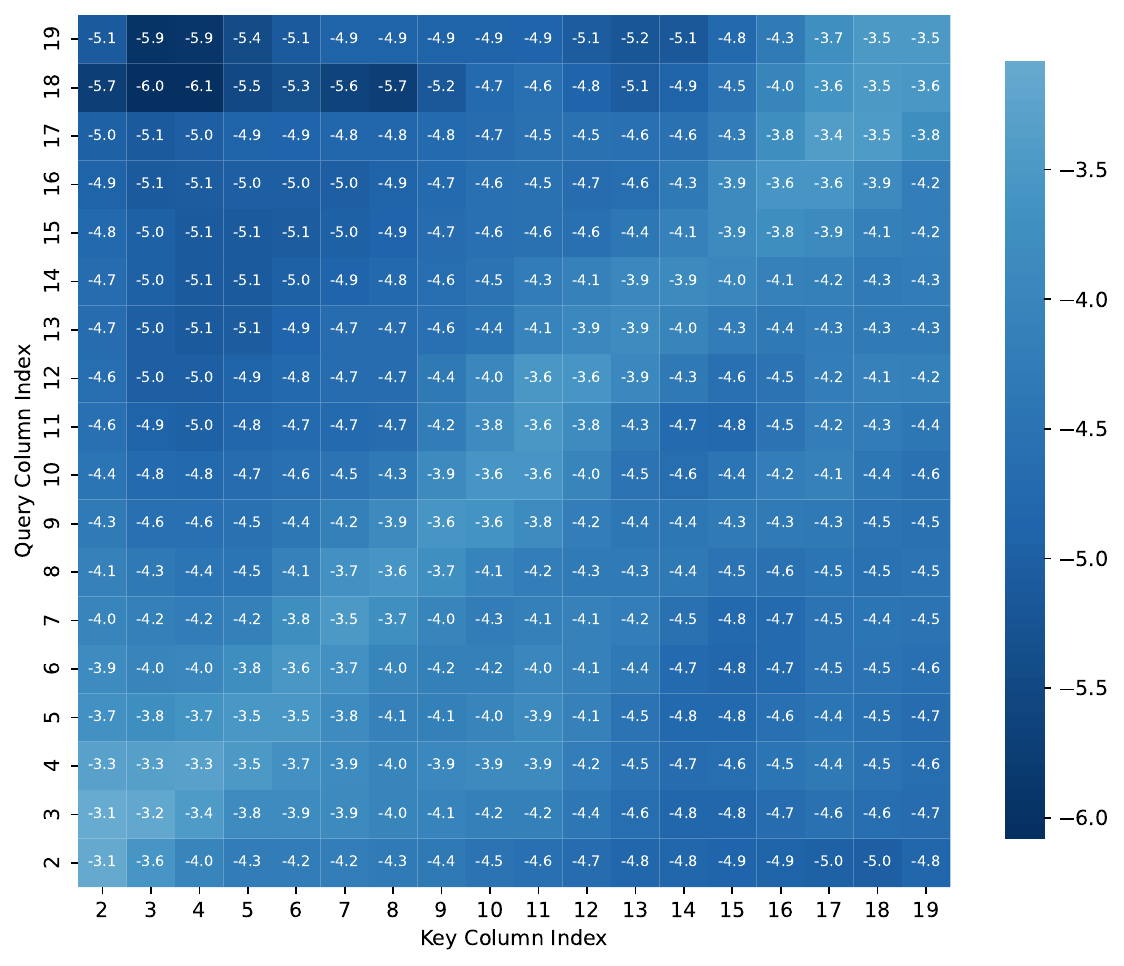}
        \caption{Llama-3.1-8B}
        \label{fig:Llama3.1-8B_qk_heatmap}
    \end{subfigure}
    
    \caption{
        Visualization of the geometric interaction score $S_{j,l}$ in coordinate-encoding heads of various LLMs. 
        The heatmaps illustrate the dot product between column-averaged query vectors ($\bar{q}_j$) and key vectors ($\bar{k}_l$) after injecting RoPE corresponding to their column indices. 
    }
    \label{fig:qk_heatmap_llms}
\end{figure*}

\subsubsection{Stage II: Coordinate Localization}

Following the methodology established in \S\ref{subsec:stage 2}, we extend our mechanistic analysis of coordinate localization to a broader range of models. We compute the geometric interaction score to investigate whether these models utilize structural geometry to locate target cells. The interaction score $S_{j,l}$ is computed by injecting RoPE into column-averaged query ($\bar{q}_j$) and key ($\bar{k}_l$) vectors, as illustrated in Figure~\ref{fig:qk_heatmap_llms}.
For Qwen3-0.6B and Qwen2.5-32B, the Column-RoPE probe reveals a highly distinct diagonal pattern (Figures~\ref{fig:Qwen3-0.6B_qk_heatmap} and \ref{fig:Qwen2.5-32B_qk_heatmap}), consistent with the observations for Qwen3-4B in the main text. 
\begin{itemize}
    \item \textbf{Qwen3-0.6B:} The prominent diagonal $j=l$ indicates that even at a smaller scale, the model identifies specialized heads in middle layers (Layers 13-18) that prioritize tokens based on their relative column indices.
    \item \textbf{Qwen2.5-32B:} The larger model exhibits a cleaner diagonal with minimal off-diagonal noise, suggesting that larger models develop even more precise coordinate-encoding projections to leverage the periodic properties of RoPE for structural navigation.
\end{itemize}
These results confirm that the Qwen series consistently transitions from semantic binding to a table-mediated localization stage, where the residual stream explicitly utilizes the table's internal geometry to resolve target coordinates \cite{li20252dtpe}.

The Llama series (Llama-3.2-3B and Llama-3.1-8B) presents a more nuanced geometric interaction. As shown in Figures~\ref{fig:Llama3.2-3B_qk_heatmap} and \ref{fig:Llama3.1-8B_qk_heatmap}, while a diagonal pattern exists, it is characterized by significantly \textit{negative} interaction scores (darker blue regions).
In attention mechanisms, a large negative dot product between query and key vectors effectively serves as a soft-mask, suppressing attention to specific positions. The presence of a negative diagonal suggests that Llama models may utilize coordinate-encoding heads as inhibitory filters. In this paradigm, rather than explicitly selecting the target column, the model may be suppressing irrelevant columns or structural noise to isolate the relevant KV-cache signals \cite{wang2023forbidden}.
This observation aligns with the paradigm noted Figure~\ref{fig:patch_llms}. 
Since Llama models exhibit negligible causal effects on table tokens (as shown in Figure~\ref{fig:patch_llms}), the localization likely occurs via long-range attention from the final token position directly to the KV-cache. The coordinate-encoding mechanism in Llama appears to act as a latent geometric pointer that facilitates this long-range retrieval without storing intermediate localization states in the table-region residual stream.

In conclusion, despite the divergence in the sign of the interaction scores, all tested models exhibit a clear reliance on RoPE-based structural signals during Stage II. This confirms that the transition from semantic matching to geometric localization is a universal prerequisite for table understanding across modern LLM models.

\begin{figure*}[t]
    \centering
    \begin{subfigure}[b]{0.24\textwidth} 
        \centering
        \includegraphics[width=\linewidth]{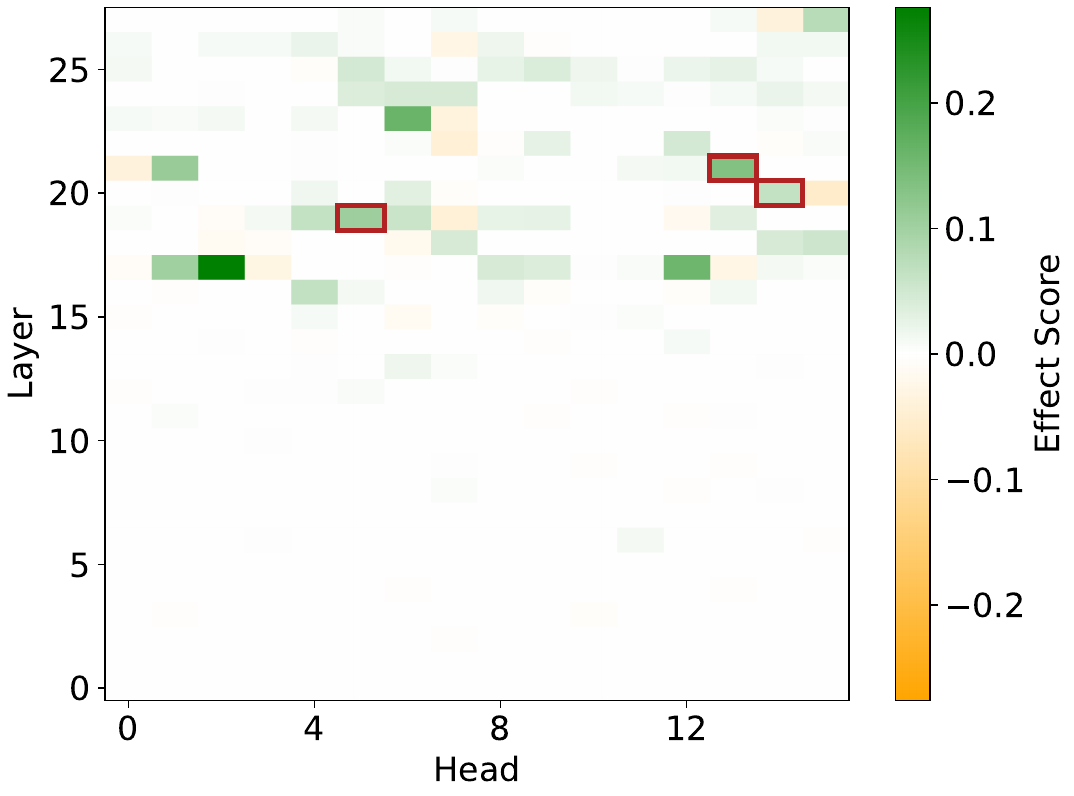}
        \caption{Qwen3-0.6B}
        \label{fig:Qwen3-0.6B_last_patch}
    \end{subfigure}
    \hfill
    \begin{subfigure}[b]{0.24\textwidth}
        \centering
        \includegraphics[width=\linewidth]{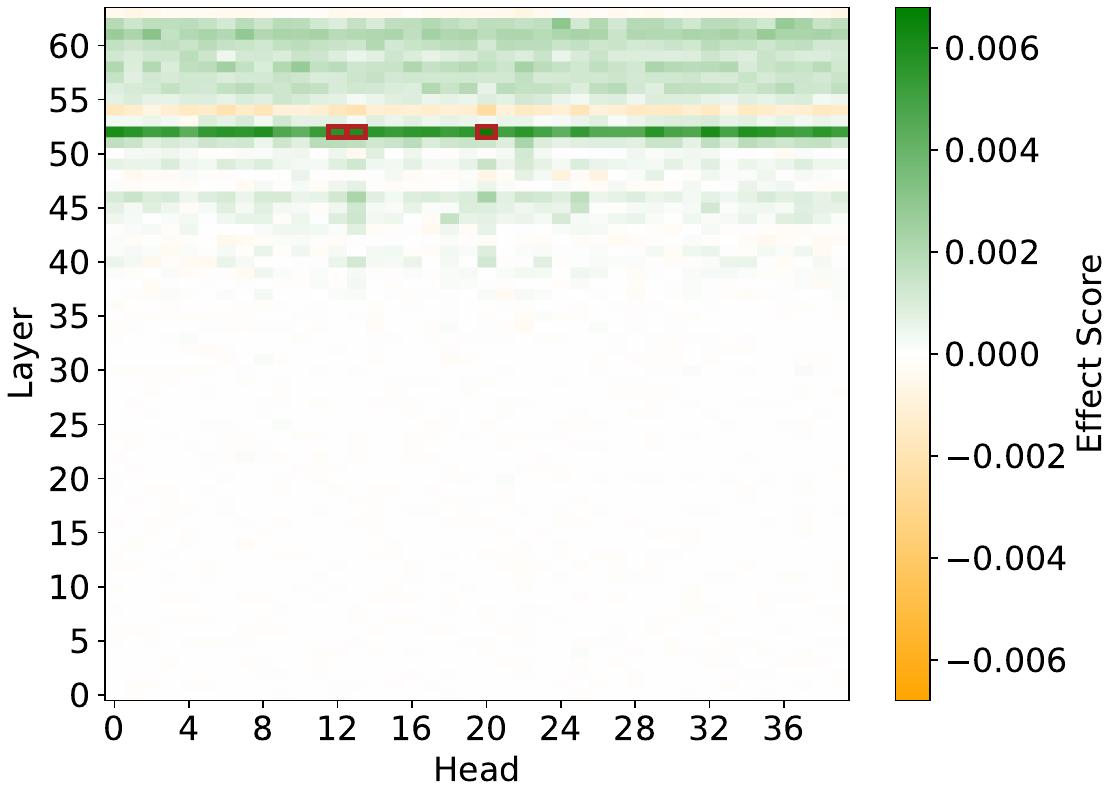}
        \caption{Qwen2.5-32B}
        \label{fig:Qwen2.5-32B_last_patch}
    \end{subfigure}
    \hfill
    \begin{subfigure}[b]{0.24\textwidth}
        \centering
        \includegraphics[width=\linewidth]{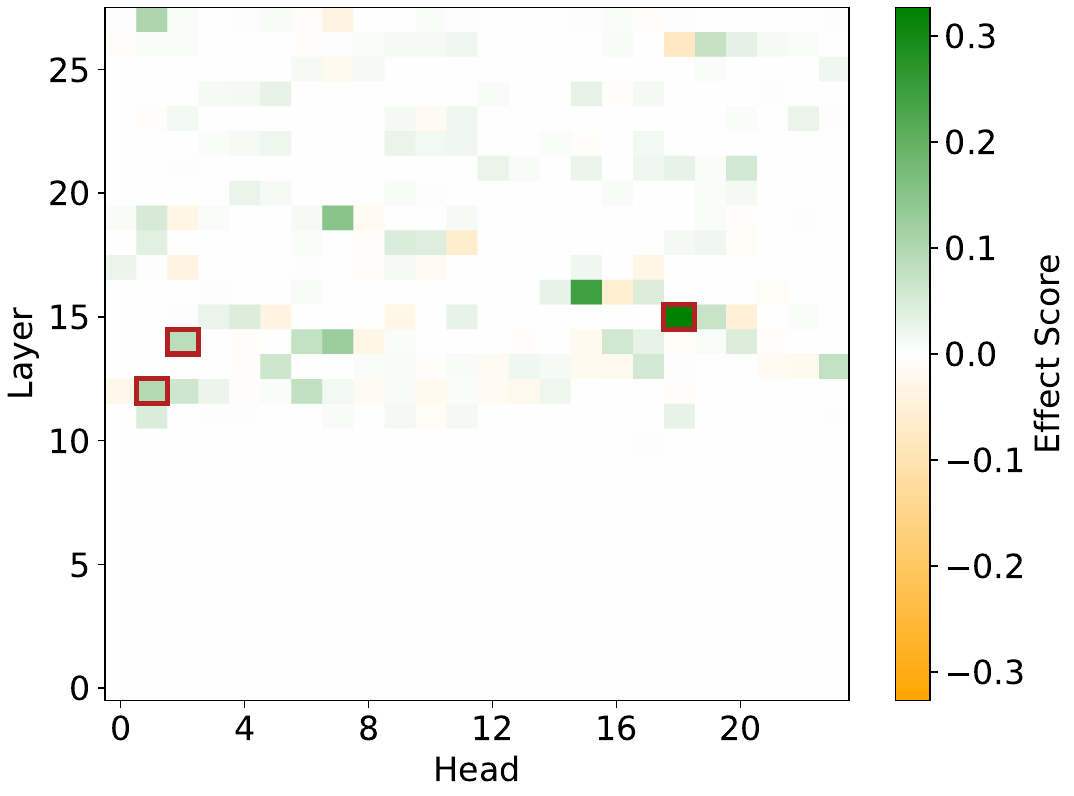}
        \caption{Llama-3.2-3B}
        \label{fig:Llama3.2-3B_last_patch}
    \end{subfigure}
    \hfill
    \begin{subfigure}[b]{0.24\textwidth}
        \centering
        \includegraphics[width=\linewidth]{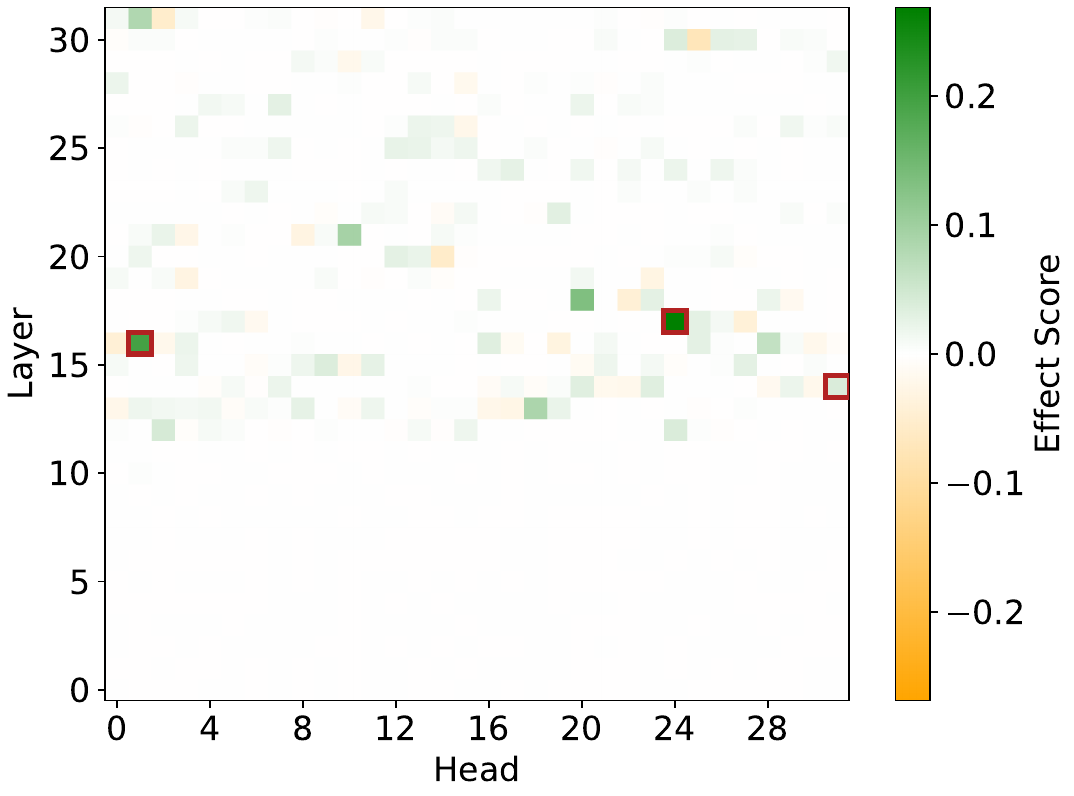}
        \caption{Llama-3.1-8B}
        \label{fig:Llama3.1-8B_last_patch}
    \end{subfigure}
    
    \caption{
        Effect Scores of attention heads at the final token position on various LLMs. 
        The red boxes highlight three critical heads that exhibit positive Effect Scores and act as the primary information movers.
    }
    \label{fig:last_patch_llms}
\end{figure*}

\subsubsection{Stage III: Information Propagation}

To verify the generalizability of the information propagation mechanism, we replicate the final-token activation patching experiment on other models. Following the setup in \S\ref{subsec:stage 3}, we intervene on the query constraints to target a counterfactual cell and compute the Effect Scores for all attention heads at the final token position. The results are visualized in Figure~\ref{fig:last_patch_llms}.
For Qwen3-0.6B (Figure~\ref{fig:Qwen3-0.6B_last_patch}), Llama-3.2-3B (Figure~\ref{fig:Llama3.2-3B_last_patch}), and Llama-3.1-8B (Figure~\ref{fig:Llama3.1-8B_last_patch}), we observe a clear concentration of high Effect Scores in the late layers (Stage III), consistent with the findings for Qwen3-4B.
\begin{itemize}
    \item \textbf{Qwen Serious:} In Qwen3-0.6B and Qwen2.5-32B, causal mass emerges prominently starting from Layer 19 and Layer 47, respectively. The identified \textit{information mover heads} (highlighted in red) exhibit high precision in reading the resolved value from the localized table coordinates.
    \item \textbf{Llama Series:} In Llama-3.2-3B and Llama-3.1-8B, the propagation stage is also localized, appearing abruptly in the late layers. This supports our previous hypothesis: since Llama bypasses intermediate table-mediated storage, its late-layer heads must perform a direct, high-impact retrieval from the KV-cache to move information to the final residual stream.
\end{itemize}


Across all models, the emergence of positive Effect Scores at the final token position serves as a definitive marker for Stage III. The temporal alignment of these heads with the final layers (as specified in Table~\ref{tab:model_stages}) confirms that the three-stage pipeline is a robust mechanistic property of transformer-based table understanding.

\begin{figure*}[t]
    \centering
    \begin{subfigure}[b]{0.48\textwidth} 
        \centering
        \includegraphics[width=\linewidth]{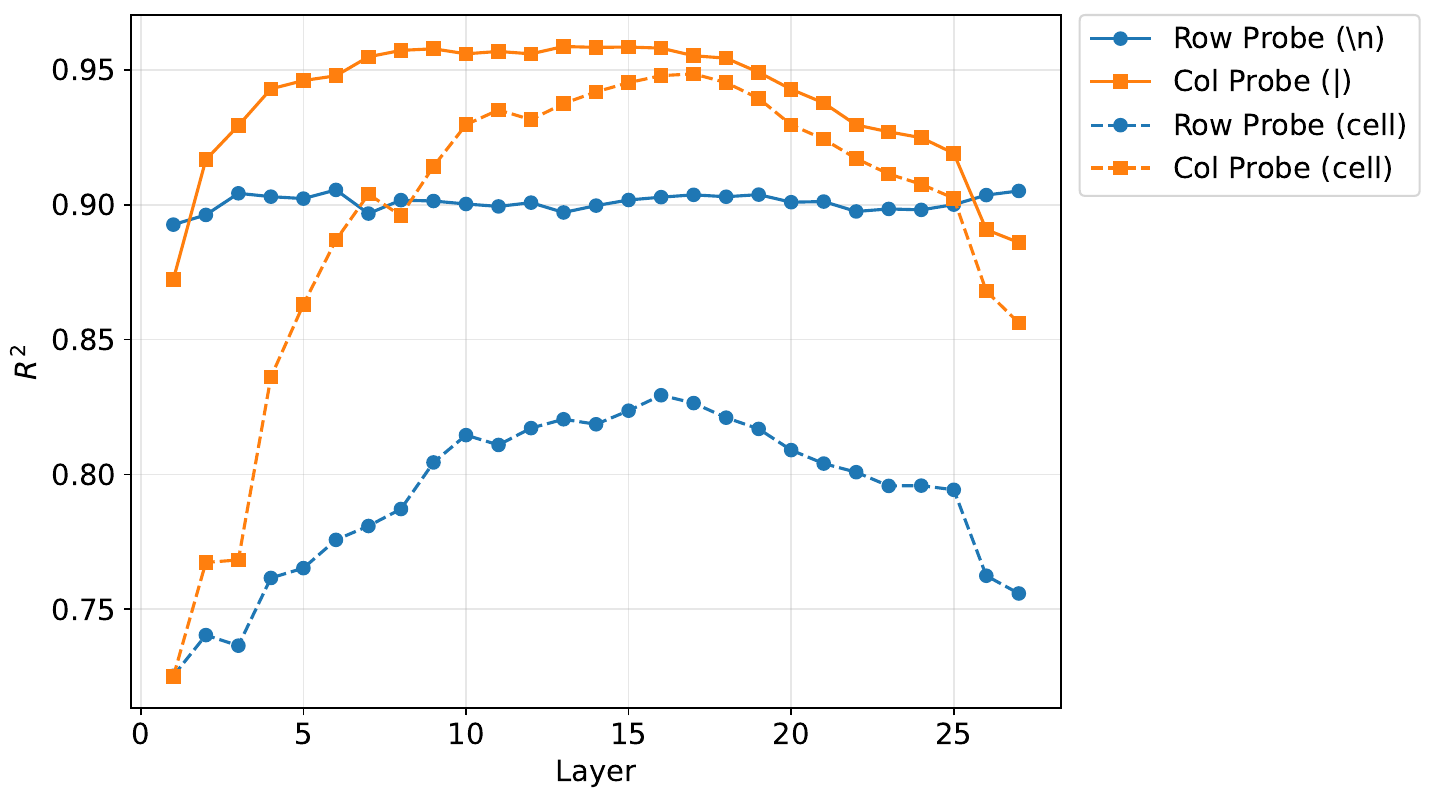}
        \caption{Qwen3-0.6B}
        \label{fig:Qwen3-0.6B_probe_r2}
    \end{subfigure}
    \hfill
    \begin{subfigure}[b]{0.48\textwidth}
        \centering
        \includegraphics[width=\linewidth]{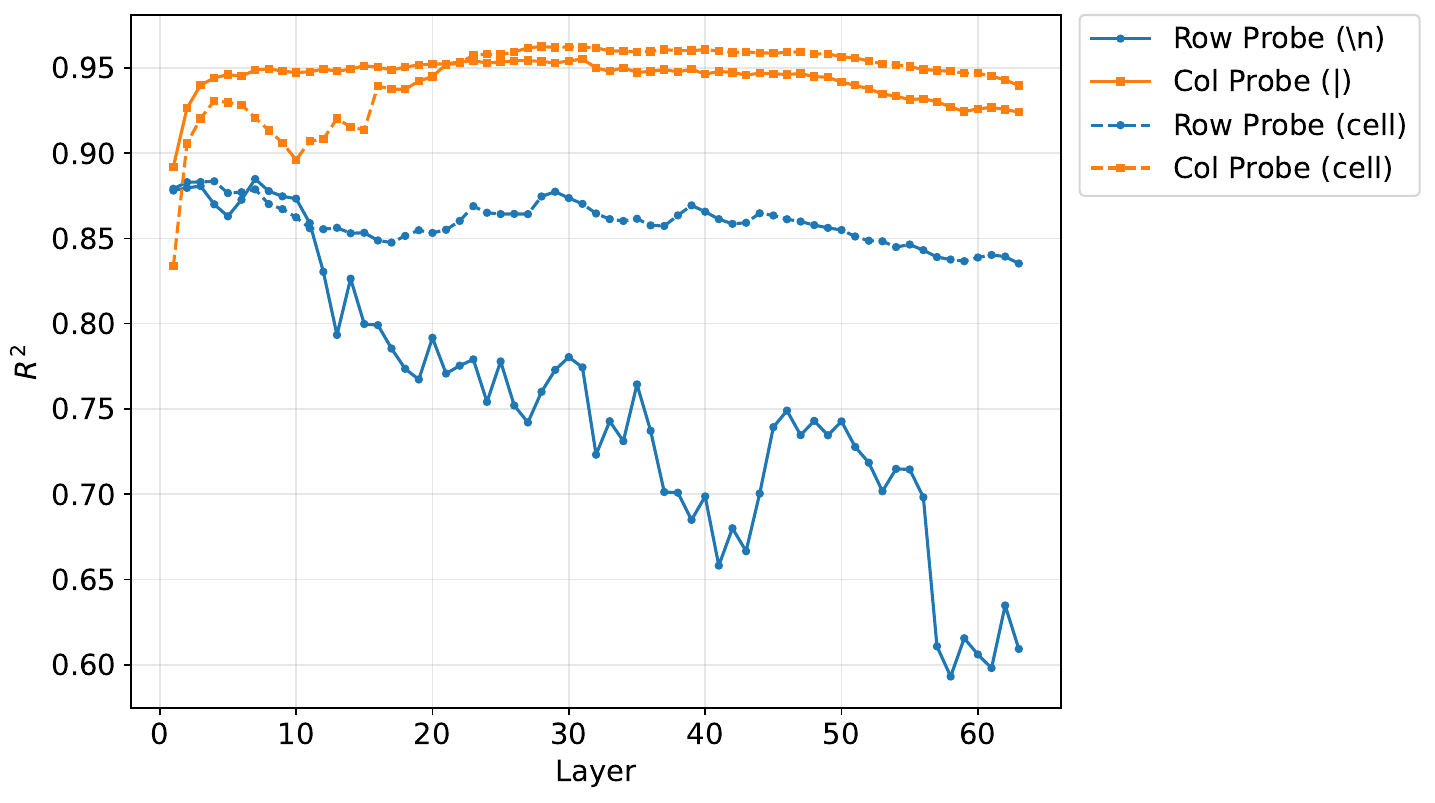}
        \caption{Qwen2.5-32B}
        \label{fig:Qwen2.5-32B_probe_r2}
    \end{subfigure}
    \hfill
    \begin{subfigure}[b]{0.48\textwidth}
        \centering
        \includegraphics[width=\linewidth]{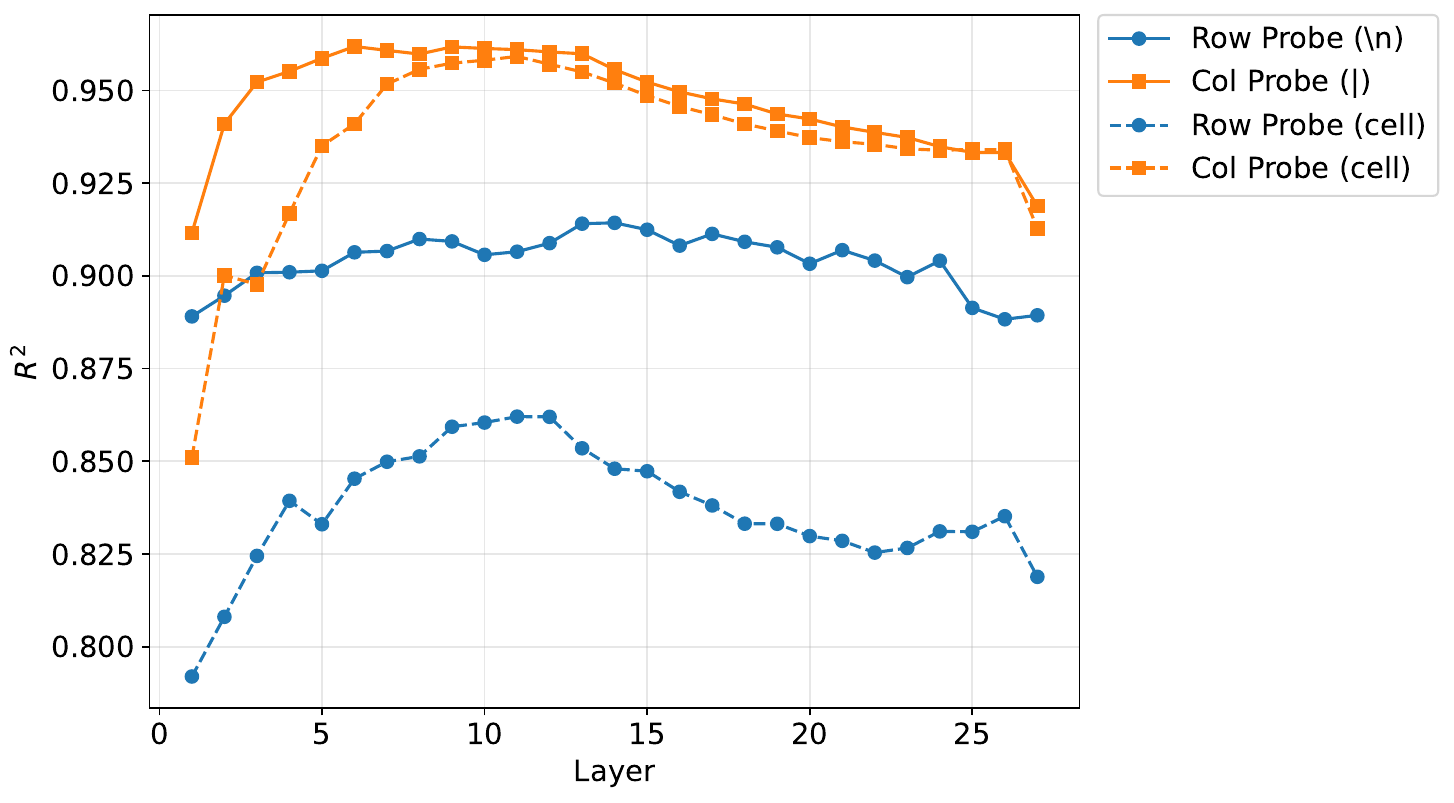}
        \caption{Llama-3.2-3B}
        \label{fig:Llama3.2-3B_probe_r2}
    \end{subfigure}
    \hfill
    \begin{subfigure}[b]{0.48\textwidth}
        \centering
        \includegraphics[width=\linewidth]{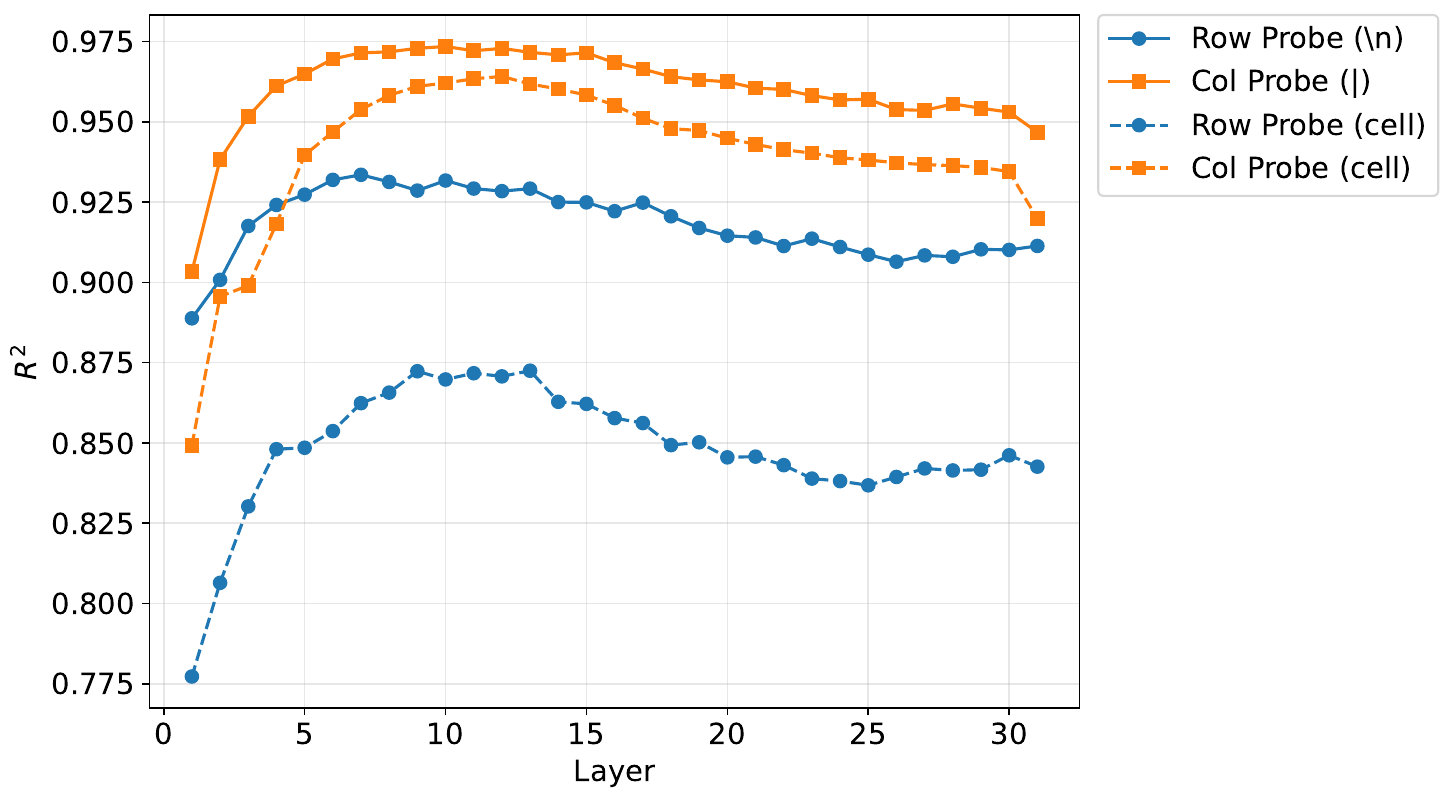}
        \caption{Llama-3.1-8B}
        \label{fig:Llama3.1-8B_probe_r2}
    \end{subfigure}
    
    \caption{
      $R^2$ of column probes and row probes trained for cells and delimiters across layers on various LLMs. 
    }
    \label{fig:probe_r2_llms}
\end{figure*}


\subsection{Implicit Coordinate System via Delimiter Counting}
\label{subapp:coordinate}

In this subsection, we extend the investigation of the implicit coordinate system to the broader model suite to evaluate whether the delimiter-driven counting mechanism is a universal feature of transformer-based table understanding.

We visualize the $R^2$ of linear probes for the supplementary models in Figure~\ref{fig:probe_r2}. Similar to the findings for Qwen-4B, all models exhibit a distinct performance gap between delimiter tokens and cell tokens in the early-to-middle layers.
All models achieve near-perfect column $R^2$ ($>0.95$) on delimiters extremely early (by Layer 5-10). The cell probes lag behind, converging to high accuracy only in the middle layers. This reinforces the hypothesis that delimiters act as the primary anchors for structural states, which are subsequently synchronized to cell hidden states.
Also, the column probes of all models exhibit consistently higher performance than row probes, confirming that the models utilize column indices as primary location mechanism, with treating row positions as secondary features derived from row boundaries.
Using trained probes from various LLMs that exhibit high $R^2$ scores, we successfully predict column and row indices. Our results reveal distinct sawtooth patterns and stepwise functions, which are consistent with the findings presented in \S\ref{subsec:Representations of Column and Row Indices}.

\begin{figure*}[t]
    \centering
    \begin{subfigure}[b]{0.48\textwidth} 
        \centering
        \includegraphics[width=\linewidth]{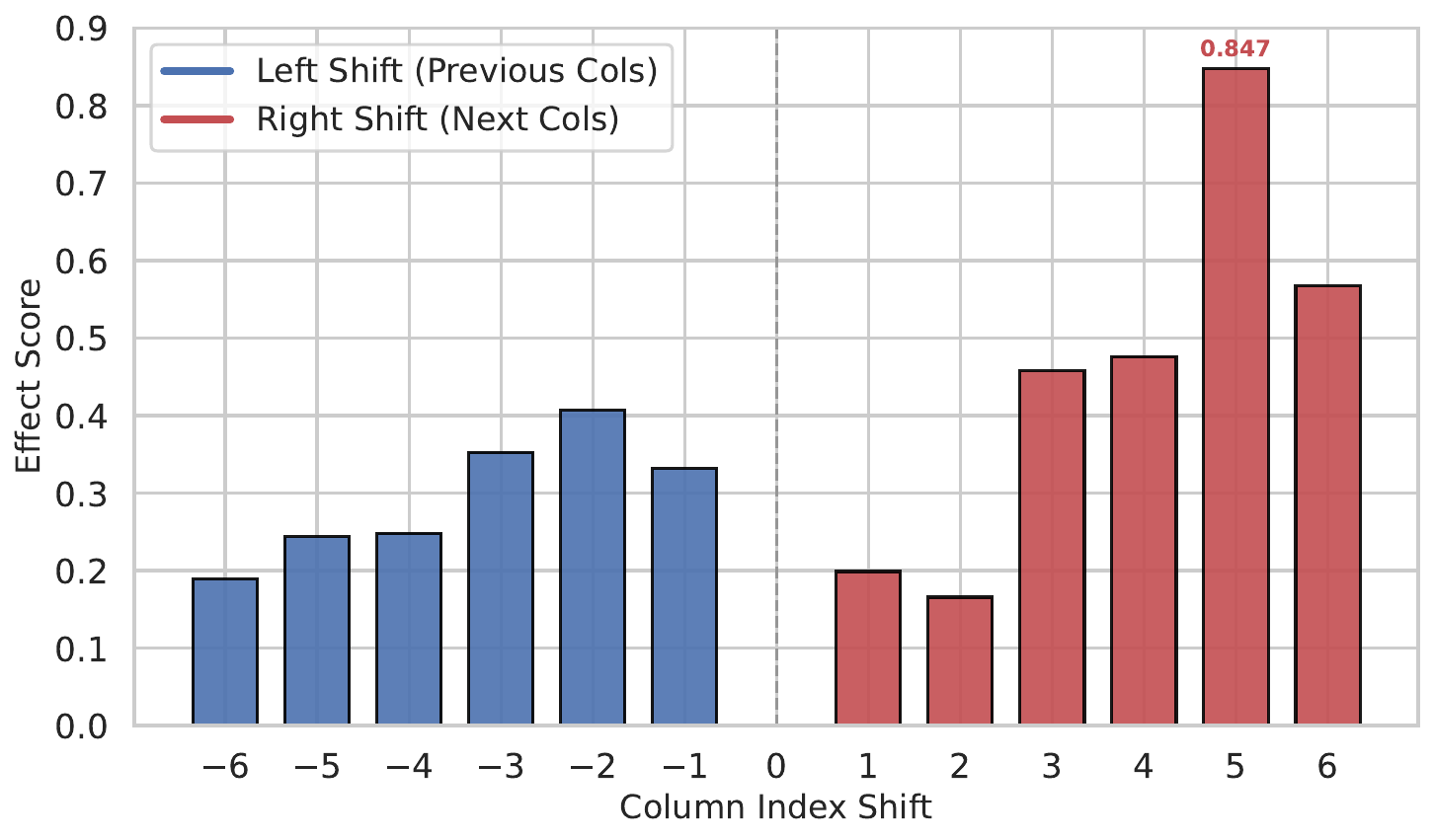}
        \caption{Qwen3-0.6B}
        \label{fig:Qwen3-0.6B_intervene_single_effect}
    \end{subfigure}
    \hfill
    \begin{subfigure}[b]{0.48\textwidth}
        \centering
        \includegraphics[width=\linewidth]{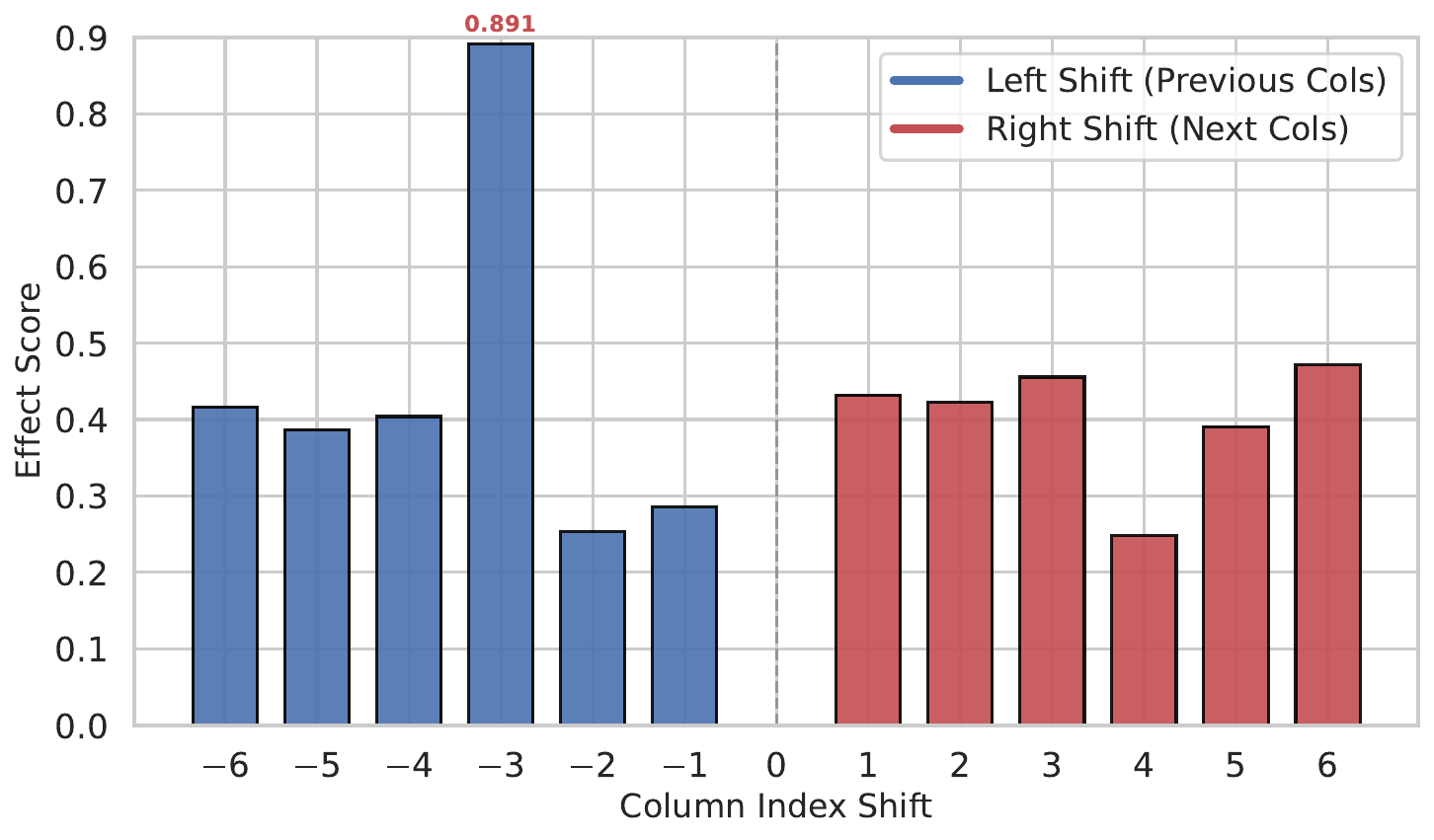}
        \caption{Qwen2.5-32B}
        \label{fig:Qwen2.5-32B_intervene_single_effect}
    \end{subfigure}
    \hfill
    \begin{subfigure}[b]{0.48\textwidth}
        \centering
        \includegraphics[width=\linewidth]{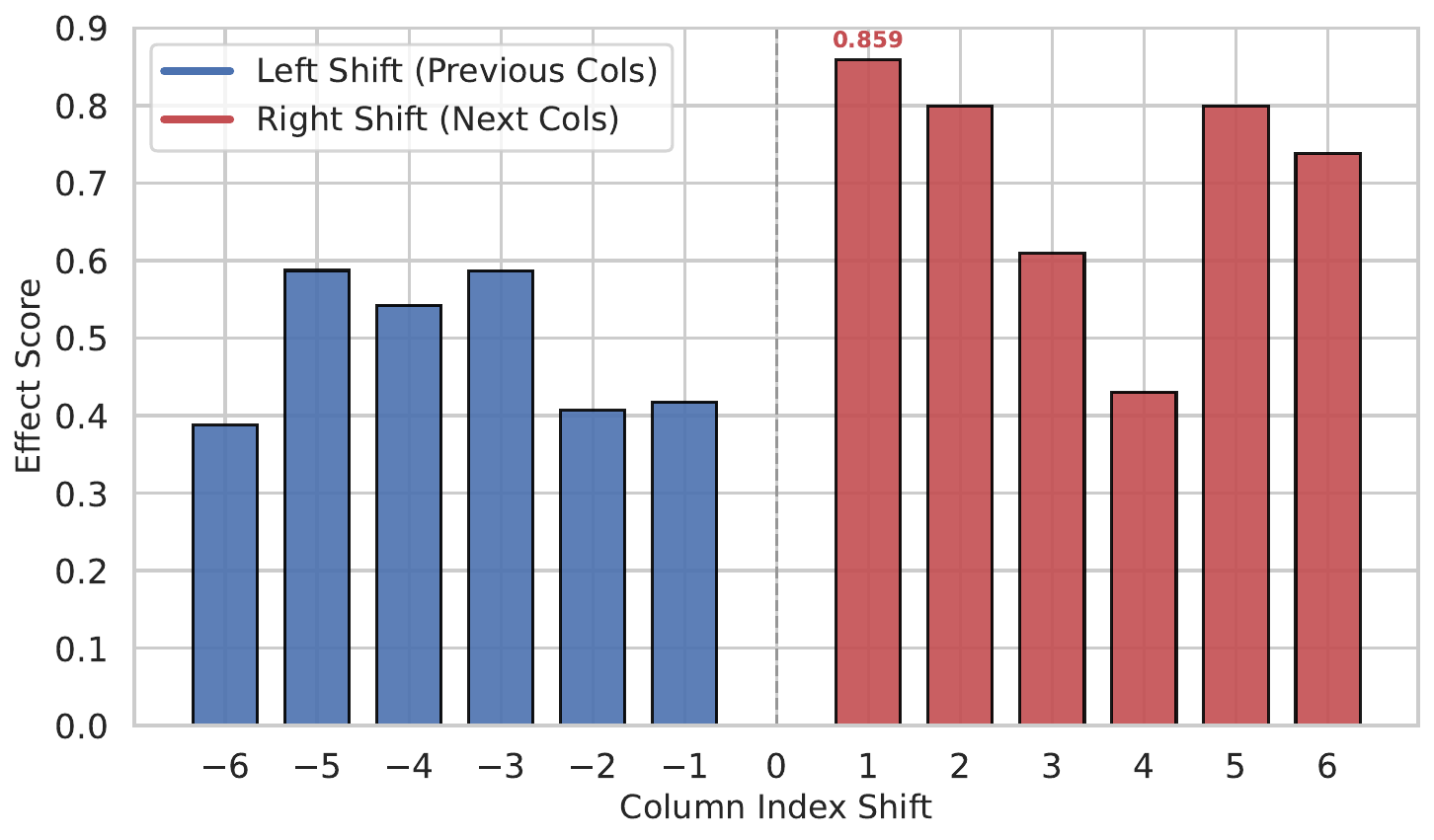}
        \caption{Llama-3.2-3B}
        \label{fig:Llama3.2-3B_intervene_single_effect}
    \end{subfigure}
    \hfill
    \begin{subfigure}[b]{0.48\textwidth}
        \centering
        \includegraphics[width=\linewidth]{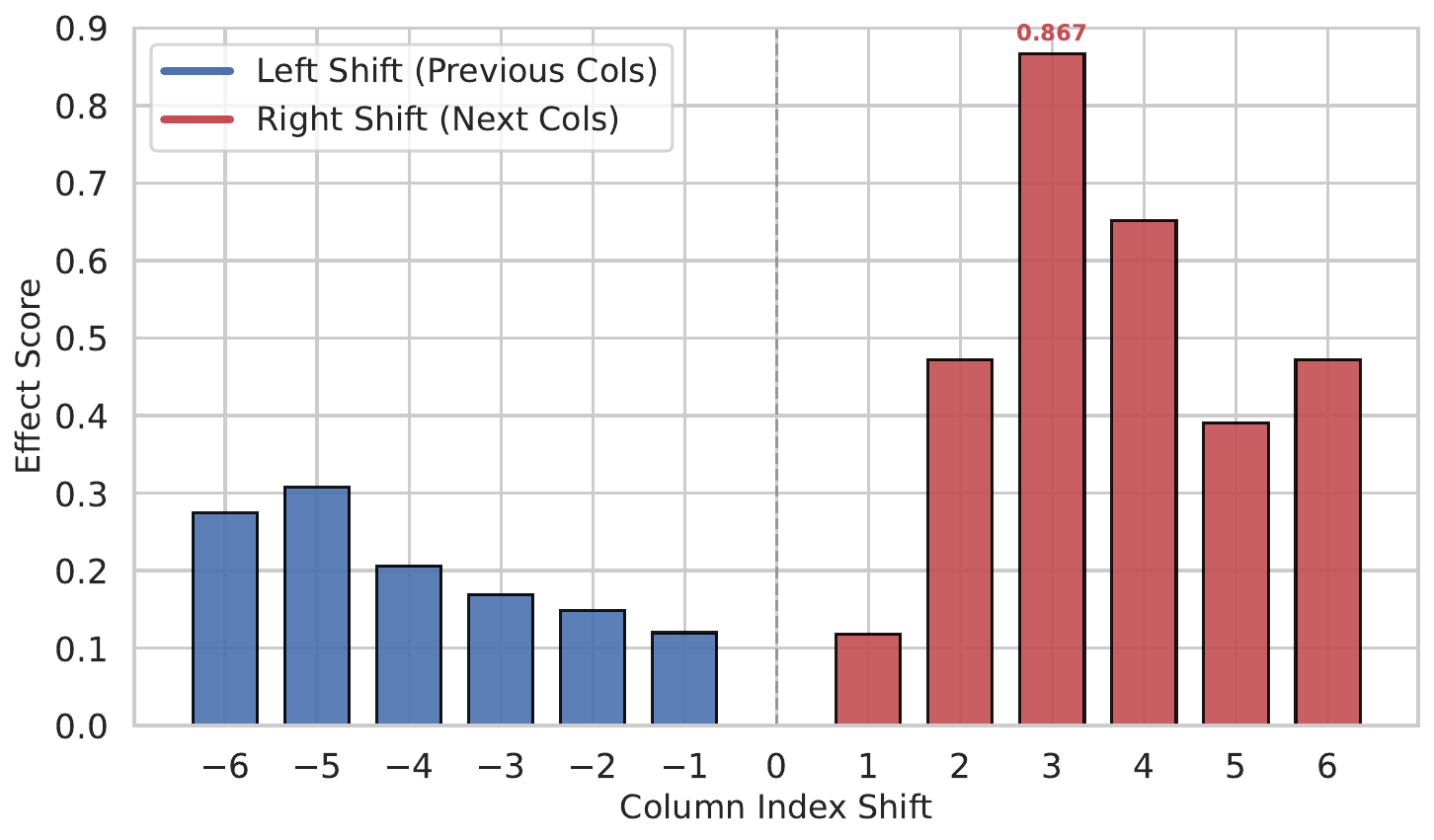}
        \caption{Llama-3.1-8B}
        \label{fig:Llama3.1-8B_intervene_single_effect}
    \end{subfigure}
    
    \caption{
      Effect Scores resulting from injecting the unit shift vector scaled by integer factors $k$ on various LLMs. The positive integers denote rightward shifts and negative integers denote leftward shifts. 
    }
    \label{fig:intervene_single_effect_llms}
\end{figure*}

\begin{figure*}[t]
    \centering
    \begin{subfigure}[b]{0.48\textwidth} 
        \centering
        \includegraphics[width=\linewidth]{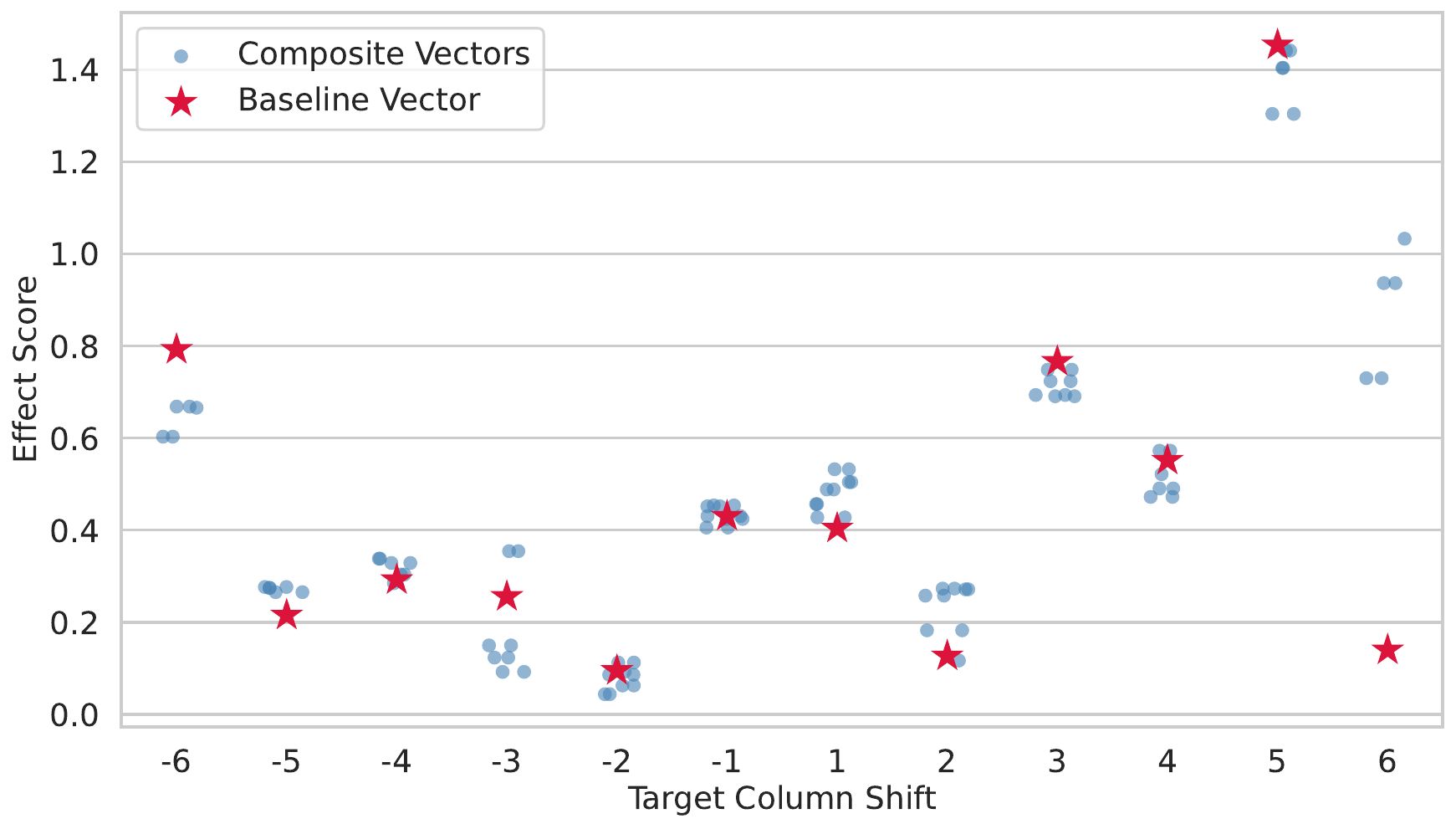}
        \caption{Qwen3-0.6B}
        \label{fig:Qwen3-0.6B_intervene_comb_effect}
    \end{subfigure}
    \hfill
    \begin{subfigure}[b]{0.48\textwidth}
        \centering
        \includegraphics[width=\linewidth]{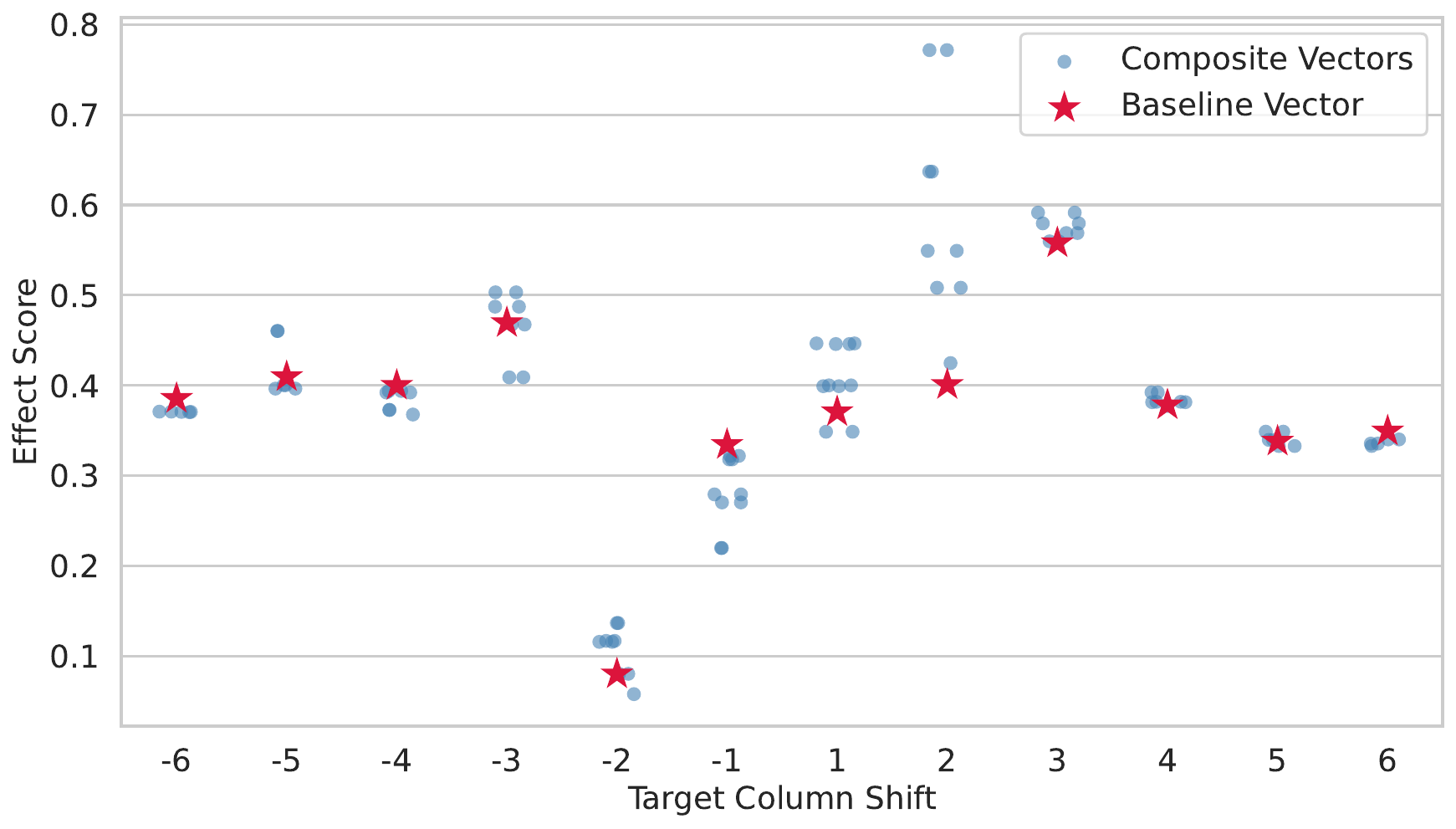}
        \caption{Qwen2.5-32B}
        \label{fig:Qwen2.5-32B_intervene_comb_effect}
    \end{subfigure}
    \hfill
    \begin{subfigure}[b]{0.48\textwidth}
        \centering
        \includegraphics[width=\linewidth]{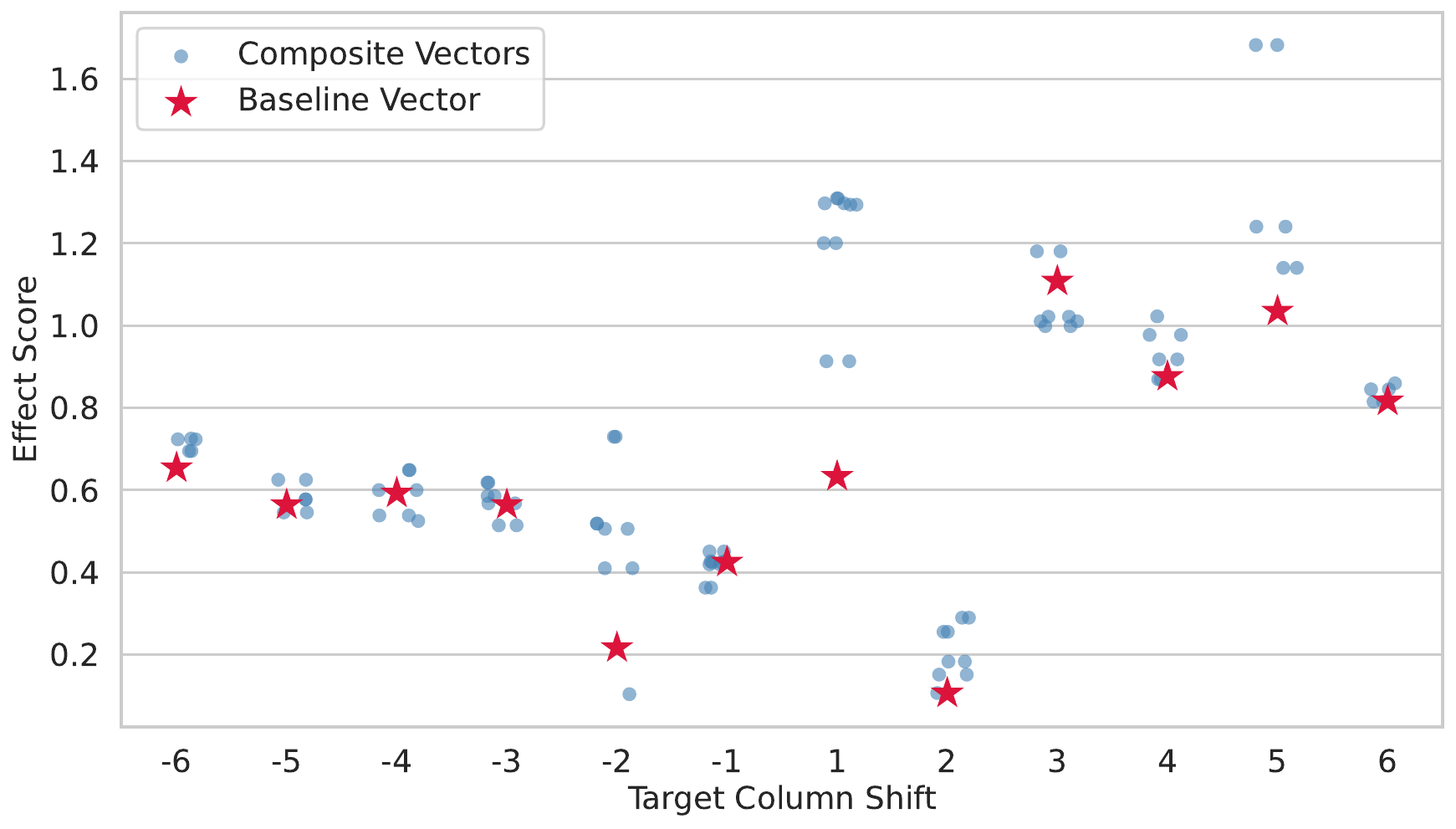}
        \caption{Llama-3.2-3B}
        \label{fig:Llama3.2-3B_intervene_comb_effect}
    \end{subfigure}
    \hfill
    \begin{subfigure}[b]{0.48\textwidth}
        \centering
        \includegraphics[width=\linewidth]{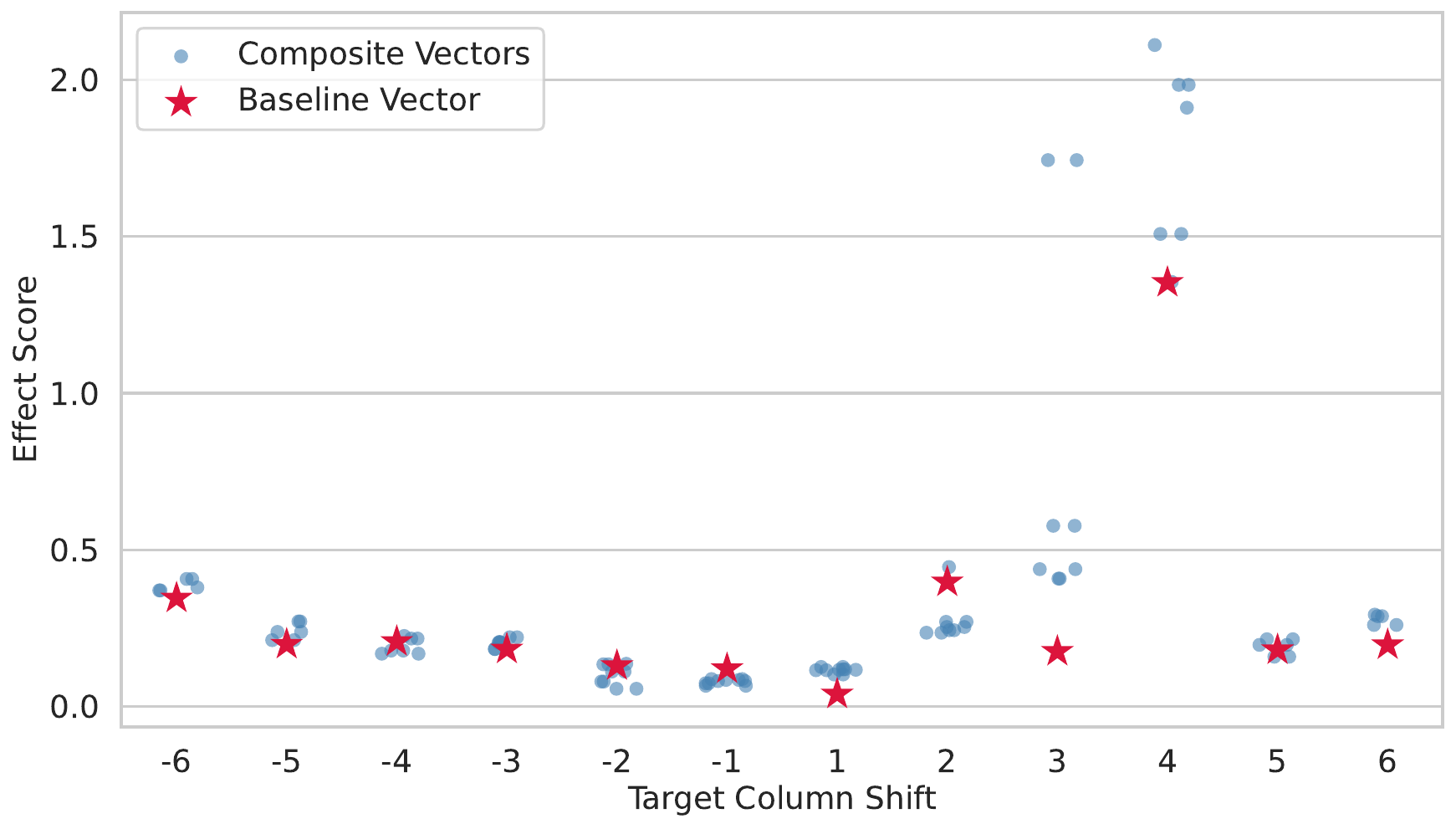}
        \caption{Llama-3.1-8B}
        \label{fig:Llama3.1-8B_intervene_comb_effect}
    \end{subfigure}
    
    \caption{
      Comparison of Effect Scores between Baseline Vectors (red stars) and Composite Vectors (blue dots) on various LLMs. The Baseline Vector represents the direct extraction of the shift vector for target offset $k$. The Composite Vectors are formed by summing two distinct vectors $\vec{v}_{a} + \vec{v}_{b}$ such that $a + b = k$.
    }
    \label{fig:intervene_comb_effect_llms}
\end{figure*}

\subsection{The Linear Geometry of Tabular Indices}
\label{subapp:geometry}

To assess the cross-model consistency of the linear geometry hypothesis, we replicate the vector steering and compositionality experiments (\S\ref{sec:geometry}) on more models. The goal is to determine if the representation of column indices as a linear subspace is a universal property of transformer-based table understanding.

\subsubsection{Steering with Unit Shift Vectors}
Following the setup in \S\ref{sec:unit_steering}, we inject the scaled unit shift vector $\vec{v}_{unit}$ into the residual stream of the column header tokens. Figure~\ref{fig:intervene_single_effect_llms} illustrates the resulting Effect Scores across varying integer factors $k$.

\begin{itemize}
    \item \textbf{Qwen Series:} Both Qwen3-0.6B and Qwen2.5-32B exhibit a robust linear response to the steering vector. In Qwen3-0.6B (Figure~\ref{fig:Qwen3-0.6B_intervene_single_effect}), the Effect Score peaks at $k=5$ with a value of $0.847$, demonstrating that even at a sub-billion parameter scale, the model maintains a structured linear progression for column navigation. Qwen2.5-32B (Figure~\ref{fig:Qwen2.5-32B_intervene_single_effect}) displays a similar trend, particularly for negative shifts, with $k=-3$ achieving a peak score of $0.891$. 
    \item \textbf{Llama Series:} The Llama models also demonstrate successful steering, with Llama-3.2-3B (Figure~\ref{fig:Llama3.2-3B_intervene_single_effect}) and Llama-3.1-8B (Figure~\ref{fig:Llama3.1-8B_intervene_single_effect}) reaching peak Effect Scores of $0.859$ and $0.867$ respectively. This success confirms that despite the sparser causal mass observed in earlier patching experiments, the Llama residual stream indeed maintains a functional coordinate system. The injected shift vector effectively alters the latent geometric pointers, which subsequently guide the late-layer attention heads to retrieve values from the steered column indices in the KV-cache.
\end{itemize}

\subsubsection{Compositionality of Column Vectors}
We evaluate the additivity of the shift vectors by comparing composite vectors $\vec{v}_{comp} = \vec{v}_{a} + \vec{v}_{b}$ against the baseline vectors $\vec{v}_{k}$ ($a+b=k$). The results are presented in Figure~\ref{fig:intervene_comb_effect_llms}.

The visual alignment between the red baseline stars and the blue composite dots across all four models provides definitive evidence for the linearity of the tabular index space.
\begin{itemize}
    \item In the \textbf{Qwen series} (Figures~\ref{fig:Qwen3-0.6B_intervene_comb_effect} and \ref{fig:Qwen2.5-32B_intervene_comb_effect}), the composite dots tightly cluster around the baseline stars. For Qwen2.5-32B, at target offsets like $k=2$ and $k=3$, the composite vectors yield performance comparable to the directly extracted baseline, confirming that the representation space is locally Euclidean.
    \item In the \textbf{Llama series} (Figures~\ref{fig:Llama3.2-3B_intervene_comb_effect} and \ref{fig:Llama3.1-8B_intervene_comb_effect}), we observe high-fidelity compositionality. Notably, in Llama-3.1-8B, the composite vectors at $k=3$ and $k=4$ exhibit a reinforcement effect, where the steering efficacy (Effect Score $> 1.5$) remains high across various $(a, b)$ combinations. This suggests that the directionality of the column-shift vector is highly stable across different positions in the latent space.
\end{itemize}

Across models and scales, the internal representation of column indices behaves as a linear subspace where structural navigation is equivalent to vector translation. This linear geometry explains how LLMs generalize to table dimensions unseen during training by simply extending the magnitude of these directional vectors within the coordinate-encoding subspace.

\begin{figure*}[t]
    \centering
    \begin{subfigure}[b]{0.24\textwidth} 
        \centering
        \includegraphics[width=\linewidth]{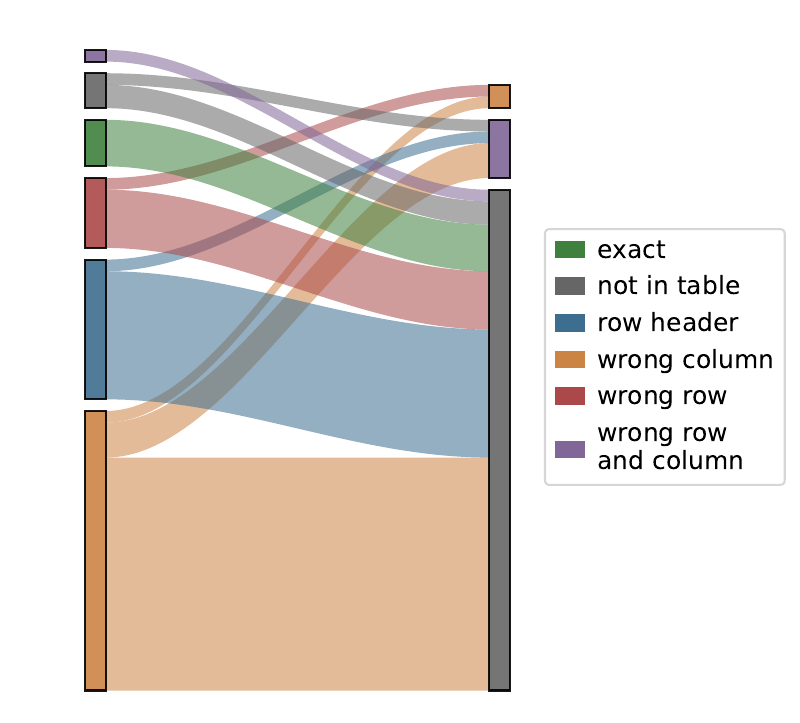}
        \caption{Qwen3-0.6B}
        \label{fig:Qwen3-0.6B_sankey_multi}
    \end{subfigure}
    \hfill
    \begin{subfigure}[b]{0.24\textwidth}
        \centering
        \includegraphics[width=\linewidth]{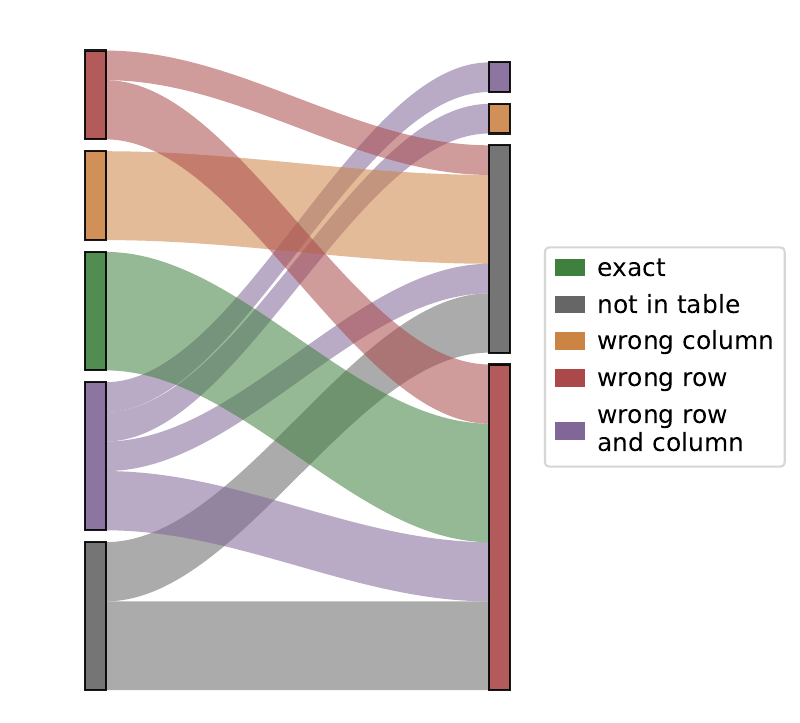}
        \caption{Qwen2.5-32B}
        \label{fig:Qwen2.5-32B_sankey_multi}
    \end{subfigure}
    \hfill
    \begin{subfigure}[b]{0.24\textwidth}
        \centering
        \includegraphics[width=\linewidth]{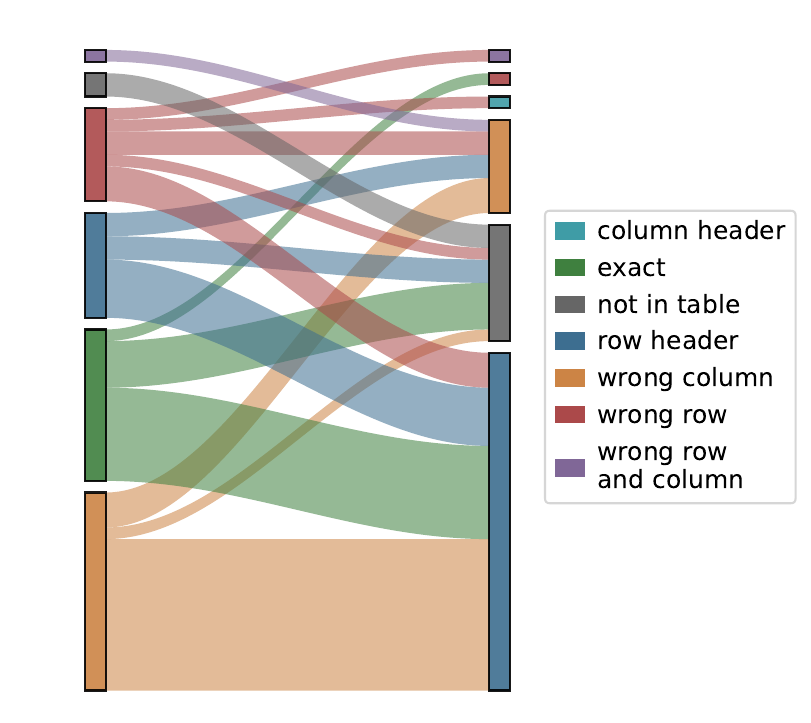}
        \caption{Llama-3.2-3B}
        \label{fig:Llama3.2-3B_sankey_multi}
    \end{subfigure}
    \hfill
    \begin{subfigure}[b]{0.24\textwidth}
        \centering
        \includegraphics[width=\linewidth]{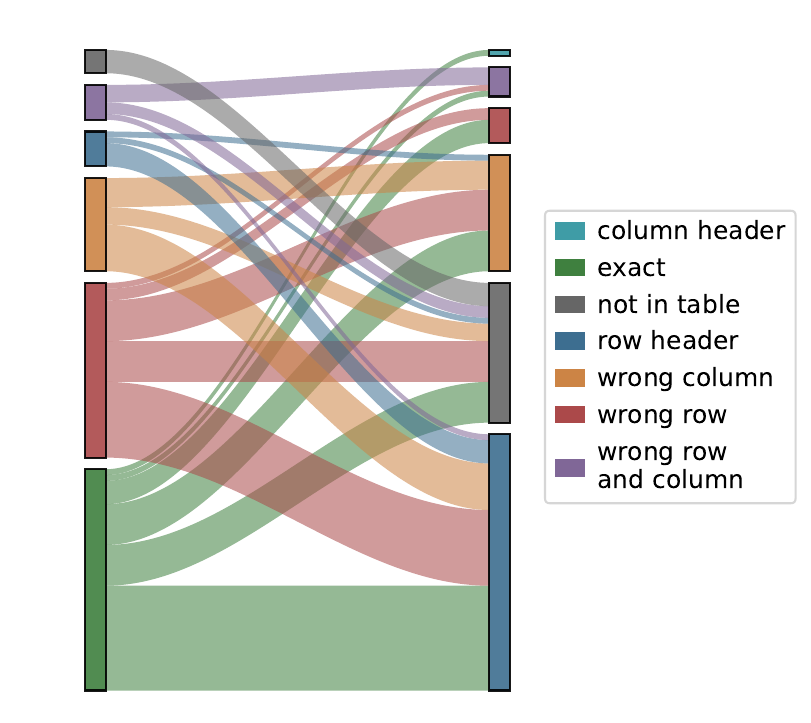}
        \caption{Llama-3.1-8B}
        \label{fig:Llama3.1-8B_sankey_multi}
    \end{subfigure}
    
    \caption{
    Mechanistic analysis on the multi-cell location task: the impact of ablating the row alignment heads of various LLMs identified in Stage I on multi-row queries. 
    }
    \label{fig:sankey_multi_llms}
\end{figure*}

\begin{figure*}[t]
    \centering
    \begin{subfigure}[b]{0.24\textwidth} 
        \centering
        \includegraphics[width=\linewidth]{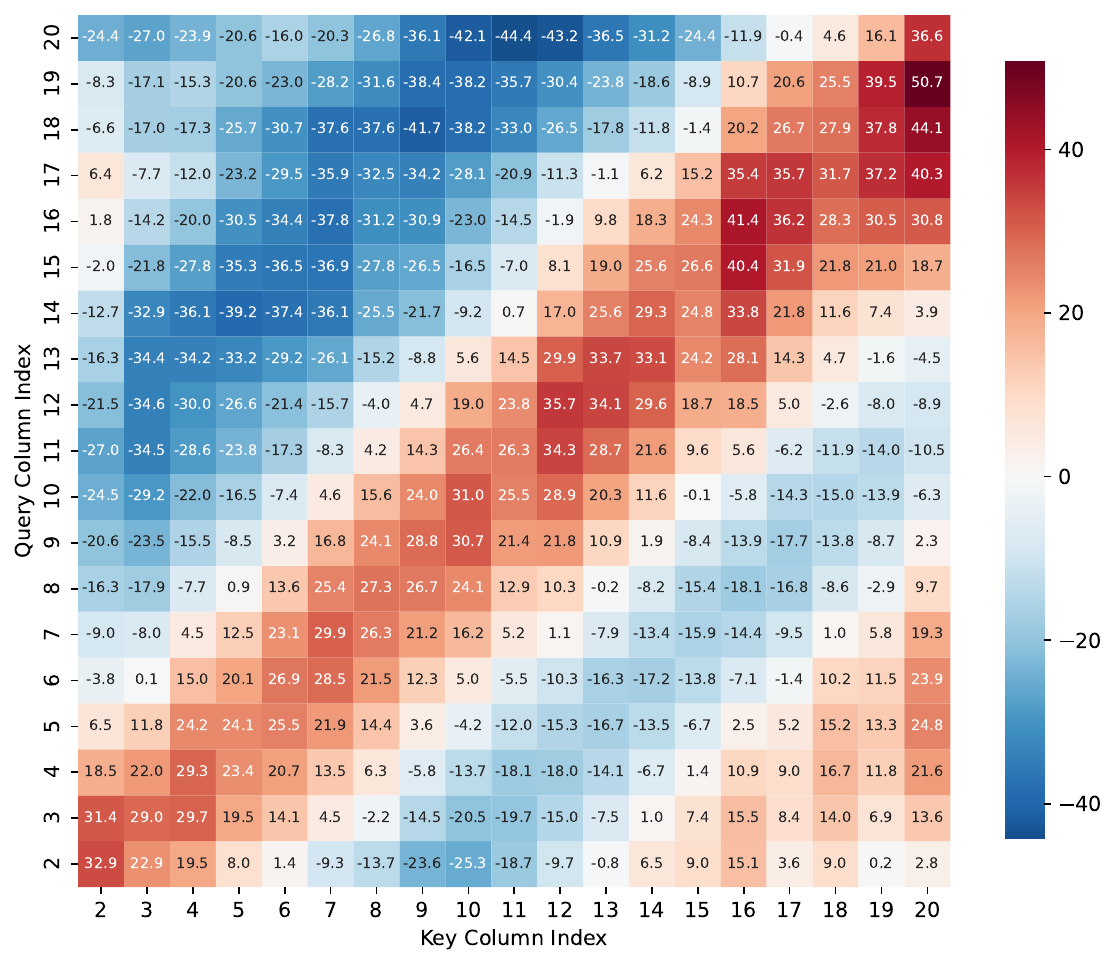}
        \caption{Qwen3-0.6B}
        \label{fig:Qwen3-0.6B_qk_heatmap_multi}
    \end{subfigure}
    \hfill
    \begin{subfigure}[b]{0.24\textwidth}
        \centering
        \includegraphics[width=\linewidth]{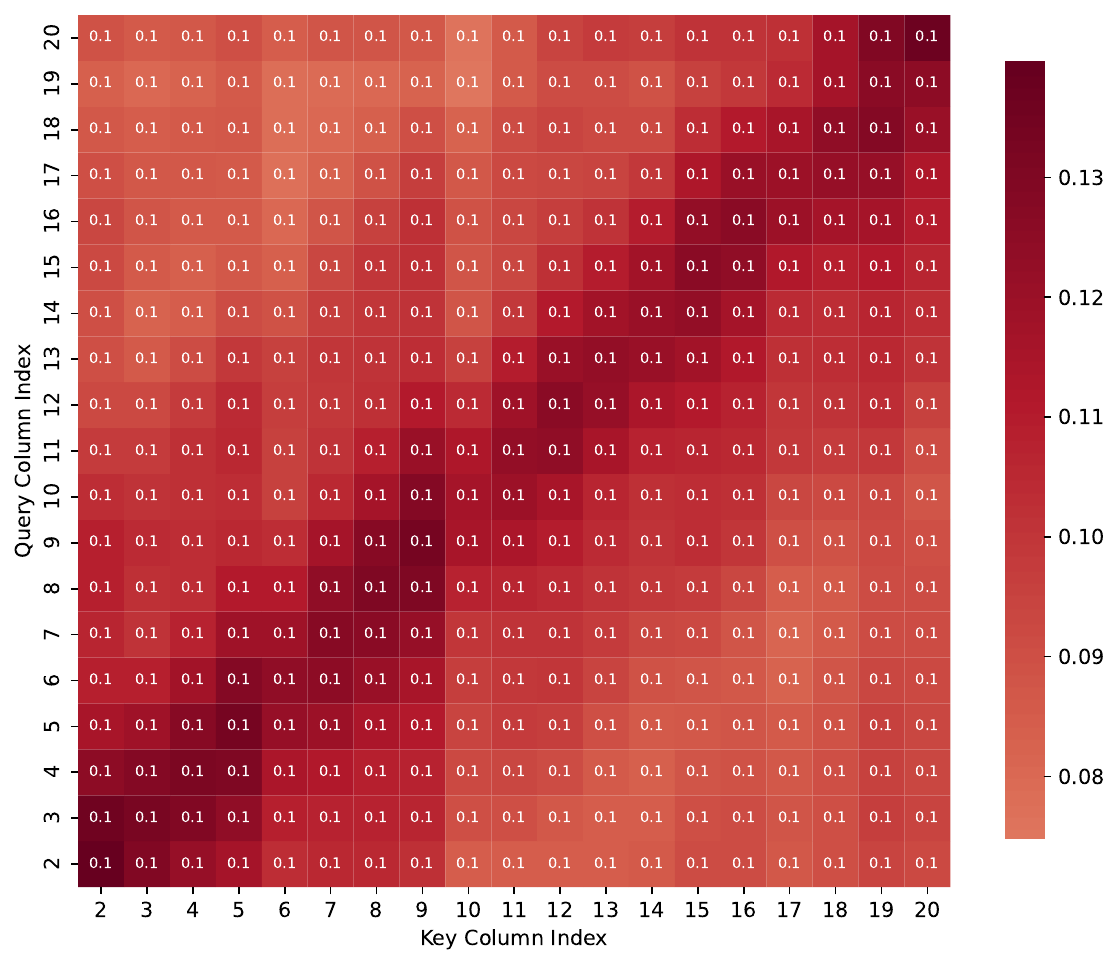}
        \caption{Qwen2.5-32B}
        \label{fig:Qwen2.5-32B_qk_heatmap_multi}
    \end{subfigure}
    \hfill
    \begin{subfigure}[b]{0.24\textwidth}
        \centering
        \includegraphics[width=\linewidth]{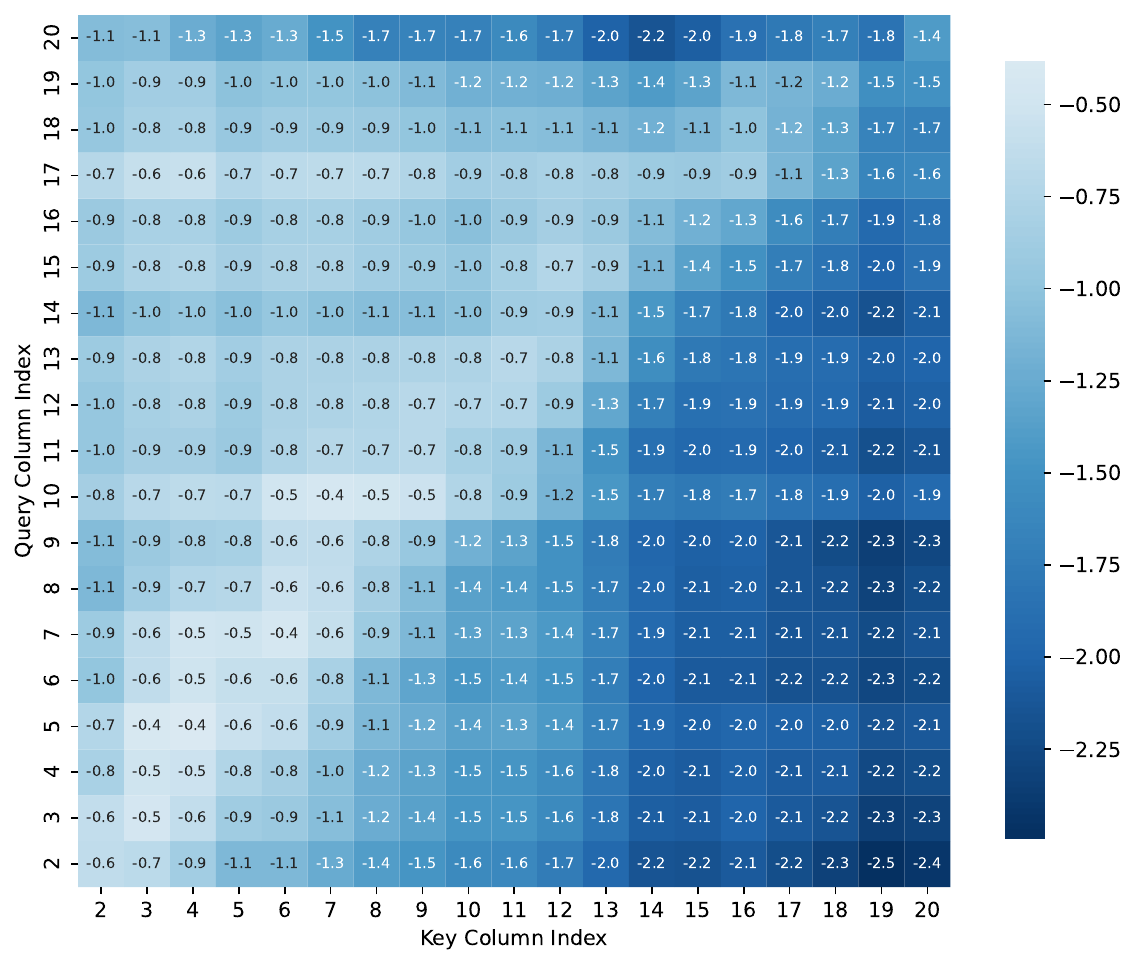}
        \caption{Llama-3.2-3B}
        \label{fig:Llama3.2-3B_qk_heatmap_multi}
    \end{subfigure}
    \hfill
    \begin{subfigure}[b]{0.24\textwidth}
        \centering
        \includegraphics[width=\linewidth]{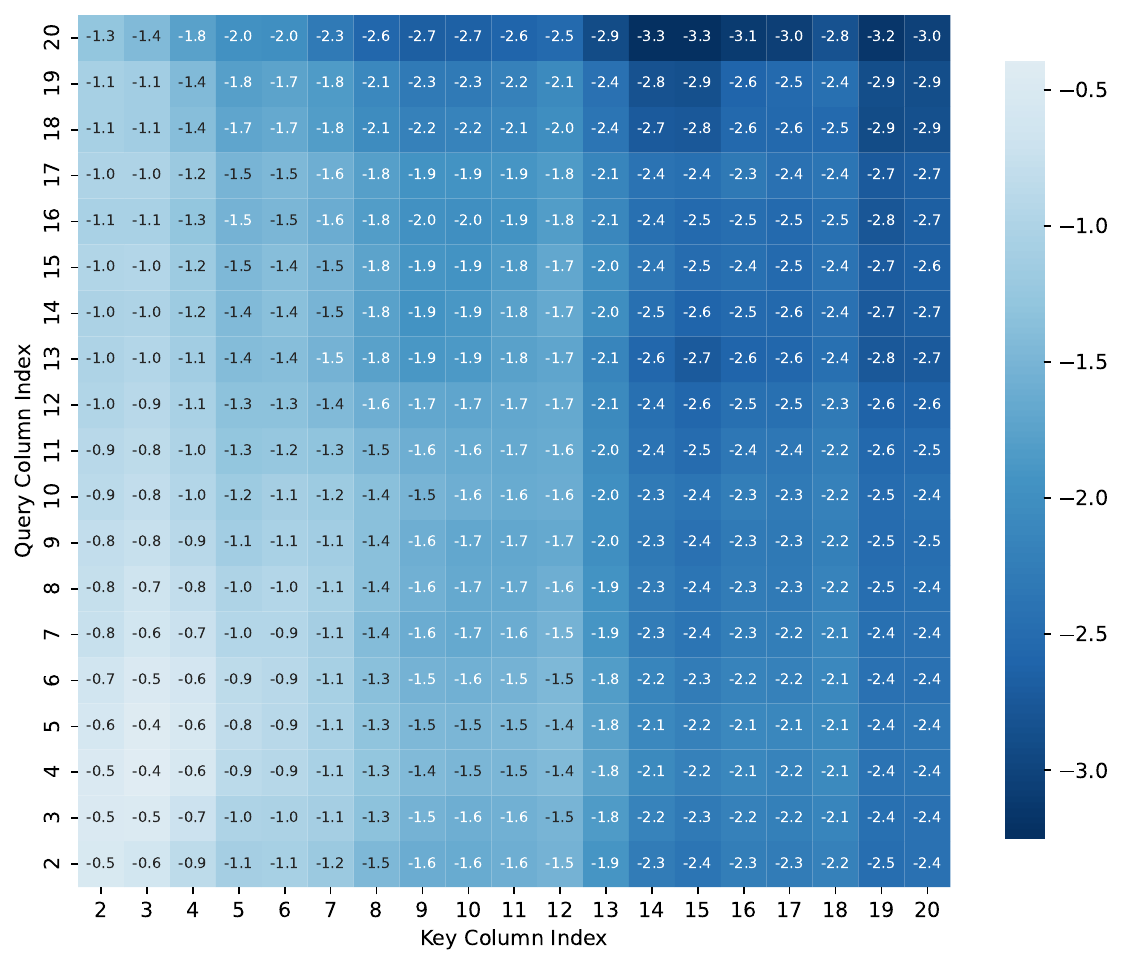}
        \caption{Llama-3.1-8B}
        \label{fig:Llama3.1-8B_qk_heatmap_multi}
    \end{subfigure}
    
    \caption{
        Mechanistic analysis on the multi-cell location task: the Query-Key interaction scores for the top-20 coordinate-encoding heads of LLMs from Stage II on multi-column queries. 
    }
    \label{fig:qk_heatmap_multi_llms}
\end{figure*}

\subsection{Generalization to Multi-Cell Location}
\label{subapp:generalization multi-cell}

To evaluate the universality and robustness of the parallel processing pipeline, we extend the mechanistic analysis of multi-cell retrieval to other models. Following the experimental setup in \S\ref{sec:complex}, we examine whether the specialized functional heads identified in the atomic task maintain their roles when processing multiple constraints concurrently.

\subsubsection{Parallel Semantic Binding across LLMs}
As illustrated in Figure~\ref{fig:sankey_multi_llms}, zero-ablating the top-20 \textit{row alignment heads} triggers a consistent performance collapse across all models, validating that parallel semantic binding is a fundamental mechanism in transformer-based table understanding.
\begin{itemize}
    \item \textbf{Qwen Series:} For Qwen3-0.6B and Qwen2.5-32B (Figures~\ref{fig:Qwen3-0.6B_sankey_multi} and \ref{fig:Qwen2.5-32B_sankey_multi}), we observe a catastrophic drop in \texttt{exact} matches. For instance, Qwen2.5-32B shows a significant transition toward \texttt{not in table} and \texttt{wrong row} errors. This confirms that these heads are essential for anchoring the multiple query row constraints to their respective table headers. The similar transitions suggest that the row alignment operator naturally scales to multi-target scenarios via the attention mechanism's inherent parallelism.
    \item \textbf{Llama Series:} In Llama-3.2-3B and Llama-3.1-8B (Figures~\ref{fig:Llama3.2-3B_sankey_multi} and \ref{fig:Llama3.1-8B_sankey_multi}), the ablation results in a substantial increase in \texttt{row header} errors. The prevalence of \texttt{row header} errors further supports the hypothesis that without these alignment heads, the Llama series fails to bridge the gap between the query and the table content, defaulting to a literal copy of the query tokens.
\end{itemize}

\subsubsection{Multi-Threaded Coordinate Localization across LLMs}
Figure~\ref{fig:qk_heatmap_multi_llms} presents the interaction scores $S_{j,l}$ for the \textit{coordinate-encoding heads} under multi-column queries. The results demonstrate that the structural navigation mechanism generalizes effectively to parallel index resolution.
\begin{itemize}
    \item \textbf{Qwen Series:} Figures~\ref{fig:Qwen3-0.6B_qk_heatmap_multi} and \ref{fig:Qwen2.5-32B_qk_heatmap_multi} exhibit clear, multi-modal diagonal patterns. 
    This indicates that the coordinate-encoding heads utilize RoPE to create a multi-threaded geometric alignment, where the query representations for different columns concurrently target their respective spatial locations in the serialized table.
    \item \textbf{Llama Series:} Consistent with the findings in the atomic task, Llama-3.2-3B and Llama-3.1-8B (Figures~\ref{fig:Llama3.2-3B_qk_heatmap_multi} and \ref{fig:Llama3.1-8B_qk_heatmap_multi}) exhibit a negative diagonal pattern. The dark blue regions aligned with the queried column indices suggest that Llama utilizes these heads as inhibitory filters to mask non-target columns. In the multi-cell setting, this mechanism functions by suppressing all columns except those specified in the query, effectively isolating multiple KV-cache signals for parallel retrieval. While the sign of the interaction score differs from Qwen, the preservation of the structural alignment across multiple column indices confirms that the geometric localization logic is preserved.
\end{itemize}

In conclusion, the consistency of these patterns across varied models and scales reinforces the three-stage information flow. Regardless of model size or specific attention behavior (excitatory vs. inhibitory), LLMs resolve complex multi-cell queries by reusing the same specialized circuit identified in simpler tasks, executing semantic and geometric operations in a single, parallelized forward pass.

\end{document}